%% file: main.tex
\documentclass[10pt,twocolumn,letterpaper]{article}

\usepackage[pagenumbers]{cvpr} 

\input{preamble}

\definecolor{cvprblue}{rgb}{0.21,0.49,0.74}
\usepackage[pagebackref,breaklinks,colorlinks,allcolors=cvprblue]{hyperref}

\def\paperID{7314} 
\def\confName{CVPR}
\def\confYear{2026}

\title{CURE: Curriculum-guided Multi-task Training for Reliable Anatomy Grounded Report Generation}

\author{
    Pablo Messina\textsuperscript{1,2,3}, 
    Andrés Villa\textsuperscript{4}, 
    Juan León Alcázar\textsuperscript{4}, 
    Karen Sanchez\textsuperscript{4}, \\
    Carlos Hinojosa\textsuperscript{4}, 
    Denis Parra\textsuperscript{1,2,3}, 
    Alvaro Soto\textsuperscript{1,2}, 
    Bernard Ghanem\textsuperscript{4} \\
    \textsuperscript{1}Pontificia Universidad Católica de Chile, 
    \textsuperscript{2}CENIA, 
    \textsuperscript{3}iHEALTH, 
    \textsuperscript{4}KAUST \\
    {\tt\small pamessina@uc.cl, \{andres.villa,juancarlo.alcazar,karen.sanchez,carlos.hinojosa\}@kaust.edu.sa} \\ {\tt\small \{dparra,asoto\}@ing.puc.cl, bernard.ghanem@kaust.edu.sa}
}

\begin{document}
\maketitle

\input{sec_0_abstract}
\input{sec_1_intro}
\input{sec_2_related_work}
\input{sec_3_methodology}

\input{sec_4_experiments}

\input{sec_5_conclusion}


\section*{Acknowledgments}
This work was conducted while P. Messina was a remote research intern at the Image and Video Understanding Lab (IVUL) at KAUST, under the supervision of B. Ghanem. P. Messina was supported by the ANID Scholarship Program (Doctorado Becas Chile 2019-21191569). We also acknowledge the support of Fondecyt grant 1231724. This work was also funded by ANID - Millennium Science Initiative Program - ICN2021\_004 (iHEALTH) as well as ICN17\_002 (IMFD), and by the National Center for Artificial Intelligence (CENIA) FB210017, Basal Funds for Centers of Excellence (ANID). The research reported in this publication was supported by funding from King Abdullah University of Science and Technology (KAUST) - Center of Excellence for Generative AI, under award number 5940.

{
    \small
    \bibliographystyle{ieeenat_fullname}
    \bibliography{main}
}

\input{sec_6_suppl}

\end{document}

%% file: preamble.tex
\usepackage{tabularx}
\usepackage{booktabs}
\usepackage{multirow}
\usepackage{array}
\usepackage[htt]{hyphenat}
\usepackage[table]{xcolor}
\usepackage{listings}
\usepackage{xcolor}
\usepackage{colortbl}
\usepackage{pifont}
\usepackage[accsupp]{axessibility}

\newcommand{\raw}[1]{%
  \texttt{\detokenize{#1}}%
}

\definecolor{badred}{RGB}{220, 50, 50}
\definecolor{goodgreen}{RGB}{50, 150, 50}

\lstdefinestyle{promptstyle}{
    basicstyle=\ttfamily\scriptsize, 
    breaklines=true,                 
    breakatwhitespace=true,          
    columns=fullflexible,            
    frame=single,                    
    rulecolor=\color{gray!50},       
    backgroundcolor=\color{gray!5},  
    aboveskip=1em,
    belowskip=1em
}

\newcommand{\mysection}[1]{\vspace{2pt}\noindent\textbf{#1}}

%% file: sec_0_abstract.tex
\begin{abstract}
Medical vision–language models can automate the generation of radiology reports but struggle with accurate visual grounding and factual consistency. Existing models often misalign textual findings with visual evidence, leading to unreliable or weakly grounded predictions. We present ``CURE'', an error-aware curriculum learning framework that improves grounding and report quality without any additional data. CURE tunes a multimodal instructional model on phrase grounding, grounded report generation, and anatomy-grounded report generation using public datasets. The method dynamically adjusts sampling based on model performance emphasizing harder samples to improve spatial and textual alignment. CURE improves grounding accuracy by +0.35 IoU, boosts report quality by +0.192 CXRFEScore, and reduces hallucinations by 18.6\%. CURE is a data-efficient framework that enhances both grounding accuracy and report reliability.
Code is available at \url{https://github.com/PabloMessina/CURE} and model weights at \url{https://huggingface.co/pamessina/medgemma-4b-it-cure}.
\end{abstract}

%% file: sec_1_intro.tex
\vspace{-0.2cm}
\section{Introduction}
\label{sec:intro}

\begin{figure}[t]
  \centering
  \includegraphics[width=\linewidth]{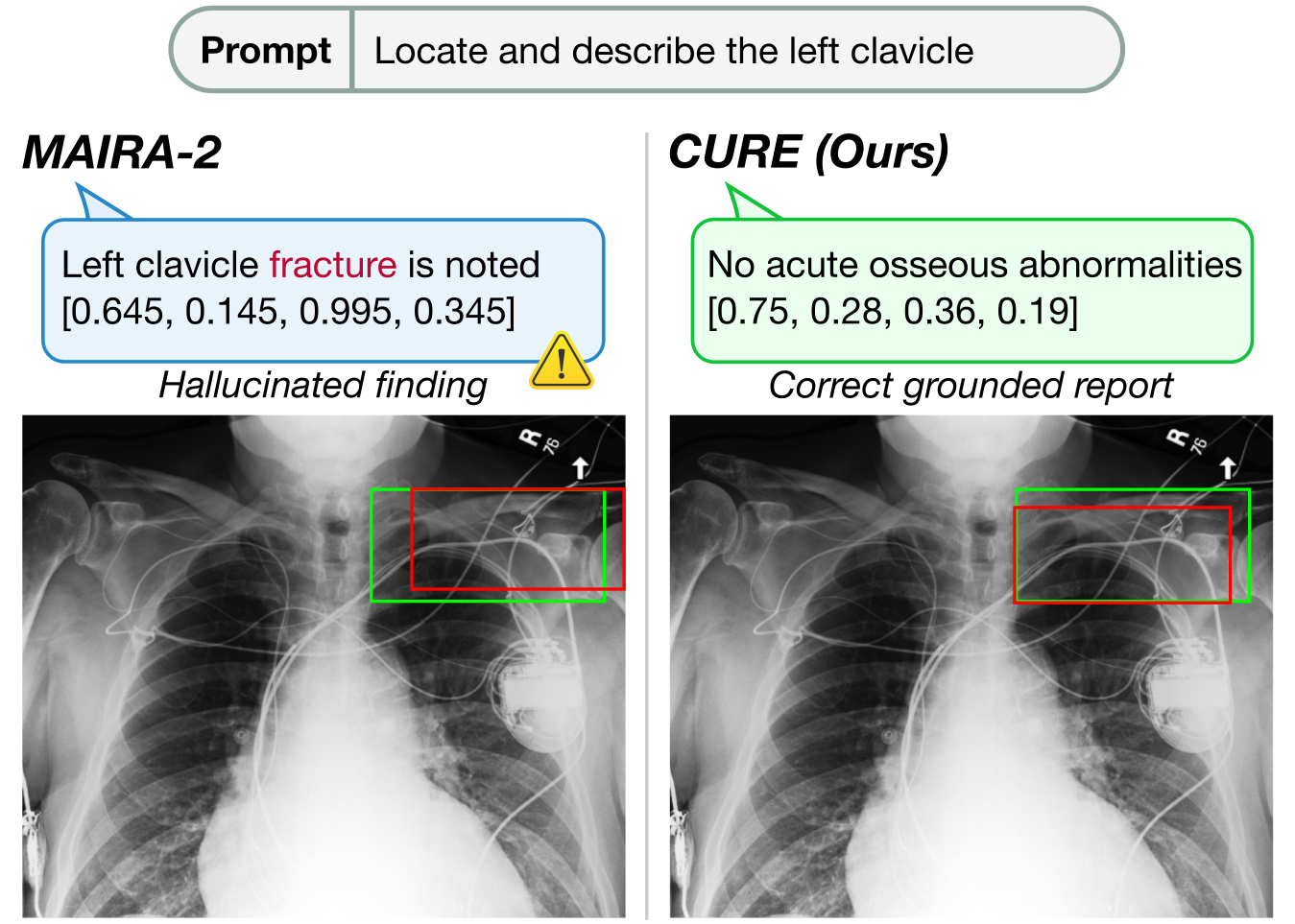}
  \caption{
  \textbf{False Positive Detection of Pathologies.} Given the same chest X-ray input from the MIMIC-CXR test set, both models approximate the location of the left clavicle. However, the baseline model (MAIRA-2) hallucinates a fracture (there is no fracture in the image), whereas our proposed model (CURE) generates a clinically correct and visually grounded description.}
  \label{fig:framework}
  \vspace{-0.4cm}
\end{figure}

In the medical domain, Vision-language Models (VLMs) enable the automatic generation of diagnostic reports from imaging exams, thereby reducing the workload of specialists and promoting standardized diagnostic pipelines \cite{maira2, llavamed, medgemma}. Despite their remarkable success in bridging visual and textual modalities \cite{clip, Siglip, llava, Qwen2.5-VL, llavamore}, VLMs still face fundamental challenges when applied to specialized domains such as medicine. The factuality and reliability of VLMs remain a significant concern, as current architectures often generate outputs that are inconsistent with their visual input, a phenomenon known as hallucinations \cite{MMVP, merlim, POPE}.

Medical VLMs~\cite{medgemma, llavarad, llavamed} rely on large-scale, domain-specific datasets to adapt general-purpose multi-modal models for medical applications. While this standard fine-tuning strategy improves performance on medical benchmarks, the resulting models often lack visual grounding, thus limiting interpretability and increasing the risk of hallucinations \cite{eagle}. For clinical adoption, it is crucial that VLMs produce factually accurate outputs and correctly ground medical findings in the relevant image regions. Strengthening factuality and grounding is therefore essential to ensure transparency, trustworthiness, and the reliable deployment of medical VLMs in real-world workflows.





To address these drawbacks, state-of-the-art models such as MAIRA-2~\cite{maira2} explicitly incorporate grounded report generation to better align local visual evidence with the textual findings. For example, MAIRA-2 is trained on a diverse set of localization-based tasks, including phrase grounding and grounded report generation. However, as shown in Figure~\ref{fig:framework}, current models often exhibit a bias in which the target visual regions become overly associated with abnormal findings, leading to false positives in medical reports.



To address these limitations, we propose \textbf{CURE} (\textbf{CU}rriculum-guided Multi-task Training for \textbf{RE}liable Anatomy Grounded Report Generation), a curriculum-based learning framework that enhances the reliability and visual grounding of medical VLMs without requiring additional data. CURE improves the standard training pipeline, and obtains improved results using only a subset of publicly available chest X-ray datasets. In particular, CURE replaces the conventional finding-generation objective used in state-of-the-art methods with an anatomy-grounded report generation (AGRG) task, thus leveraging the most detailed regional annotations provided by some datasets.




Our contributions are threefold: \textbf{i)} We introduce CURE, a novel error-aware curriculum framework that dynamically adjusts sampling distributions based on model performance, enhancing visual grounding without requiring any additional training data. \textbf{ii)} Our training strategy enables the effective transfer of grounding capabilities to medical VLMs that originally lack visual grounding mechanisms. These models surpass the current state of the art in visual grounding, despite being trained on less training data. \textbf{iii)} On the Chest ImaGenome dataset, CURE achieves an 18.6\% reduction in hallucinations across six key anatomical regions, substantially improving the reliability and trustworthiness of medical VLM outputs and setting a new state-of-the-art in visually grounded report generation.



%% file: sec_2_related_work.tex
\section{Related Work}
\label{sec:related_work}

\vspace{-0.1cm}
\mysection{Medical Vision-language Models.} The adoption of Vision-language models (VLMs) in the medical domain has advanced rapidly in recent years~\cite{singhal2023large, nath2025vila, jeong2024medical, kalpelbe2025vision}. Early approaches primarily targeted clinical report generation and captioning tasks, while more recent models such as Med-PaLM~\cite{singhal2023large, singhal2025toward}, MAIRA-2~\cite{bannur2024maira}, and the recently released MedGemma~\cite{sellergren2025medgemma} exhibit strong multi-modal reasoning capabilities across diverse medical modalities. Despite their success, the challenge of efficiently fine-tuning these large models on heterogeneous and task-specific medical datasets remains largely unsolved. Our work builds upon this line of research by introducing a structured, error-aware training curriculum that enhances performance without requiring additional data or model parameters.

\mysection{Multi-Task Learning in Medical Imaging.} Multi-task learning (MTL) is a widely adopted paradigm for jointly training models on related objectives such as classification, segmentation, and report generation~\cite{caruana1993multitask, zhang2021survey, liu2023hierarchical}. In medical imaging, MTL has been successfully applied across modalities, improving both efficiency and generalization~\cite{kim2023cross, boutillon2021multi,sainz2020multi, chen2018multi, weninger2019multi, wimmer2021multi, mohamed2025deepchest}. For instance, Sainz \etal~\cite{sainz2020multi} leveraged MTL for breast cancer screening by jointly learning classification and detection of abnormal findings in mammography; Chen \etal~\cite{chen2018multi} improved atrial segmentation and classification using MRI; and Weninger \etal~\cite{weninger2019multi} enhanced brain tumor segmentation by coupling detection and image reconstruction tasks in brain MRI. Despite these advances, MTL performance strongly depends on how task contributions are balanced during training \cite{kendall2018multi}. Improper weighting can cause task dominance and negative transfer~\cite{sener2018multi, yu2020gradient, mohamed2025deepchest}, degrading overall performance. To address this limitation, we propose an adaptive curriculum that dynamically schedules data exposure based on model performance, allowing the network to automatically prioritize under-performing tasks rather than relying on manually tuned loss weights.


\mysection{Curriculum Learning.} Curriculum learning (CL)\cite{bengio2009curriculum,wang2021survey} trains models by gradually increasing task difficulty, starting from easier samples and progressing to harder ones, in analogy to human learning. CL has been successfully applied to medical imaging~\cite {shibu2025medsam, li2023dynamic}. For instance, Shibu \etal~\cite{shibu2025medsam} introduced a MedSAM-guided CL strategy for white matter tract segmentation, where the curriculum is defined by anatomical block complexity to progressively refine spatial representations. Similarly, Li \etal~\cite{li2023dynamic} introduced a dynamic CL framework for medical image classification that derives sample difficulty from in-domain uncertainty estimates (via a Dirichlet classifier) and adapts the sampling schedule accordingly. Most CL implementations rely on predefined or heuristic difficulty measures. Instead, more recent extensions, termed self-paced learning (SPL)~\cite{kumar2010self}, estimate sample difficulty using the training loss at each iteration. Yet, SPL tends to repeatedly select easy samples, since data points with lower losses are consistently prioritized. To overcome this limitation, Jiang \etal~\cite{jiang2015self} proposed self-paced curriculum learning (SPCL), which integrates prior knowledge from predefined curricula with the adaptive nature of SPL, enabling the model to leverage both task structure and feedback during learning. Building on this principle, our method employs an error-aware sampling strategy, where sample difficulty is inferred from the model's current error distribution across tasks and data subsets. Unlike prior work, our curriculum integrates both spatial and textual feedback to guide training in grounded medical vision-language models.



%% file: sec_3_methodology.tex
\section{Methodology}
\label{sec:methodology}

Our curriculum-guided multi-task training framework, CURE, enhances the capabilities of medical VLMs by reformulating the training methodology without requiring additional data. CURE restructures existing datasets into a unified, fine-grained instructional format. In addition, it introduces an error-aware curriculum that dynamically adjusts sampling based on the model's performance across datasets and anatomical categories. This reformulated training pipeline enables the model to focus on challenging samples and underperforming regions progressively.

\begin{table}[t]
\footnotesize
\centering
\caption{
    \textbf{Dataset composition and statistics.} Number of instances for each task across the training, validation, and test splits. The MIMIC-CXR dataset serves as a superset, providing the Chest ImaGenome (CIG) and MS-CXR subsets for training and its official test split for report-generation evaluation. Evaluation-only datasets are used to assess generalization performance. \textbf{AGRG} refers to Anatomy Grounded Report Generation, \textbf{PG} to Phrase Grounding, and \textbf{GRG} to Grounded Report Generation.
}
\label{tab:dataset_statistics}
\input{tables_dataset_statistics}

\end{table}

\subsection{Error-Aware Curriculum Learning}
\label{sec:CL}

Our training setup involves a collection of $k$ distinct data sources, $\mathcal{D} = \{D_1, D_2, \dots, D_k\}$. Each data source ($D_i$) corresponds to a specific dataset and its original supervision task, such as phrase grounding (PG) in MS-CXR or anatomy grounded report generation (AGRG) in Chest ImaGenome. Under standard multi-task training, samples are drawn from $\mathcal{D}$ in proportion to the dataset size ($|D_i|$), the sampling probability for $D_i$ is defined as: $p_i = |D_i| / |\mathcal{D}|$. 

However, medical imaging datasets are inherently imbalanced in terms of relative dataset sizes and in the distribution of anatomical or semantic classes within each dataset. The second source of imbalance is further emphasized due to variations in patient demographics, acquisition protocols, and the natural frequency of clinical conditions~\cite{sanchez2022cx}.

As shown in Table~\ref{tab:dataset_statistics}, our data composition reflects this imbalance: over 12.9M instances from Chest ImaGenome dominate the much smaller MS-CXR (815 PG instances) and PadChest-GR ($\sim$12k instances) datasets. Beyond the imbalance between datasets, each dataset also exhibits substantial intra-dataset class imbalance. For example, anatomical regions in AGRG and semantic categories in PG are unevenly represented, leading the model to overfit frequent regions, neglect rare but clinically important ones, and hallucinate findings (Figure~\ref{fig:framework}). Our curriculum framework addresses both sources of imbalance.

\begin{figure*}[t]
  \centering
  \includegraphics[width=\linewidth]{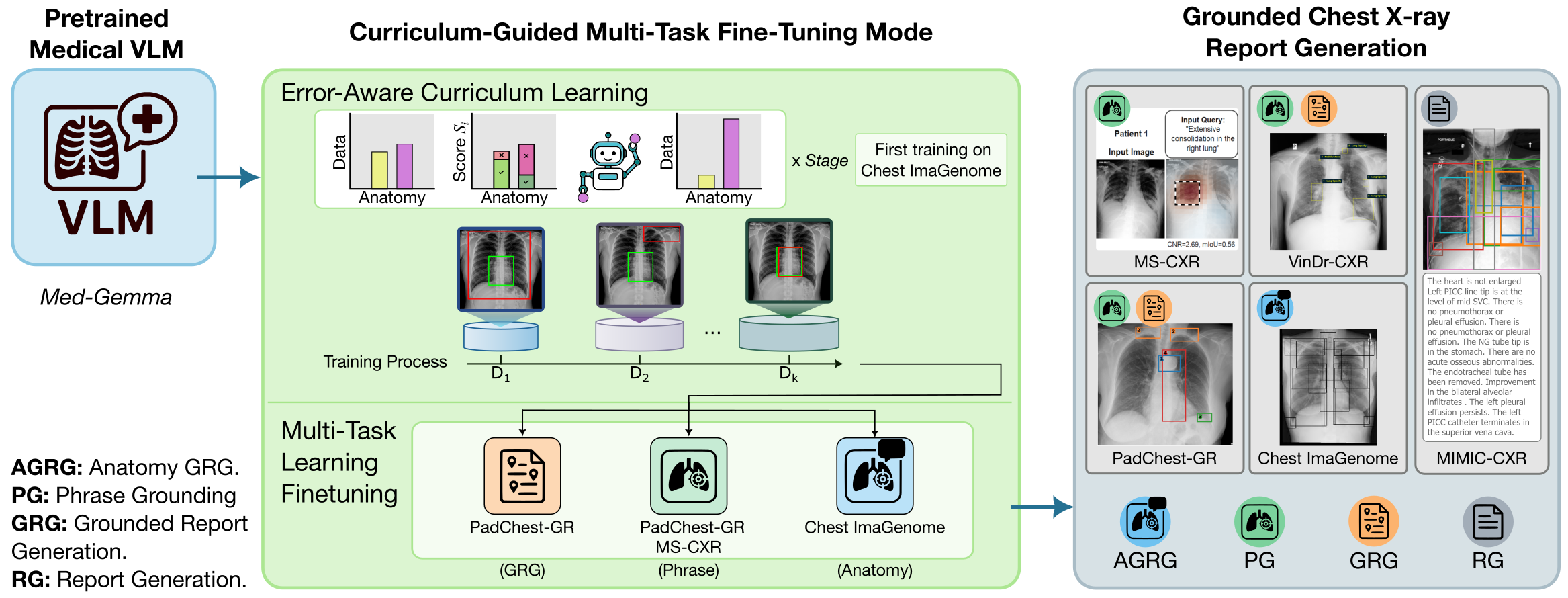}
  \caption{
    \textbf{Overview of CURE, our Curriculum-guided Multi-task Training Framework.} During training, the model is periodically evaluated every N steps on validation subsets from each task. Performance metrics (IoU, CXRFEScore) are calculated to identify task-level and category-level errors, which are then used to update the sampling weights in the training sampler. The cycle then resumes, allowing the model to focus more heavily on the data it finds most challenging. Evaluation of the RG task uses the official MIMIC-CXR test set, while VinDr-CXR is assessed in a zero-shot setting.
  }
  \label{fig:main_fig}
  \vspace{-0.4cm}
\end{figure*}

To mitigate these issues, we introduce an error-aware curriculum learning strategy~\cite{hacohen2019power} that dynamically adjusts sampling probabilities at two levels of granularity: (i) at the dataset level, where we re-weight the relative number of training samples per dataset, and (ii) at the class level, where we re-weight samples per anatomical region or semantic class. The curriculum proceeds over $n$ iterative stages. Each stage consists of three steps: training, evaluation, and sample re-weighting. The sampling probability for each data source ($p_i$) and class is modulated by the error rate ($e_i$) at the current stage. The sampling probabilities estimated at stage $n$ are used to initialize stage $n+1$, while the first stage starts with a uniform distribution \ie, $p_i = p_j$ for all $i,j$. The overall procedure is illustrated in Figure~\ref{fig:main_fig}.

\mysection{Data Source Evaluation.}
To maintain computational efficiency, each stage evaluates a random, fixed-size subset from each validation source (e.g., $\sim$200 samples for Chest ImaGenome, $\sim$150 for PadChest-GR, and $\sim$100 for MS-CXR). For each subset, we assess boxes localization quality using Intersection over Union (IoU) and textual fidelity using CXRFEScore~\cite{messina2024extracting}, which captures clinical semantic similarity. These metrics determine the updated sampling weights for the next stage.

\mysection{Inter-Dataset Curriculum.} After each evaluation stage, we compute an aggregate performance score ($s_i$) for each data source $D_i$ as a weighted sum of the two metrics:
\begin{equation}
s_i = \alpha \cdot \text{IoU}_i + (1 - \alpha) \cdot \text{CXRFEScore}_i,
\end{equation}
where $\alpha$ controls the trade-off between localization accuracy and semantic quality. The error for each source is defined as $e_i = 1 - s_i$. The sampling probability for the next training stage is then obtained by normalizing these errors:
\begin{equation}
p_i = \frac{e_i}{\sum_{j=1}^{K} e_j}.
\end{equation}

\mysection{Intra-Dataset Curriculum.}
Within each dataset, the curriculum operates at a finer granularity by re-weighting categories, anatomical regions, or semantic groups based on their per-class error.
For phrase grounding (PG) in MS-CXR~\cite{mscxr}, we group samples by the eight original phrase classes (e.g., pneumonia, consolidation). In PadChest-GR~\cite{padchest-gr}, where annotations span 155 labels organized into 26 higher-level label groups (e.g., atelectasis, cardiomegaly), we apply the curriculum at the group level.

The most fine-grained application of the curriculum occurs in Chest ImaGenome~\cite{chestImagenome} for the anatomy grounded report generation (AGRG) task. AGRG comprises three subtasks: \textit{Locate}, \textit{Describe}, and \textit{Locate and describe}, sampled uniformly to preserve task balance. The curriculum acts within each subtask by re-weighting anatomical locations according to their error. 

Because annotation coverage differs across subtasks, each one has its own set of available locations (i.e., 36 with bounding box for \textit{Locate}, 38 with text for \textit{Describe}, and 29 with both annotations for \textit{Locate and describe}). This formulation allows the model to focus on anatomical regions that are spatially or semantically challenging, improving both localization and descriptive quality.

For the grounded report generation (GRG) task in PadChest-GR, we do not apply intra-dataset re-weighting. These reports contain multiple co-occurring findings spanning several anatomical and pathological categories. Since there is no straightforward or well-defined way to categorize such multi-label instances for fine-grained rebalancing, we adopt uniform sampling for this task.

\subsection{Fine-Grained Task Formulation}
\label{sec:task_formulation}

CURE re-formulates diverse tasks into a unified, fine-grained instructional format. Each training instance from the source dataset $D_i$ is represented as a triplet \texttt{(image, instruction, response)}, allowing heterogeneous supervision (bounding boxes, phrases, anatomical labels, and descriptive sentences) to be learned under a consistent multi-task framework.



\mysection{Phrase Grounding (PG).} For MS-CXR and PadChest-GR, the original annotations link phrases within a report to one or more corresponding bounding boxes on the image. We transform these annotations into direct grounding instructions. For each annotated phrase, we create a sample using the template prompt \texttt{Ground the phrase: \{phrase\}}. When a phrase is associated with multiple bounding boxes, all of them are included in the response. The expected output is the phrase followed by one or more normalized bounding box coordinates: \texttt{{phrase}: [cx$_1$,cy$_1$,w$_1$,h$_1$]...[cx$_n$,cy$_n$,w$_n$,h$_n$]}.

PadChest-GR provides $8,489$ sentences and 155 unique, fine-grained clinical labels. To enrich our training data, we exploit this dual annotation by generating additional instances: for each \texttt{(sentence, box list)} pair, we create \texttt{(label, box list)} pairs. For example, from ``Minimal biapical pleural thickening'' with label ``apical pleural thickening'' and bounding box \texttt{[cx,cy,w,h]}, we produce one instance for the sentence and another for the label. This augmentation, applied only to training and validation splits, nearly doubles the PadChest-GR PG data and helps the model ground natural descriptions and canonical clinical terms. The test set remains un-augmented.



\mysection{Grounded Report Generation (GRG).}
In the PadChest-GR's GRG task, the model must produce a complete report in which grounded findings are explicitly linked to bounding boxes. We use the instruction: \texttt{Generate a grounded report}. The target response is the ground-truth report, where phrases corresponding to available bounding box annotations are augmented with their coordinates directly in the text. For example, an output may take the form: \texttt{\{phrase$_1$\} [cx$_1$,cy$_1$,w$_1$,h$_1$]. \{phrase$_2$\} [cx$_{2}$,cy$_{2}$,w$_{2}$,h$_{2}$]}, where each phrase is linked to its corresponding list of bounding boxes.




\mysection{Anatomy-Grounded Report Generation (AGRG).}
Chest ImaGenome provides detailed scene-graphs linking anatomical locations in MIMIC-CXR frontal-view images to both descriptive sentences and bounding boxes. This detailed structure is ideal for our AGRG fine-grained task decomposition as it allows for a significant expansion of the training data: a single image can yield multiple training instances, one for each annotated location. Concretely, the $\sim$237k images in the training split generate an average of 9 to 36 instructional instances per image, depending on the subtask, culminating in millions of viable training samples reported in Table~\ref{tab:dataset_statistics}. From this large pool, we create three distinct subtasks:

\begin{itemize}
    \item \textbf{Locate:} To train pure spatial localization, we use the prompt \texttt{Locate the \{location\}}. The expected output follows the format: \texttt{Location of the \{location\}: [cx,cy,w,h]}.

    \item \textbf{Describe:} To train contextual description independent of localization, we use the prompt \texttt{Describe the \{location\}}. The target output is formatted as: \texttt{Description of the \{location\}: \{description\}}.

    \item \textbf{Locate and Describe:} To train the model to perform both tasks simultaneously, we prompt: \texttt{Locate and describe the \{location\}}. The response combines both formats: \texttt{Location of the \{location\}: [cx,cy,w,h]. Description: \{description\}}.
\end{itemize}

This multi-prompt formulation explicitly disentangles and teaches the model the diverse tasks of localization and description. This process generates over 12.9M potential training instances. However, it is important to note that due to computational constraints, only a fraction of this pool is sampled during training, as detailed in the footnote of Table~\ref{tab:dataset_statistics}.

%% file: tables_dataset_statistics.tex
\footnotesize
\centering

\setlength{\tabcolsep}{1pt}

\begin{tabular}{llrrr}
\toprule
\textbf{Dataset} & \textbf{Task} & \textbf{\# Train} & \textbf{\# Val} & \textbf{\# Test} \\
\midrule
\multicolumn{5}{l}{\textit{\textbf{Training \& In-Domain Evaluation}}} \\
\multirow{3}{*}{MIMIC-CXR (CIG) } & AGRG \textit{Locate} & 8.5M\ddag & 69.9K\dag & \multirow{3}{*}{1,000*} \\
& AGRG \textit{Describe} & 2.3M\ddag & 18.8K\dag & \\
& AGRG \textit{Loc., Desc.} & 2.1M\ddag & 17.4K\dag & \\
\cmidrule(l){1-5}
MIMIC-CXR (MS-CXR) & PG & 815\ddag & 169\dag & 176 \\
\cmidrule(l){1-5}
MIMIC-CXR (Test Split) & Report Generation & — & — & 3,155 \\
\cmidrule(l){1-5}
\multirow{2}{*}{PadChest-GR} & GRG & 3,185\ddag & 455\dag & 915 \\
& PG & 8,871**\ddag & 1,297\dag** & 1,238 \\
\midrule
\multicolumn{5}{l}{\textit{\textbf{Zero-Shot Generalization}}} \\
\multirow{2}{*}{VinDr-CXR} & GRG & — & — & 3,000 \\
& PG & — & — & 2,108 \\
\bottomrule
\multicolumn{5}{p{0.47\textwidth}}{\footnotesize{
*A sampled subset of 1,000 from 123,319 total available instances.}} \\
\multicolumn{5}{p{0.47\textwidth}}{\footnotesize{
**Includes both original report phrases and fine-grained labels as phrases.}} \\
\multicolumn{5}{p{0.47\textwidth}}{\footnotesize{
\dag Each curriculum cycle evaluates a stratified random subset of the validation set.}} \\
\multicolumn{5}{p{0.47\textwidth}}{\footnotesize{
\ddag Due to computational constraints, only a fraction of the total available instances are used. With 9,000 training steps and an effective batch size of 25, a maximum of 225,000 instances ($\approx$1.74\% of the total) are processed.}} \\
\vspace{-0.6cm}
\end{tabular}

%% file: sec_4_experiments.tex
\section{Experiments}
\label{sec:experiments}

\begin{table}[t]
\footnotesize
\centering
\caption{
    \textbf{Results for Phrase Grounding (PG).} We report Micro-Average IoU (IoU Mi. \(\uparrow\)) and Macro-Average IoU (IoU Ma. \(\uparrow\)) on three test sets: MS-CXR, PadChest-GR, and zero-shot VinDr-CXR. CURE consistently improves localization performance across all metrics and datasets, including VinDr-CXR, which was not seen during training.
}
\label{tab:main_pg_results}

\input{tables_main_pg_results}
\vspace{-0.4cm}
\end{table}



We evaluate CURE on Anatomy-Grounded Report Generation (AGRG), Grounded Report Generation (GRG), Phrase Grounding (PG), and traditional report generation (RG). For RG, we repurpose the AGRG and GRG setups (see Supplementary Table~\ref{tab:task_summary}). We benchmark against state-of-the-art methods and ablate core components to assess their impact on grounding accuracy and report quality.

\mysection{Datasets.} CURE adopts a multi-task fine-tuning strategy on the same three publicly available chest X-ray datasets used by our baseline (MAIRA-2), enabling a fair and comparable evaluation.
Chest ImaGenome~\cite{chestImagenome} is a subset derived from the MIMIC-CXR~\cite{mimiccxr} dataset that focuses exclusively on frontal-view X-rays. It provides scene graphs linking anatomical regions to bounding boxes and their corresponding original MIMIC-CXR report sentences, and is used for the Anatomy-Grounded Report Generation task.
PadChest-GR~\cite{padchest-gr} contains radiology reports with phrases explicitly grounded in bounding boxes, supporting both Grounded Report Generation and Phrase Grounding tasks.
MS-CXR~\cite{mscxr} is a smaller dataset derived from MIMIC-CXR, providing bounding box annotations for report phrases, and is used exclusively for Phrase Grounding.
For zero-shot evaluation, we also include VinDr-CXR~\cite{vindrcxr}, using its official test split to assess the model's generalization to unseen data distributions.

\mysection{Evaluation Metrics.}
For the \textit{Grounded Report Generation} task, localization performance is measured using the mean Intersection over Union (IoU), while textual report quality is evaluated using CheXbert \cite{chexbert} metrics (such as Precision, Recall, F1-score, and Cosine Similarity), CXRFEScore \cite{messina-etal-2024-extracting}, RaTEScore~\cite{ratescore}, and RadGraph F1 \cite{delbrouck-etal-2024-radgraph}.
For the \textit{Phrase Grounding task}, we report both micro- and macro-averaged IoU to evaluate across and within categories.
All reported metrics include standard deviations derived from 1000 bootstrap samples.
Please refer to Supplementary Section~\ref{sec:suppl_datasets} for a detailed overview of the tasks, corresponding prompt formats, and expected output structures for each dataset used in our experiments.



We use MedGemma-4B-IT \cite{medgemma} as the base model, fine-tuned with LoRA \cite{lora} at rank 16 in 4-bit precision. All training stages use the AdamW optimizer \cite{adamw} with a consistent set of hyperparameters: an effective batch size of 25 (per-device batch size 5 with 5 gradient accumulation steps), a learning rate of \(2\times10^{-4}\), a linear scheduler with a 0.03 warmup ratio, and gradient clipping with maximum gradient norm set to 0.3. Optimizer and scheduler states reset between phases; the initial 3000-step pre-training and subsequent 6000-step multi-task phase use separate optimizer instances, preserving only model weights. Data augmentation includes spatial transformations and contrast-limited adaptive histogram equalization (CLAHE) \cite{clahe}.


\mysection{Baseline Methods.} We compare CURE against two strong baselines.
First, we evaluate MedGemma-4B-IT~\cite{medgemma} to establish the pretrained model’s baseline performance.
Second, we include MAIRA-2~\cite{bannur2024maira}, a state-of-the-art open-source medical vision–language model that jointly learns grounding and report generation tasks.

\begin{table}[b]
\vspace{-0.2cm}
\centering
\caption{
    \textbf{Results for Anatomy-Grounded Report Generation (AGRG) on Chest ImaGenome (CIG).}
    We report mean IoU (\(\uparrow\)), CheXbert F1 (micro/macro) (\(\uparrow\)), 
    CheXbert cosine similarity (\(\uparrow\)), and CXRFEScore (\(\uparrow\)).
    \textbf{Bold} values indicate the best performance for each metric.
}
\label{tab:main_agrg_results}

\input{tables_main_agrg_results}
\end{table}

\begin{table*}[t]
\centering
\footnotesize
\caption{
    \textbf{Results for Grounded Report Generation (GRG) on PadChest-GR and zero-shot VinDr-CXR.}
    We report mean IoU (\(\uparrow\)), CheXbert F1 (micro/macro) (\(\uparrow\)), 
    CheXbert cosine similarity (\(\uparrow\)), and CXRFEScore (\(\uparrow\)).
    Bold values indicate the best score for each metric.
}
\label{tab:main_grg_results}

\input{tables_main_grg_results}
\vspace{-0.2cm}
\end{table*}

\subsection{Comparison against State-of-the-Art}

\mysection{Visual Grounding.} Table~\ref{tab:main_pg_results} shows that CURE consistently outperforms MAIRA-2 across all metrics (Micro and Macro IoU) and datasets, including MS-CXR and PadChest-GR.  CURE achieves larger relative gains on PadChest-GR than on MS-CXR, highlighting the effectiveness of the proposed error-aware reweighting strategy.

\mysection{Anatomy-Grounded Report Generation.}
On Chest ImaGenome, CURE gains +0.35 IoU over MAIRA-2, doubling localization performance (Table~\ref{tab:antGro}). Since MedGemma-4B-IT lacks innate visual grounding, this confirms our pipeline successfully instills it. Furthermore, CURE surpasses MAIRA-2's IoU on GRG (Table~\ref{tab:main_grg_results}), despite MAIRA-2's pretraining on the proprietary USMix dataset~\cite{maira2} (193K text-only, 69K grounded reports).

\mysection{Report Generation Performance.} We evaluate CURE across three report generation tasks: on the MIMIC-CXR test set for standard report generation, and on Chest-ImaGenome and PadChest-GR for the AGRG and GRG tasks, respectively. Table~\ref{tab:main_agrg_results} presents AGRG results on a 1000 sample subset of the Chest ImaGenome test set (see Table~\ref{tab:dataset_statistics}). While our model's most significant improvement is in spatial grounding, it also shows enhanced performance in most text-based metrics for this fine-grained, in-domain task. CURE achieves the highest scores on F1-Mi (0.529), cosine similarity (0.691), and CXRFEScore (0.549), outperforming the baseline models.

Table~\ref{tab:main_grg_results} reports GRG performance on PadChest-GR. This task naturally favors MAIRA-2, owing to its additional training on the proprietary USMix dataset, which allows it to surpass CURE on most text-based metrics. In contrast, CURE consistently delivers higher grounding accuracy, achieving the best mean IoU (0.265). Although CURE does not outperform MAIRA-2 on text-based metrics, it substantially boosts MedGemma-4B-IT’s performance in these metrics. This highlights the effectiveness of our training paradigm, particularly given that PadChest-GR (GRG) contains far fewer training instances than its version for PG and Chest ImaGenome for AGRG.

Table~\ref{tab:main_mimiccxr_report_gen_results} shows performance on the MIMIC-CXR test set, comprising 3,155 frontal views. For fair comparison, all models generate reports from the same frontal view. Our CURE (AGRG) variant, which generates reports by concatenating the descriptions for 29 distinct anatomical locations, outperforms all the baselines in recall scores (R-Ma: 0.539, R-Mi: 0.749), and the combined CURE (AGRG+GRG) further improves this (R-Ma: 0.582, R-Mi: 0.781), demonstrating that our fine-grained approach effectively captures a wide range of findings. The top-performing model on most metrics is CXRMate-RRG24~\cite{cxrmate-rrg24}, the winner of a recent radiology report generation competition. Its leading RadGraph F1 score is consistent with its training, which was optimized via reinforcement learning using RadGraph F1 as a reward. Comparatively, our model remains competitive on semantic similarity metrics, achieving the highest CheXbert Cosine Similarity (0.792), RaTEScore (0.597), and a close second on CXRFEScore (0.655). CURE also achieves the highest CheXbert F1 macro (0.415).

\begin{table*}[ht]
\centering
\footnotesize
\caption{
    \textbf{Results for Report Generation (RG) on the MIMIC-CXR test set.}
    We report CheXbert F1 (F1-Ma/Mi) (\(\uparrow\)), Precision (P-Ma/Mi) (\(\uparrow\)), and Recall (R-Ma/Mi) (\(\uparrow\)), each macro (Ma) and micro (Mi) averaged together with CheXbert Cosine Similarity (Cos.) (\(\uparrow\)), CXRFEScore (CXRFES) (\(\uparrow\)), RaTEScore (RaTES) (\(\uparrow\)), and RadGraph F1 (RadF1) (\(\uparrow\)).
    \textbf{Bold} and \underline{underlined} values indicate the best and second-best models per metric, respectively.
}
\label{tab:main_mimiccxr_report_gen_results}

\input{tables_main_mimiccxr_report_gen_results}
\vspace{-0.3cm}
\end{table*}

\mysection{Evaluating Hallucination and Reliability.}
To complement our empirical evaluation, we conduct a targeted analysis of hallucination and report consistency on the AGRG task. The evaluation is performed on a subset of the Chest ImaGenome test set, comprising 300 examples for six key anatomical locations (1800 images in total). We prompted CURE using its \texttt{Locate and describe} instruction and prompted MAIRA-2 to ground and describe the same location. We then employed Gemini 2.5 Flash Lite to perform a Natural Language Inference (NLI) comparison between each model's generated description for a specific anatomy and the full ground-truth MIMIC-CXR report.

As detailed in Table~\ref{tab:cig_hallucination_results}, CURE substantially improves reliability. On average, it reduces abnormal finding hallucinations from MAIRA-2's 26.50\% to 8.78\%, halves contradictions (17.44\% vs. 33.22\%), and more than doubles entailment (39.50\% vs. 15.94\%).
The improvement is particularly stark for bone structures like the clavicles, where CURE's hallucination rate is only 1.00\% compared to MAIRA-2's rates of over 59\%.

While MAIRA-2 exhibits a slightly lower hallucination rate for a few anatomies (e.g., `Cardiac Silhouette'), CURE consistently achieves a significantly lower contradiction rate and higher entailment rate across almost all categories. This improved performance likely stems from a key difference in the training data composition. Standard phrase grounding, used by MAIRA-2, is inherently biased towards abnormal findings. In contrast, our AGRG formulation exposes the model to both normal and abnormal descriptions for each anatomical region, leading to a more balanced and reliable generative process that mitigates the tendency to hallucinate abnormalities.

\begin{table}[b]
\vspace{-0.3cm}
\centering
\caption{
    \textbf{Hallucination analysis for AGRG on a Chest ImaGenome subset.}
    We report abnormal finding hallucination rates (\%) and assess report consistency using a Natural Language Inference (NLI) framework, reporting the fraction (\%) of Contradiction (Cont.) and Entailment (Entail). Lower hallucination and contradiction rates, together with higher entailment rates, indicate better clinical faithfulness and grounding quality.
}
\label{tab:cig_hallucination_results}

\input{tables_cig_hallucination_results}
\end{table}

\begin{table*}[t]
\scriptsize
\centering
\caption{\textbf{Ablation Study.}
We evaluate the contribution of each component across three grounding tasks. CXRS denotes the CXRFEScore metric. For Phrase Grounding (PG), we report Micro-Averaged IoU on MS-CXR (MS), PadChest-GR (PC), and VinDr-CXR (VD). Note that: CL(\textit{f}) indicates curriculum learning with a reweighting frequency of \textit{f} steps, CIG(\textit{s}) denotes a Chest ImaGenome pre-training stage of \textit{s} steps, HPS refers to hyperparameter search. \textbf{Bold} and \underline{underlined} values indicate the best and second-best models per metric, respectively.
}
\label{tab:ablation_study_detailed}

\input{tables_main_ablations}
\end{table*}

\mysection{Zero-Shot Performance.} To assess the generalization capabilities of CURE beyond the datasets seen during training, we further evaluate the model on the VinDr-CXR test split for both PG and GRG (see Tables~\ref{tab:main_pg_results} and \ref{tab:main_grg_results}). As shown in Table~\ref{tab:main_pg_results}, CURE surpasses MAIRA-2 in grounding performance on the PG task, achieving higher Micro and Macro IoU. Similarly, on the GRG task, CURE equips MedGemma-4B-IT with a strong grounding ability, outperforming MAIRA-2 in IoU and narrowing the gap in text-based metrics, despite MAIRA-2's advantage from training on proprietary GRG-specific data. These results demonstrate that the capabilities learned through CURE generalize effectively to out-of-domain datasets and task formats.

\subsection{Ablation Study}

To assess the contribution of each component in our training pipeline, we conduct a detailed ablation study summarized in Table~\ref{tab:ablation_study_detailed}. For context, we include MAIRA-2 as a strong external baseline. Our analysis begins with a multi-task fine-tuned baseline (\texttt{v1}), trained without data augmentation, curriculum learning, or specialized pre-training.


\begin{figure*}[h!]
    \centering
    \setlength{\tabcolsep}{2pt}
    \begin{tabular}{cccc}
      \multicolumn{2}{c}{\textbf{``Nodule or mass'' (VinDr-CXR)}} & \multicolumn{2}{c}{\textbf{``Surgical staples'' (PadChest-GR)}} \\[2pt]

      \textbf{MAIRA-2} & \textbf{CURE} & \textbf{MAIRA-2} & \textbf{CURE} \\[2pt]

      \includegraphics[height=3.9cm]{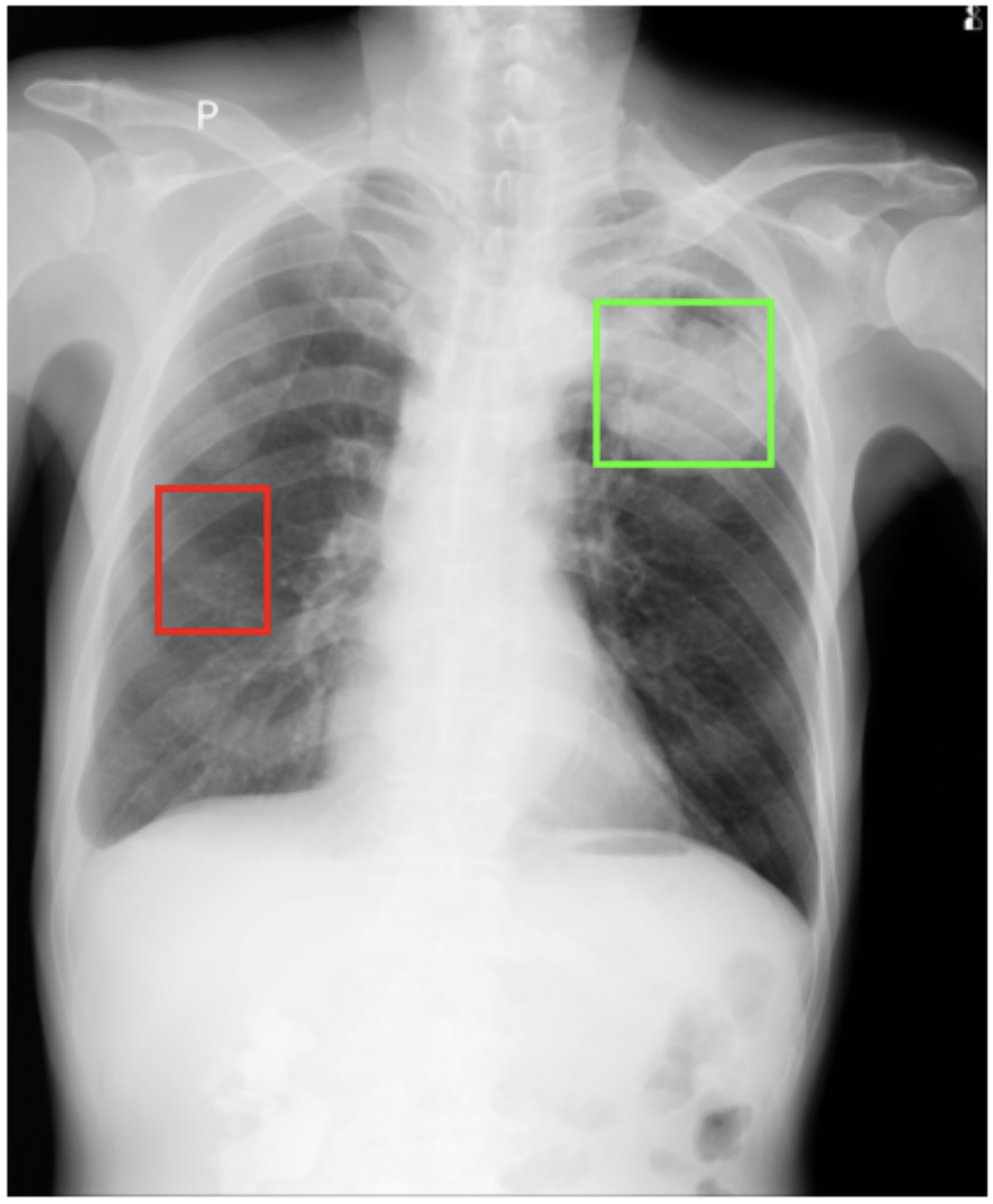} &
      \includegraphics[height=3.9cm]{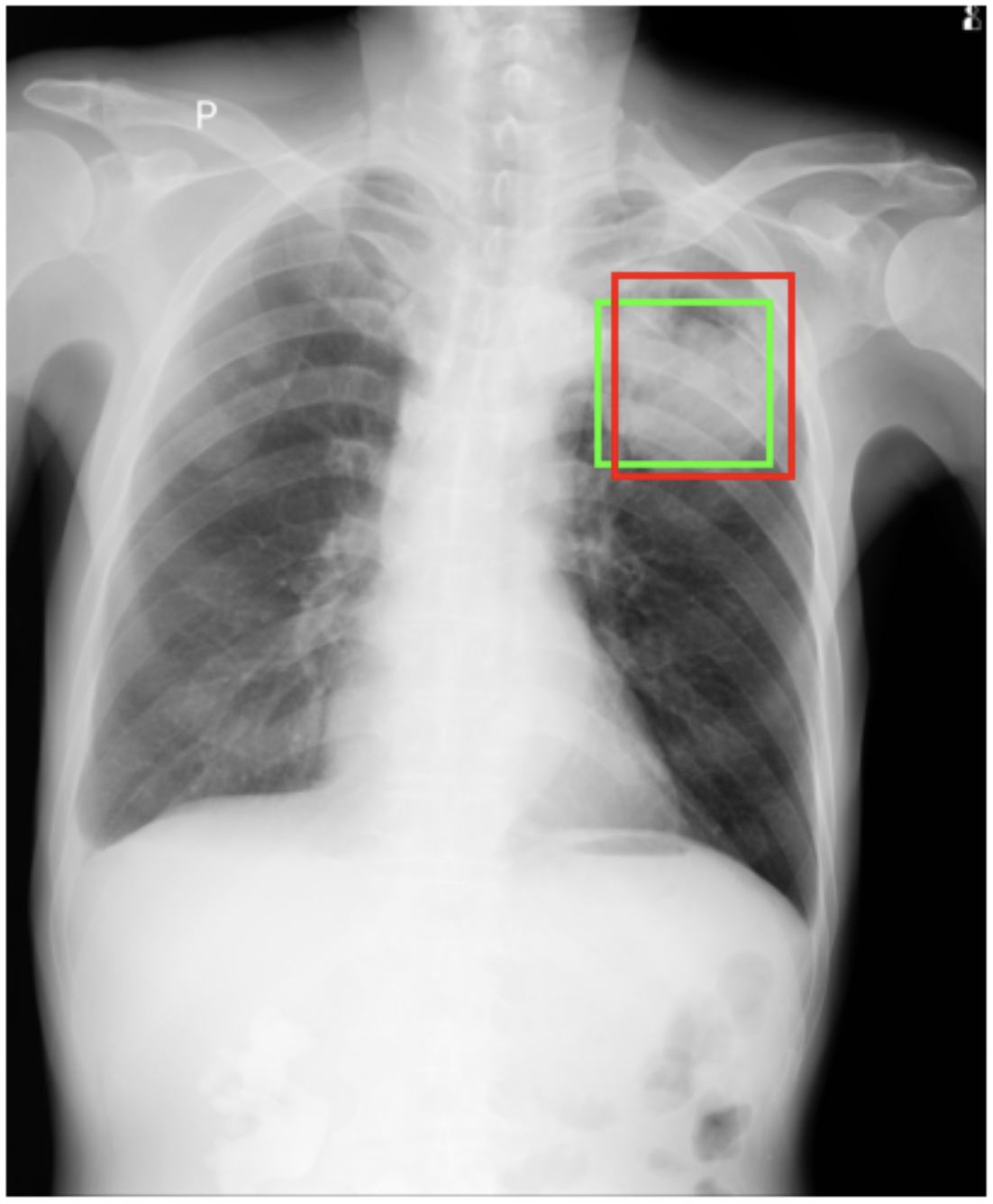} &
      \includegraphics[height=3.9cm]{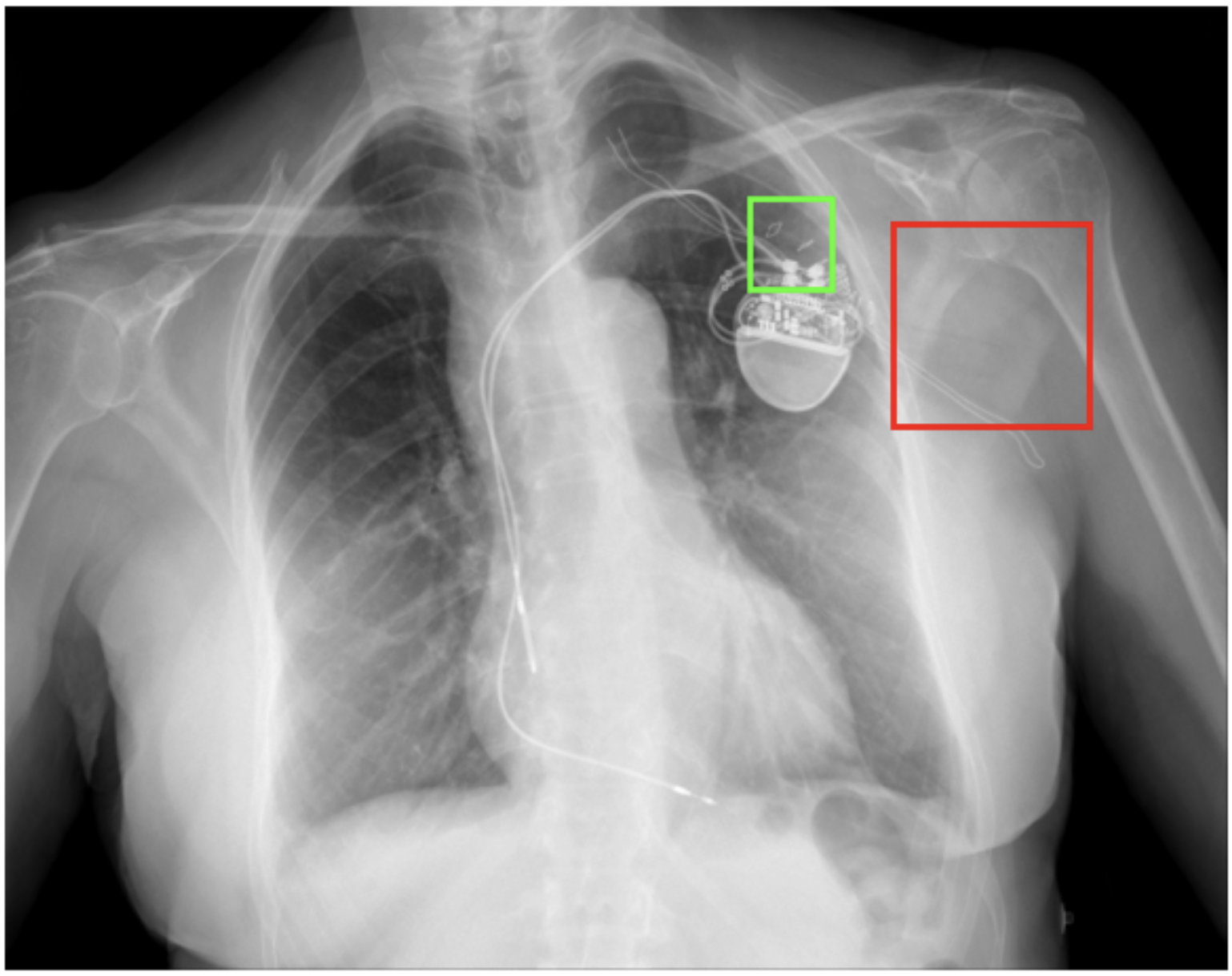} &
      \includegraphics[height=3.9cm]{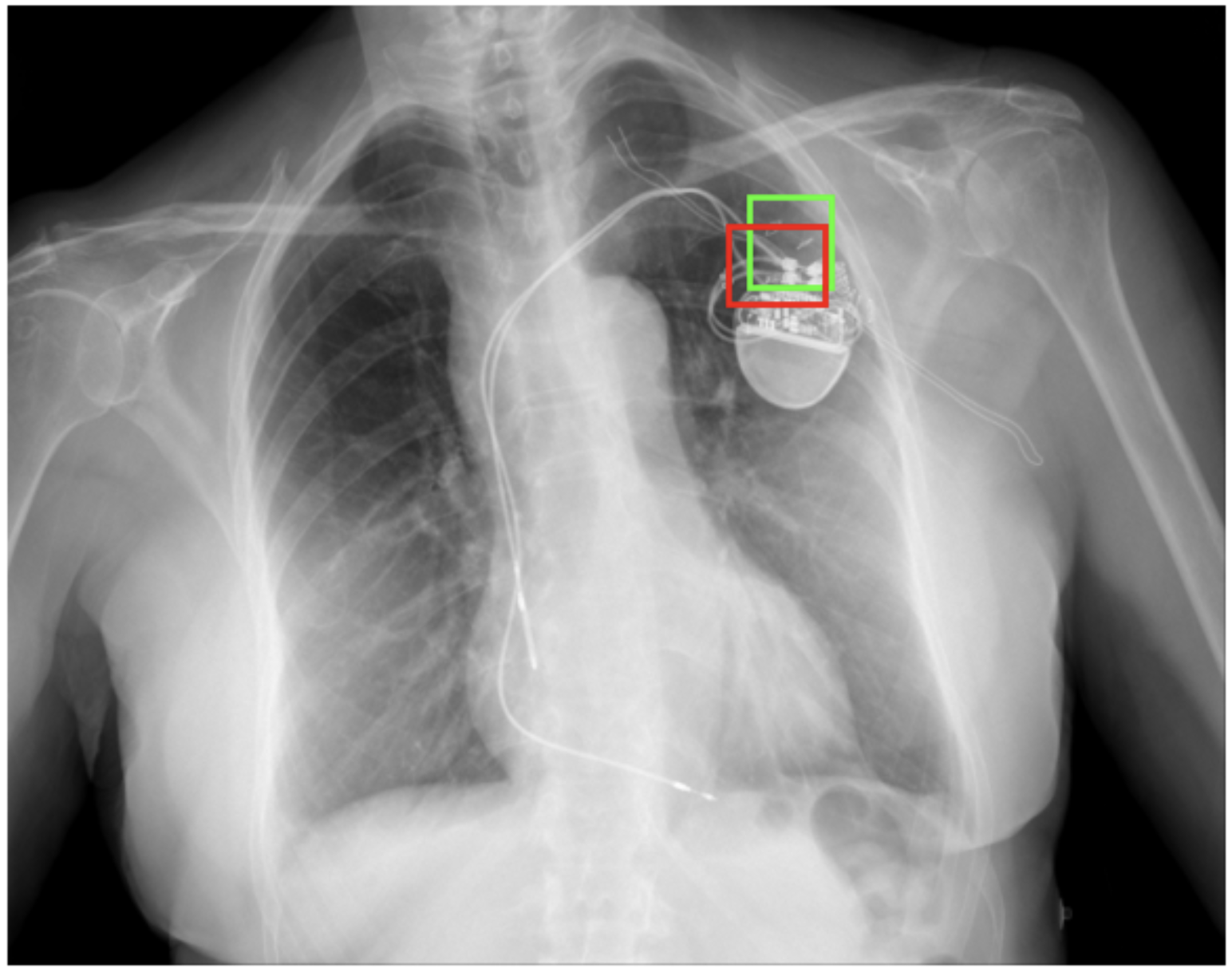}
    \end{tabular}
    \caption{\textbf{Qualitative Examples.} Qualitative phrase grounding (PG) results on challenging examples from the VinDr-CXR and PadChest-GR datasets. The left panels show the detection of a \textit{``Nodule or mass''} (VinDr-CXR), while the right panels demonstrate the grounding of \textit{``Surgical staples''} (PadChest-GR). Ground-truth regions are shown in green for reference, and model predictions from \textbf{MAIRA-2} and \textbf{CURE} are shown in red.}
    \label{fig:qualitative}
    \vspace{-0.3cm}
\end{figure*}

Even this simplest variant already surpasses MAIRA-2 on the AGRG task and on the zero-shot VinDr-CXR PG task, highlighting the effectiveness of our fine-grained task reformulation. However, \texttt{v1} underperforms on GRG IoU and on the MS-CXR PG task. Adding \textbf{data augmentation} (\texttt{v2}) yields small but consistent improvements in PG performance across all three grounding datasets.

We then examine the effect of \textbf{curriculum learning} by varying the frequency of our performance-based reweighting strategy (\texttt{v3-v5}). A reweighting interval of 3000 steps (\texttt{v5}) tends to produce better performance, outperforming more frequent updates. We therefore adopt this interval in all subsequent experiments.

A larger gain emerges when introducing a dedicated \textbf{Chest ImaGenome pre-training stage} (\texttt{v6-v8}). Increasing the pre-training duration to 3000 steps (\texttt{v8}) enables the model to match MAIRA-2 on the MS-CXR PG task (0.495), emphasizing the value of grounding-specific pre-training before the curriculum stage. 

Finally, we perform a hyperparameter search (HPS) for both the pre-training stage and the multi-task fine-tuning phase (\texttt{v9}). Our final model, \textbf{CURE (\texttt{v9})}, achieves the best overall performance, notably surpassing MAIRA-2 on the challenging GRG (PC) IoU benchmark (0.265 vs. 0.256), while maintaining strong results across all other tasks.

\subsection{Qualitative Analysis}
Figure~\ref{fig:qualitative} compares the phrase grounding performance of CURE and MAIRA-2 on challenging examples from VinDr-CXR~\cite{vindrcxr} and PadChest-GR~\cite{padchest-gr}. 
In the left panel (VinDr-CXR), MAIRA-2 fails to localize a \textit{``Nodule or mass''}, incorrectly placing a bounding box in the lower right lung field. CURE, conversely, accurately localizes the true nodule in the upper left lung, closely aligning with the ground truth. Similarly, in the right panel (PadChest-GR) for \textit{``Surgical staples''}, MAIRA-2's prediction falls outside the relevant anatomical region, whereas CURE successfully bounds the staples and implanted device. 
These qualitative examples are consistent with CURE's overall higher phrase grounding metrics on both datasets, illustrating its improved spatial localization and semantic alignment compared to the baseline.

%% file: tables_main_pg_results.tex
\resizebox{0.48\textwidth}{!}{
\setlength{\tabcolsep}{2pt} 
\begin{tabular}{@{}lcccccc@{}}
\toprule
\small
\multirow{2}{*}{\textbf{Model}} &
\multicolumn{2}{c}{\textbf{MS-CXR}} &
\multicolumn{2}{c}{\textbf{PadChest-GR}} &
\multicolumn{2}{c}{\textbf{VinDr-CXR (Zero-Shot)}} \\ 
\cmidrule(l){2-3} \cmidrule(l){4-5} \cmidrule(l){6-7}
& IoU Mi. $\uparrow$ & IoU Ma. $\uparrow$ & IoU Mi. $\uparrow$ & IoU Ma. $\uparrow$ & IoU Mi. $\uparrow$ & IoU Ma. $\uparrow$\\ 
\midrule
MAIRA-2 & 0.495 $\pm$ 0.016 & 0.453 $\pm$ 0.016 & 0.280 $\pm$ 0.008 & 0.288 $\pm$ 0.009 & 0.161 $\pm$ 0.005 & 0.114 $\pm$ 0.010 \\
\midrule


CURE & \textbf{0.552} $\pm$ 0.015 & \textbf{0.495} $\pm$ 0.015 & \textbf{0.453} $\pm$ 0.006 & \textbf{0.438} $\pm$ 0.007 & \textbf{0.243} $\pm$ 0.005 & \textbf{0.205} $\pm$ 0.012 \\
\bottomrule
\label{tab:phrase_gr}
\end{tabular}%
}

%% file: tables_main_agrg_results.tex
\resizebox{0.48\textwidth}{!}{%
\setlength{\tabcolsep}{2pt} 
\begin{tabular}{@{}lccccc@{}}
\toprule
\textbf{Model} &
IoU $\uparrow$ &
F1-Mi $\uparrow$ &
F1-Ma $\uparrow$ &
Cos. $\uparrow$ &
CXRFEScore $\uparrow$ \\ 
\midrule
MAIRA-2 & 0.249 $\pm$ 0.008 & 0.377 $\pm$ 0.016 & 0.098 $\pm$ 0.009 & 0.587 $\pm$ 0.010 & 0.357 $\pm$ 0.010 \\
MedGemma-4B-IT & -- & 0.266 $\pm$ 0.012 & 0.227 $\pm$ 0.014 & 0.662 $\pm$ 0.004 & 0.467 $\pm$ 0.006 \\
\midrule
CURE & \textbf{0.601} $\pm$ 0.008 & \textbf{0.529} $\pm$ 0.017 & \textbf{0.234} $\pm$ 0.018 & \textbf{0.691} $\pm$ 0.009 & \textbf{0.549} $\pm$ 0.011 \\
\bottomrule
\label{tab:antGro}
\end{tabular}%
}

%% file: tables_main_grg_results.tex
\resizebox{1.0\textwidth}{!}{%
\setlength{\tabcolsep}{2pt} 
\begin{tabular}{@{}lccccccccccc@{}}
\toprule
\multicolumn{1}{c}{\multirow{2}{*}{\textbf{Model}}} &
\multicolumn{5}{c}{\textbf{PadChest-GR}} &
\multicolumn{5}{c}{\textbf{VinDr-CXR (Zero-Shot)}} \\
\cmidrule(l){2-6} \cmidrule(l){7-11}
& IoU $\uparrow$ & F1-Mi $\uparrow$ & F1-Ma $\uparrow$ & Cos. $\uparrow$ & CXRFEScore $\uparrow$ &
  IoU $\uparrow$ & F1-Mi $\uparrow$ & F1-Ma $\uparrow$ & Cos. $\uparrow$ & CXRFEScore $\uparrow$ \\
\midrule
MAIRA-2 & 0.256 $\pm$ 0.011 & \textbf{0.591} $\pm$ 0.015 & \textbf{0.321} $\pm$ 0.019 & \textbf{0.844} $\pm$ 0.004 & \textbf{0.616} $\pm$ 0.011 & 0.217 $\pm$ 0.007 & \textbf{0.546} $\pm$ 0.008 & \textbf{0.256} $\pm$ 0.011 & 0.824 $\pm$ 0.002 & 0.591 $\pm$ 0.005 \\
MedGemma-4B-IT & -- & 0.144 $\pm$ 0.009 & 0.203 $\pm$ 0.014 & 0.733 $\pm$ 0.003 & 0.517 $\pm$ 0.005 & -- & 0.209 $\pm$ 0.006 & 0.212 $\pm$ 0.008 & 0.779 $\pm$ 0.001 & \textbf{0.596} $\pm$ 0.003 \\
\midrule
CURE & \textbf{0.265} $\pm$ 0.011 & 0.507 $\pm$ 0.015 & 0.270 $\pm$ 0.013 & 0.819 $\pm$ 0.005 & 0.574 $\pm$ 0.010 & \textbf{0.262} $\pm$ 0.007 & 0.505 $\pm$ 0.008 & 0.246 $\pm$ 0.009 & \textbf{0.832} $\pm$ 0.003 & 0.540 $\pm$ 0.007 \\
\bottomrule
\end{tabular}%
\vspace{-0.4cm}
}

%% file: tables_main_mimiccxr_report_gen_results.tex
\resizebox{1.0\textwidth}{!}{%
\setlength{\tabcolsep}{2pt} 
\begin{tabular}{@{}lccccccccccc@{}}
\toprule
\textbf{Model} &
\textbf{F1-Ma} $\uparrow$ & \textbf{F1-Mi} $\uparrow$ &
\textbf{P-Ma} $\uparrow$ & \textbf{P-Mi} $\uparrow$ &
\textbf{R-Ma} $\uparrow$ & \textbf{R-Mi} $\uparrow$ &
\textbf{Cos.} $\uparrow$ & \textbf{CXRFES} $\uparrow$ & \textbf{RaTES} $\uparrow$ & \textbf{RadF1} $\uparrow$\\
\midrule
CXRMate-RRG24 & \underline{0.414} $\pm$ 0.006 & \textbf{0.589} $\pm$ 0.004 & \textbf{0.493} $\pm$ 0.012 & \underline{0.617} $\pm$ 0.005 & 0.415 $\pm$ 0.006 & 0.563 $\pm$ 0.005 & 0.764 $\pm$ 0.001 & \textbf{0.656} $\pm$ 0.002 & 0.577 $\pm$ 0.002 & \textbf{0.255} $\pm$ 0.002 \\
MAIRA-2 (w/ grounding) & 0.304 $\pm$ 0.006 & 0.489 $\pm$ 0.005 & \underline{0.442} $\pm$ 0.021 & \textbf{0.639} $\pm$ 0.006 & 0.283 $\pm$ 0.006 & 0.397 $\pm$ 0.005 & 0.751 $\pm$ 0.002 & 0.603 $\pm$ 0.002 & 0.496 $\pm$ 0.002 & 0.120 $\pm$ 0.002 \\
MAIRA-2 (w/o grounding) & 0.386 $\pm$ 0.006 & 0.554 $\pm$ 0.004 & 0.425 $\pm$ 0.009 & 0.578 $\pm$ 0.005 & 0.384 $\pm$ 0.006 & 0.533 $\pm$ 0.005 & 0.693 $\pm$ 0.002 & 0.576 $\pm$ 0.002 & 0.501 $\pm$ 0.002 & 0.143 $\pm$ 0.002 \\
MedGemma-4B-IT & 0.382 $\pm$ 0.004 & 0.547 $\pm$ 0.004 & 0.332 $\pm$ 0.005 & 0.452 $\pm$ 0.004 & 0.494 $\pm$ 0.005 & 0.692 $\pm$ 0.005 & 0.714 $\pm$ 0.001 & 0.580 $\pm$ 0.002 & 0.532 $\pm$ 0.001 & 0.112 $\pm$ 0.001 \\
\midrule
CURE (GRG) & 0.314 $\pm$ 0.006 & 0.463 $\pm$ 0.005 & \underline{0.442} $\pm$ 0.009 & 0.605 $\pm$ 0.006 & 0.290 $\pm$ 0.006 & 0.376 $\pm$ 0.004 & 0.725 $\pm$ 0.002 & 0.526 $\pm$ 0.002 & 0.447 $\pm$ 0.002 & 0.077 $\pm$ 0.002 \\
CURE (AGRG) & 0.400 $\pm$ 0.005 & 0.559 $\pm$ 0.004 & 0.355 $\pm$ 0.008 & 0.446 $\pm$ 0.004 & \underline{0.539} $\pm$ 0.005 & \underline{0.749} $\pm$ 0.004 & \underline{0.783} $\pm$ 0.001 & 0.645 $\pm$ 0.002 & \underline{0.592} $\pm$ 0.001 & \underline{0.181} $\pm$ 0.001 \\
CURE (AGRG+GRG) & \textbf{0.415} $\pm$ 0.005 & \underline{0.562} $\pm$ 0.004 & 0.365 $\pm$ 0.010 & 0.439 $\pm$ 0.004 & \textbf{0.582} $\pm$ 0.005 & \textbf{0.781} $\pm$ 0.004 & \textbf{0.792} $\pm$ 0.001 & \underline{0.655} $\pm$ 0.002 & \textbf{0.597} $\pm$ 0.001 & 0.176 $\pm$ 0.001 \\
\bottomrule
\end{tabular}%
\vspace{-0.4cm}
}

%% file: tables_cig_hallucination_results.tex
\resizebox{0.46\textwidth}{!}{%
\begin{tabular}{@{}llrrr@{}}
\toprule
\textbf{Model} & \textbf{Anatomy} &
\textbf{Abn. } $\downarrow$ &
\textbf{Cont. } $\downarrow$ &
\textbf{Entail. } $\uparrow$ \\
\midrule
MAIRA-2 & Cardiac Silhouette & \bf 2.00 & \bf 8.00 & 25.00 \\
CURE & Cardiac Silhouette & 25.67 & 27.67 & \bf 47.33 \\
\midrule
MAIRA-2 & Left Clavicle & 59.00 & 22.67 & 5.00 \\
CURE & Left Clavicle & \bf 1.00 & \bf 7.00 & \bf 32.67 \\
\midrule
MAIRA-2 & Left Lung & 12.00 & 56.33 & 21.67 \\
CURE & Left Lung & \bf 7.00 & \bf 32.33 & \bf 41.67 \\
\midrule
MAIRA-2 & Right Clavicle & 62.67 & 20.33 & 1.67 \\
CURE & Right Clavicle & \bf 1.00 & \bf 7.33 & \bf 27.67  \\
\midrule
MAIRA-2 & Right Lung & \bf 10.67 & 53.67 & 29.33 \\
CURE & Right Lung & 11.67 & \bf 27.00 & \bf 46.00 \\
\midrule
MAIRA-2 & Spine & 12.67 & 38.33 & 13.00 \\
CURE & Spine & \bf 6.33 & \bf 3.33 & \bf 41.67 \\
\midrule
MAIRA-2 & \textit{Mean Anatomies} & 26.50 & 33.22 & 15.94 \\
CURE & \textit{Mean Anatomies} & \bf 8.78 & \bf 17.44 & \bf 39.50 \\
\bottomrule
\end{tabular}%
}

%% file: tables_main_ablations.tex
\resizebox{\textwidth}{!}{%
\setlength{\tabcolsep}{2pt} 
\begin{tabular}{@{}l cc cc cc ccc@{}}
\toprule
\multicolumn{1}{c}{\multirow{2}{*}{\textbf{Model Configuration}}} & \multicolumn{2}{c}{\textbf{AGRG (CIG)}} & \multicolumn{2}{c}{\textbf{GRG (PC)}} & \multicolumn{2}{c}{\textbf{GRG (VD)}} & \multicolumn{3}{c}{\textbf{PG (IoU Mi. $\uparrow$)}} \\
\cmidrule(l){2-3} \cmidrule(l){4-5} \cmidrule(l){6-7} \cmidrule(l){8-10}
& IoU $\uparrow$ & CXRS $\uparrow$ & IoU $\uparrow$ & CXRS $\uparrow$ & IoU $\uparrow$ & CXRS $\uparrow$ & MS & PC & VD \\ 
\midrule
MAIRA-2 (External Baseline) & 0.249 $\pm$ 0.008 & 0.357 $\pm$ 0.010 & \underline{0.256} $\pm$ 0.011 & \textbf{0.616} $\pm$ 0.011 & 0.217 $\pm$ 0.007 & 0.591 $\pm$ 0.005 & \underline{0.495} $\pm$ 0.016 & 0.280 $\pm$ 0.008 & 0.161 $\pm$ 0.005 \\
\midrule
v1: Base (w/o Aug, w/o CL, w/o CIG) & 0.380 $\pm$ 0.008 & 0.517 $\pm$ 0.011 & 0.171 $\pm$ 0.009 & 0.589 $\pm$ 0.011 & 0.207 $\pm$ 0.006 & 0.630 $\pm$ 0.007 & 0.388 $\pm$ 0.017 & 0.356 $\pm$ 0.007 & 0.191 $\pm$ 0.004 \\
v2: + Aug & 0.360 $\pm$ 0.009 & 0.522 $\pm$ 0.011 & 0.185 $\pm$ 0.010 & \underline{0.599} $\pm$ 0.011 & 0.221 $\pm$ 0.007 & 0.648 $\pm$ 0.007 & 0.398 $\pm$ 0.019 & 0.366 $\pm$ 0.007 & 0.203 $\pm$ 0.005 \\
\midrule
v3: + Aug + CL(1.5k) & 0.399 $\pm$ 0.009 & 0.521 $\pm$ 0.011 & 0.179 $\pm$ 0.010 & 0.592 $\pm$ 0.011 & 0.224 $\pm$ 0.007 & 0.630 $\pm$ 0.007 & 0.409 $\pm$ 0.019 & 0.383 $\pm$ 0.007 & 0.210 $\pm$ 0.005 \\
v4: + Aug + CL(2k) & 0.394 $\pm$ 0.009 & 0.513 $\pm$ 0.011 & 0.193 $\pm$ 0.010 & 0.596 $\pm$ 0.011 & 0.222 $\pm$ 0.006 & \underline{0.651} $\pm$ 0.007 & 0.393 $\pm$ 0.017 & 0.383 $\pm$ 0.007 & 0.196 $\pm$ 0.005 \\
v5: + Aug + CL(3k) & 0.411 $\pm$ 0.009 & 0.526 $\pm$ 0.011 & 0.180 $\pm$ 0.010 & 0.595 $\pm$ 0.012 & 0.217 $\pm$ 0.007 & \textbf{0.671} $\pm$ 0.007 & 0.430 $\pm$ 0.018 & 0.393 $\pm$ 0.007 & 0.205 $\pm$ 0.005 \\
\midrule
v6: + Aug + CIG(1k) + CL(3k) & 0.454 $\pm$ 0.008 & 0.518 $\pm$ 0.011 & 0.195 $\pm$ 0.009 & 0.591 $\pm$ 0.011 & 0.232 $\pm$ 0.006 & 0.628 $\pm$ 0.007 & 0.457 $\pm$ 0.016 & 0.394 $\pm$ 0.007 & 0.219 $\pm$ 0.005 \\
v7: + Aug + CIG(2k) + CL(3k) & 0.448 $\pm$ 0.008 & \underline{0.533} $\pm$ 0.011 & 0.203 $\pm$ 0.010 & 0.586 $\pm$ 0.011 & 0.227 $\pm$ 0.006 & 0.590 $\pm$ 0.007 & 0.467 $\pm$ 0.016 & 0.403 $\pm$ 0.007 & 0.222 $\pm$ 0.005 \\
v8: + Aug + CIG(3k) + CL(3k) & \underline{0.486} $\pm$ 0.008 & 0.521 $\pm$ 0.011 & 0.207 $\pm$ 0.010 & 0.582 $\pm$ 0.011 & \underline{0.233} $\pm$ 0.006 & 0.601 $\pm$ 0.007 & \underline{0.495} $\pm$ 0.016 & \underline{0.421} $\pm$ 0.007 & \underline{0.224} $\pm$ 0.005 \\
\midrule
\textbf{v9 (CURE)}: + Aug + CIG(3k) + CL(3k) + HPS & \textbf{0.601} $\pm$ 0.008 & \textbf{0.549} $\pm$ 0.011 & \textbf{0.265} $\pm$ 0.011 & 0.574 $\pm$ 0.010 & \textbf{0.262} $\pm$ 0.007 & 0.540 $\pm$ 0.007 & \textbf{0.552} $\pm$ 0.015 & \textbf{0.453} $\pm$ 0.006 & \textbf{0.243} $\pm$ 0.005 \\
\bottomrule
\end{tabular}%
}

%% file: sec_5_conclusion.tex
\section{Conclusion}
\label{sec:conclusion}

\vspace{-0.2cm}
We introduced CURE, an error-aware curriculum learning framework that advances visual grounding and factual reliability in medical vision–language models through adaptive multi-task training on existing public datasets. By dynamically prioritizing underperforming samples and categories, CURE delivers consistent improvements over the baseline across diverse tasks, including phrase grounding, where it outperforms in all Micro and Macro IoU metrics on datasets such as MS-CXR, PadChest-GR, and VinDr-CXR. Moreover, CURE enables effective grounding in the base MedGemma-4B-IT model— which originally lacks visual grounding capabilities—achieving a +0.35 IoU improvement over MAIRA-2 in anatomy-grounded report generation, more than doubling localization accuracy.
Finally, CURE reduces hallucinations by 18.6\% across key anatomical regions, halves contradictions, and doubles entailment, demonstrating measurable progress in clinical faithfulness for medical VLMs.


%% file: sec_6_suppl.tex
\clearpage
\maketitlesupplementary

\section{Implementation and Training Details}
\label{sec:suppl_implementation}

This section provides a detailed overview of the experimental setup, including hardware, hyperparameters, and the curriculum learning configuration to ensure full reproducibility.

\subsection{Model and Hardware Setup}
Our framework is implemented in PyTorch \cite{paszke2019pytorch} using the Hugging Face ecosystem \cite{wolf2019huggingface}, particularly the \texttt{SFTTrainer} from the TRL library for supervised fine-tuning. All experiments were conducted within a SLURM-managed High-Performance Computing (HPC) cluster equipped with NVIDIA RTX A6000 GPUs, each providing 48 GB of VRAM.

Each training run for our final model was executed on a single GPU with 60 GB of system memory RAM. The total training time for the final 9000-step CURE model was approximately 45 hours, comprising a 15-hour pre-training stage (3000 steps) followed by a 30-hour multi-task fine-tuning stage (6000 steps).

\subsection{Hyperparameter Details}
The complete training process for the final version of CURE spans 9000 steps and is strictly divided into two main phases. For clarity, the chronological pipeline is structured as follows:
\begin{itemize}
    \item \textbf{Phase 1: Pre-training (3000 steps).} The model is trained exclusively on the Chest ImaGenome dataset.
    \item \textbf{Phase 2: Multi-task Fine-tuning (6000 steps).} The optimizer and scheduler states are reset, preserving only the model weights from Phase 1. This phase applies our curriculum learning framework and is further split into two stages:
    \begin{itemize}
        \item \textit{Stage 2a: Warm-up (3000 steps).} Uniform sampling is applied across all datasets and tasks.
        \item \textit{Stage 2b: Cyclic Re-weighting (3000 steps).} A performance evaluation dictates new sampling weights, which are then fixed for these concluding steps.
    \end{itemize}
\end{itemize}
Key hyperparameters, which remained consistent across all stages unless otherwise noted, are detailed in Table~\ref{tab:suppl_hyperparams}.

\begin{table}[h!]
\centering
\caption{\textbf{Hyperparameter Configuration.} Detailed hyperparameters for the pre-training and multi-task fine-tuning stages.}
\label{tab:suppl_hyperparams}
\input{tables_hyperparameters}
\end{table}





\subsection{Curriculum Learning Details}
\label{sec:suppl_currlearn_details}

Our curriculum learning framework is applied during the multi-task fine-tuning phase (Phase 2) to dynamically prioritize underperforming tasks. The protocol operates via two conceptual mechanisms:
\begin{enumerate}
    \item \textbf{Warm-up (Stage 2a):} By sampling all datasets and intra-dataset categories uniformly, the model receives balanced exposure to all tasks. This establishes a stable performance baseline before adaptation begins.
    \item \textbf{Error-Aware Re-weighting (Stage 2b):} The model's performance is evaluated on validation sets to recalculate sampling weights, forcing the network to focus on its most frequent errors. 
\end{enumerate}
While our framework supports continuous cyclic re-weighting (e.g., recalculating weights every $M$ steps), our ablation studies (Table~\ref{tab:ablation_study_detailed}) showed that a single weight update—calculated immediately after the warm-up and fixed for the remainder of training—yielded the most effective and stable configuration for our final \textbf{CURE} model.

The re-weighting mechanism itself operates at two levels of granularity:



\mysection{Curriculum Scoring \& Re-weighting.}
Our curriculum operates at two levels: \textbf{inter-dataset} (across different data sources) and \textbf{intra-dataset} (across fine-grained categories, such as the 8 phrase classes in MS-CXR). For both levels, we compute an aggregate performance score $s$ using a task-adaptive metric:
\begin{equation}
s = \alpha \cdot \text{IoU} + (1 - \alpha) \cdot \text{CXRFEScore}.
\end{equation}
The weight $\alpha$ adapts to the subtask requirements: $\alpha=0$ for text-only generation (e.g., AGRG ``Describe''), $\alpha=1$ for pure localization (e.g., AGRG ``Locate''), and $\alpha=0.8$ when both modalities are evaluated. The resulting error $e = 1 - s$ updates the sampling probabilities at both the dataset and category levels, directing the model toward its most challenging concepts.

\begin{figure}[h!]
  \centering
  \includegraphics[width=\linewidth]{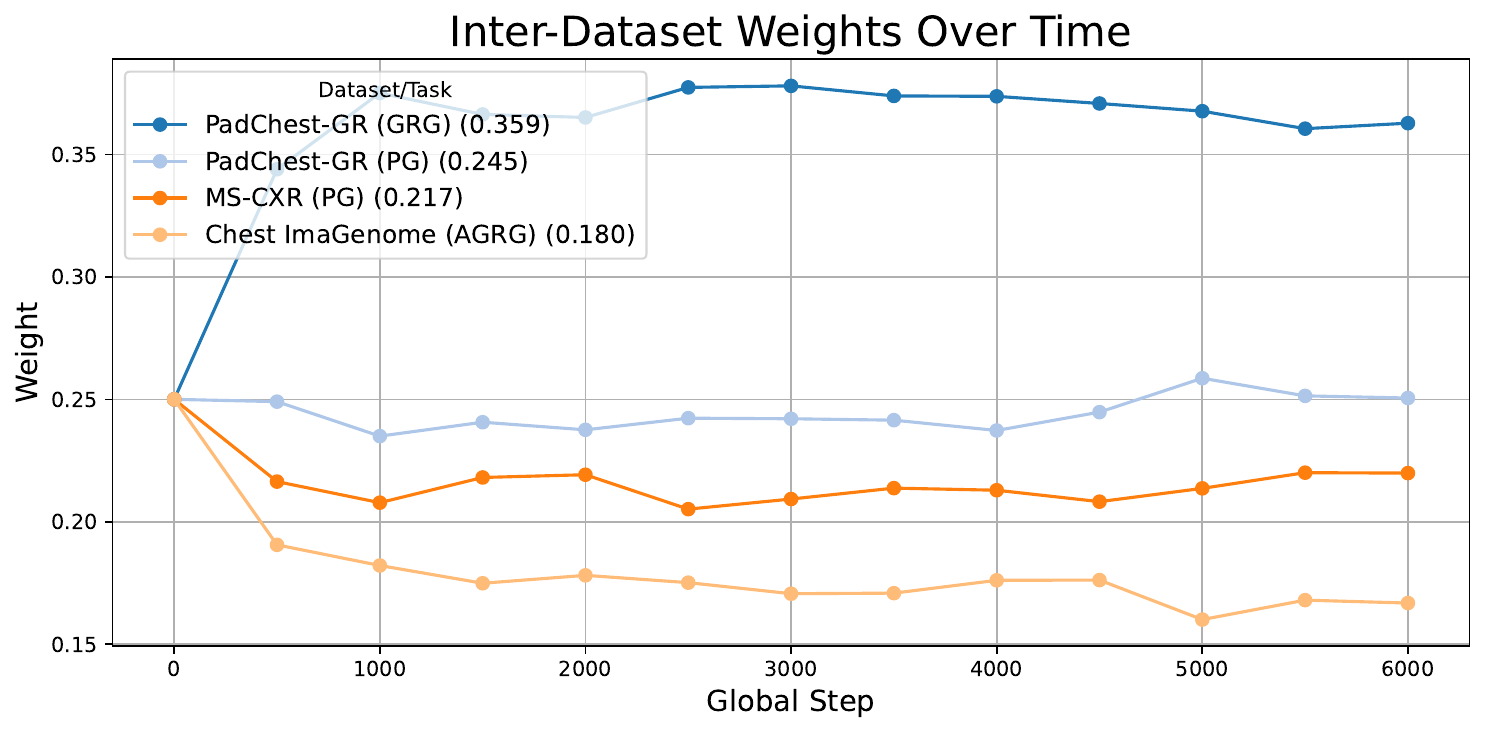}
  \caption{
    \textbf{Visualization of Inter-Dataset Weight Dynamics.} This plot illustrates the curriculum's adaptation from an experiment with frequent updates (every 500 steps). It shows how sampling probabilities for each data source evolve over time in response to the model's performance.
  }
  \label{fig:inter_weights}
\end{figure}

\begin{figure}[h!]
  \centering
  \includegraphics[width=\linewidth]{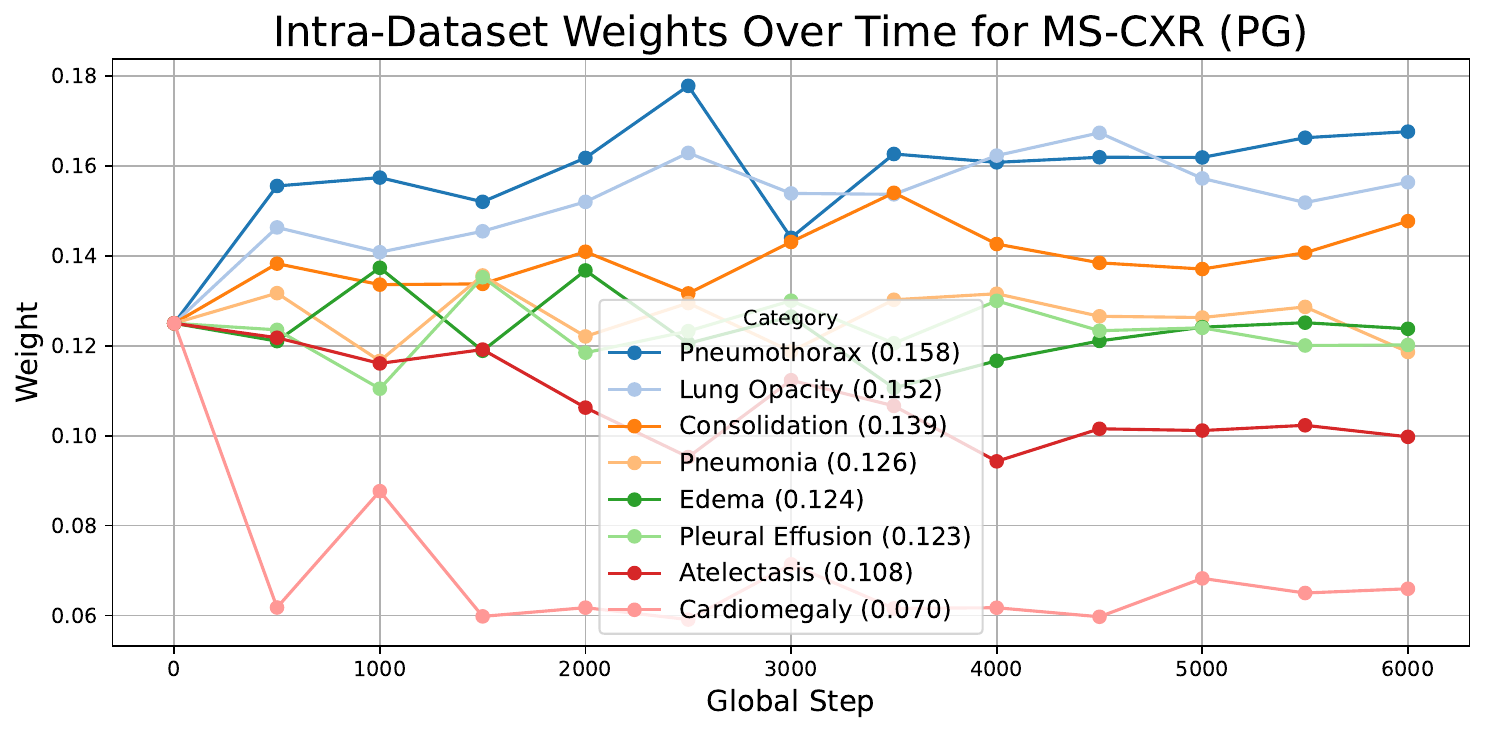}
  \caption{
    \textbf{Visualization of Intra-Dataset Weight Dynamics for MS-CXR.} This plot shows the category-level weight evolution for the 8 phrase classes in MS-CXR, taken from the same experiment with updates every 500 steps. The weights are periodically adjusted to prioritize classes with higher error rates.
  }
  \label{fig:intra_weights}
\end{figure}

To provide a more granular visualization of the curriculum's adaptive mechanism, Figures~\ref{fig:inter_weights} and \ref{fig:intra_weights} illustrate the weight evolution from an experiment with a more frequent re-weighting schedule (every 500 steps). While this specific timing differs from our final CURE model, these plots clearly demonstrate the dynamic nature of the framework in action.

\section{Datasets and Task Formulation}
\label{sec:suppl_datasets}

\subsection{Detailed Task I/O Formats}
Table \ref{tab:task_summary} provides representative examples of the instructional prompts and expected output formats for each dataset and task used during training and evaluation.

\subsection{Dataset Preprocessing and Splits}
We used the official train, validation, and test splits provided by each dataset. All images were processed using a custom pipeline built with the Albumentations library \cite{buslaev2020albumentations} and resized to a final resolution of \(448 \times 448\) pixels. The specific image transformations varied between the training and validation/test phases to ensure data diversity during training and deterministic evaluation.

\mysection{Training Pipeline.}
The training pipeline is stochastic, designed to improve model robustness to variations in X-ray acquisition. For each training image, the following transformations are applied:
\begin{itemize}
    \item \textbf{CLAHE:} Applied with a 50\% probability to simulate varying contrast levels, using a random clip limit uniformly sampled between 1.0 and 4.0, and a fixed tile grid size of \((8, 8)\).
    \item \textbf{Spatial Augmentations:}
    Spatial transformations included random resized cropping (30\% probability) and affine transformations (translation, scaling up to \(\pm10\%\), and rotation up to \(\pm15^\circ\)) applied with a 50\% probability. Horizontal flipping was disabled due to the inherent left-right asymmetry of thoracic anatomy. Color-based augmentations such as jitter or Gaussian noise were explicitly disabled.
    \item \textbf{Regularization:} To improve stability, 30\% of the training samples bypassed the spatial augmentations and instead used the deterministic validation pipeline described below.
\end{itemize}

Crucially, the entire training pipeline is bounding box-aware. When spatial transformations are applied, Albumentations simultaneously transforms the corresponding bounding box coordinates. To preserve alignment between the visual and textual modalities, the ground-truth text supervision provided to the MedGemma model is dynamically updated to reflect the augmented coordinates before training. This ensures that every augmented image remains correctly paired with its corresponding, spatially consistent text supervision.

\mysection{Validation and Test Pipeline.}
The validation and test pipelines are deterministic. Unlike the stochastic training pipeline, these splits utilized Contrast-Limited Adaptive Histogram Equalization (CLAHE) as a fixed preprocessing normalization step rather than an augmentation. Given the high dynamic range and variable exposure settings inherent to Chest X-rays, applying deterministic CLAHE (clip limit 3.0, tile grid size \((8, 8)\)) standardizes the local contrast distribution across all evaluation samples. This ensures that fine-grained clinical features—which are often obscured in low-contrast regions—are enhanced consistently for the visual encoder during inference. Finally, images were resized to \(448 \times 448\) pixels.


\begin{table*}[h]
\centering
\caption{
    \textbf{Summary of Datasets, Tasks, and I/O Formats.}
    PG = Phrase Grounding; GRG = Grounded Report Generation; AGRG = Anatomy-Grounded Report Generation; RG = Report Generation.  
    The table illustrates representative prompts and output formats for each dataset.  
    Bounding boxes are denoted as \texttt{[center\_x, center\_y, width, height]}.  
    For MIMIC-CXR (evaluation only), reports are generated using the GRG, AGRG, or hybrid AGRG+GRG approaches, and subsequently post-processed to remove bounding box coordinates prior to text-based evaluation.
}
\label{tab:task_summary}
\input{tables_task_summary}
\end{table*}


\subsection{Chest ImaGenome Evaluation Benchmark}
\label{sec:suppl_cig_subset}

We strictly adhered to the official MIMIC-CXR data splits. As detailed in the main text, the Chest ImaGenome dataset provides scene graphs for frontal-view images in MIMIC-CXR, linking anatomical bounding boxes to textual descriptions. While we utilized the original Chest ImaGenome annotations (text snippets and bounding boxes) directly for the large-scale training and validation of the Anatomy-Grounded Report Generation (AGRG) task, we devised a rigorous protocol for the test phase to ensure both computational feasibility and high-quality metric calculation.

\paragraph{Computational Constraints and Subsampling.}
Extracting all valid (image, location) pairs from the scene graphs of the official MIMIC-CXR test set yields a large pool of approximately 123,000 evaluation instances derived from 3,403 unique frontal-view images. While a comprehensive evaluation on the full set is theoretically ideal, it presents significant pragmatic challenges due to the inference latency of large multimodal models.

For example, generating anatomy-grounded reports for 1000 image-location pairs using a fine-tuned MedGemma-4B-IT takes approximately 1 hour and 20 minutes on a single NVIDIA RTX A6000 GPU. Extrapolating this to the full test set results in roughly 164 hours (nearly a week) of continuous inference time for a single model evaluation. To facilitate faster experimentation without sacrificing statistical rigor, we curated a representative, stratified subset of 1000 samples. This subset served as a fixed artifact, ensuring that all models in our experiments were evaluated on the same diverse set of examples.

\paragraph{Generating High-Quality Textual Ground Truth for Evaluation.}
For the test subset, we sought to improve the granularity and quality of the textual ground truth. The original Chest ImaGenome dataset employs a pre-LLM NLP pipeline to associate radiology report snippets with anatomical locations. While sufficient for large-scale training, these snippets can be somewhat noisy, as they consist of raw fragments from the original radiology report and may include details not strictly related to the specified anatomical location. Moreover, they were not generated using modern language models capable of producing concise, location-tailored phrasing.

To establish a robust reference standard for evaluation metrics, we maintained the original ground-truth bounding boxes but enhanced the textual annotations. We utilized \texttt{gemini-2.5-flash-lite} to synthesize concise, location-specific ``mini-reports'' derived directly from the complete original radiology reports. This ensures the model is evaluated against coherent, radiologist-style descriptions that are explicitly relevant to each anatomical region of interest. The Gemini-generated mini-reports were used only as evaluation ground truth, never for training. The prompt used for this refinement is detailed below:

\begin{lstlisting}[style=promptstyle]
You will be provided with a chest x-ray report and a specified anatomical location. Your task is to generate a JSON object in the following format: {"reasoning": "", "mini-report": ""}

Guidelines:

- reasoning: Begin your reasoning by identifying and naming anatomical regions in close proximity to the specified location. Then, briefly summarize the report as a sequence of findings/observations. Lastly, identify all findings relevant to the specified location. A finding or observation is relevant if it meets any of the following criteria: (1) it explicitly describes the specified anatomical location; (2) it explicitly describes a region anatomically very close to the specified location, where the description is highly likely to also apply to the specified location; (3) it makes a general description from which it logically and with absolute certainty follows that the description applies to the specified location as a specific instance (e.g., "both lungs are clear" implies "the right lung is clear"; "no bone abnormalities" implies "the right clavicle presents no abnormalities"); or (4) it describes devices, tubes, or other objects traversing or situated within the specified anatomical location. Present your reasoning as a single, continuous paragraph, strictly avoiding newlines and special characters.
- mini-report: From the relevant information identified in your reasoning, synthesize a concise and accurate mini-report, written in a style consistent with a radiologist's findings, specifically detailing the findings related to the specified anatomical location.
- If the report contains no findings or descriptions pertinent to the specified anatomical location, set the value of "mini-report" to "N/A".
- Make sure to use JSON format as shown above.
\end{lstlisting}

\paragraph{Stratified Sampling Strategy.}
The full test pool consists of 35,042 image-location pairs that contain descriptive findings and a larger set of image-location pairs annotated only with bounding boxes (i.e., normal or unmentioned regions). To construct the 1000-sample benchmark, we selected 700 instances with descriptive findings and 300 without.

To ensure the subset was representative of the broader test distribution, we applied a stratified sampling strategy. We generated structured annotations for the candidate mini-reports using \texttt{gemini-2.5-flash-lite} to label the presence of abnormalities and medical devices:

\begin{lstlisting}[style=promptstyle]
You will be provided with a chest X-ray report or sentence. Your task is to analyze the text and determine:

1. Whether any abnormalities or pathologies are mentioned.
2. Whether any medical devices or foreign objects are mentioned.

Output format:
Return a JSON object with the following fields:

{
  "reason": "A brief explanation of your reasoning.",
  "mentions_abnormalities": "yes" | "no",
  "mentions_devices": "yes" | "no"
}
\end{lstlisting}

Using these labels, the 700-sample partition was balanced across anatomical locations, abnormality status, and the presence of medical devices. The 300-sample partition (without specific findings) was sampled uniformly across anatomical locations to preserve anatomical diversity. This procedure yields a balanced evaluation benchmark derived strictly from the official test split.

\subsection{VinDr-CXR for Zero-Shot Generalization}
To assess model robustness against domain shifts and unseen data distributions, we employ the VinDr-CXR dataset~\cite{vindrcxr} as a zero-shot benchmark.

\paragraph{Dataset Characteristics.}
VinDr-CXR consists of 15,000 training and 3000 testing frontal-view Chest X-rays. Each image was annotated by a consensus of three radiologists for the presence of 28 common thoracic diseases and findings. These findings are categorized into 22 localizable classes (annotated with bounding boxes) and 6 global classes (image-level labels only).

\paragraph{Zero-Shot Protocol.}
We exclude the VinDr-CXR training set entirely. None of the models evaluated in this work (including CURE and all baselines) were trained or fine-tuned on any portion of VinDr-CXR. Consequently, all results reported on this dataset reflect pure zero-shot transfer capabilities.

\paragraph{Task Adaptation.}
Since VinDr-CXR provides structured classification and detection labels rather than narrative radiology reports, we adapted the annotations to align with our text-based generation tasks:
\begin{itemize}
    \item \textbf{Phrase Grounding (PG):} We mapped the short class labels (e.g., ``ILD'', ``Enlarged PA'') to full natural language phrases (e.g., ``Interstitial lung disease'', ``Enlarged pulmonary artery''). We generated evaluation instances for every localizable finding present in the test set, resulting in 2,108 zero-shot phrase grounding queries.
    \item \textbf{Grounded Report Generation (GRG):} To create reference targets for report generation, we synthesized ``pseudo-reports'' from the structured annotations. For a given image, we aggregated all positive findings; localizable findings were converted into text strings containing the finding name followed by their bounding box coordinates (e.g., ``Atelectasis [cx, cy, w, h]''), while global findings were appended as text-only sentences. These phrases were concatenated to form a complete, grounded target sequence, allowing us to compute both textual overlap and localization metrics.
\end{itemize}

\section{Evaluation Protocol}
\label{sec:suppl_eval}

\subsection{Metric Calculation}
All metrics were computed using publicly available official implementations to ensure reproducibility.

\paragraph{CheXbert Metrics.}
We used the official CheXbert implementation \cite{chexbert}, available at \url{https://pypi.org/project/f1chexbert/}, to compute clinical correctness metrics. Specifically, we report precision, recall, and F1 scores for all 14 labels under both micro and macro averaging schemes. In addition, we leveraged the BERT encoder within CheXbert to obtain dense embeddings for textual similarity analysis. Each report—both ground-truth and generated—was first segmented into individual sentences using a sentence tokenizer. The BERT model was then used to encode each sentence into an embedding vector, and we computed the cosine similarity between corresponding sentences to estimate semantic alignment. The final similarity score for a report pair was obtained by averaging these sentence-level cosine similarities, following a procedure conceptually similar to that used in CXRFEScore \cite{messina-etal-2024-extracting}.

\paragraph{RadGraph F1.}
RadGraph-based factual consistency was evaluated using the official \texttt{radgraph} library \cite{delbrouck-etal-2024-radgraph}, accessible at \url{https://pypi.org/project/radgraph/}. We adopted the recommended RG\_ER reward as the RadGraph F1 metric, which jointly measures overlap in entity and relation predictions between generated and reference reports.

\paragraph{CXRFEScore.}
We further computed CXRFEScore \cite{messina-etal-2024-extracting} to assess semantic and factual consistency via structured medical knowledge representations. CXRFEScore combines two components: a fact extractor and a fact encoder. We employed the publicly released CXRFEScore models (fact extractor and fact encoder) provided by the original authors.
The extracted facts from both generated and ground-truth reports were encoded and compared in the resulting embedding space to produce the final factual consistency score.

\paragraph{RaTEScore.}
To assess entity-aware radiology text similarity, we utilized RaTEScore \cite{ratescore}, available at \url{https://pypi.org/project/RaTEScore/}. Unlike standard lexical metrics, RaTEScore emphasizes crucial medical entities, such as diagnostic outcomes and anatomical details, and is designed to be robust against complex medical synonyms while remaining sensitive to negation expressions. We employed the default pipeline, which utilizes a fine-tuned DeBERTa model for Medical Entity Recognition (NER) and BioLORD-2023-C for synonym disambiguation, to compute the alignment between generated and reference reports.

\paragraph{Bounding Box Metrics.}
For visual grounding evaluation, Intersection-over-Union (IoU) was computed using standard bounding box evaluation scripts. In cases where either the ground truth or the model output contained multiple bounding boxes for a given region or entity, we first merged all ground-truth boxes into a single region and likewise merged all predicted boxes, then computed IoU between the two resulting union regions. This avoids ambiguity when datasets provide multiple overlapping annotations. We report the mean IoU value across all evaluated samples (micro-average). Additionally, we calculate the average IoU per class and report the mean of these class-wise averages (macro-average).

\section{Detailed Experimental Results}
\label{sec:suppl_results}

This section provides the complete, unabridged results from our experiments, including the full ablation study and per-task performance tables.

\begin{table*}[h!]
\scriptsize
\centering
\caption{
    \textbf{Experimental Configuration Summary.}
    Detailed definitions of the model configurations (v1–v15) evaluated in the ablation study (see Table~\ref{tab:suppl_ablations}).
    The table outlines the progression from the baseline model to the proposed CURE method, detailing variations in data augmentation (Aug), curriculum learning (CL) schedules, Chest ImaGenome (CIG) pre-training duration, learning rates, and sampling strategies (Inter/Intra-dataset).
}
\label{tab:suppl_configs}

\input{tables_suppl_configs}
\end{table*}

\begin{table*}[t]
\scriptsize
\centering
\caption{
\textbf{Extended Ablation Study.}
Performance of all ablation variants across three tasks: \textbf{AGRG} (Anatomy-Grounded Report Generation on Chest ImaGenome),
\textbf{GRG} (Grounded Report Generation on PadChest-GR and VinDr-CXR), and \textbf{PG} (Phrase Grounding on MS-CXR, PadChest-GR, and VinDr-CXR).
Each block reports mean Intersection-over-Union (\textbf{IoU}, micro average), CheXbert F1 (micro average, \textbf{F1}), and CXRFEScore (\textbf{CXS}) metrics.
Rows v1–v5 analyze the effects of data augmentation and curriculum learning (CL);
v6–v8 add CIG pre-training with a low learning rate (2e–5);
v9–v11 repeat those with a higher learning rate (2e–4);
and v12–v15 further explore inter- and intra-dataset sampling strategies.
Best results in each column are shown in \textbf{bold}, and second-best are \underline{underlined}.
\\[2pt]
\textit{Dataset abbreviations:} CIG = Chest ImaGenome, PC = PadChest-GR, VD = VinDr-CXR, MS = MS-CXR.
}
\label{tab:suppl_ablations}

\input{tables_suppl_ablations}
\end{table*}

\subsection{Extended Ablation Study}
\label{sec:extended_ablation}

Table \ref{tab:suppl_configs} provides detailed definitions for the model versions evaluated in our ablation study, and Table \ref{tab:suppl_ablations} presents comprehensive quantitative results. This section offers a step-by-step analysis of the training dynamics that led to the final CURE method. We examine the progression in four stages: the impact of baseline augmentations (v1–v2), the optimization of curriculum update schedules (v3–v5), the interaction between pre-training and learning rate scaling (v6–v11), and finally, a detailed isolation of sampling strategy effects (v12–v15).

\paragraph{Baseline and Data Augmentation (v1--v2).} Our baseline model (\textbf{v1}) employs uniform sampling without augmentation. As observed in Table \ref{tab:suppl_ablations}, introducing bounding-box-aware augmentation (\textbf{v2}) results in consistent, though modest, improvements in Phrase Grounding (PG) metrics across datasets (e.g., MS-CXR IoU improves from 0.388 to 0.398). This suggests that spatial transformations help the model generalize better to anatomical coordinates that may differ slightly from the training prototypes, without requiring changes to the model architecture.

\paragraph{Curriculum Learning Frequency (v3--v5).} Configurations v3 through v5 explore the frequency of curriculum re-weighting updates. We observe that a longer accumulation window of 3000 steps (\textbf{v5}) yields comparable or slightly better performance compared to more frequent updates (1500 or 2000 steps). This indicates that the model may benefit from longer exposure to a fixed data distribution, giving more time to the model's performance on current tasks to plateau before the curriculum logic re-adjusts the sampling ratios.

\paragraph{Impact of Learning Rate on Pre-training (v6--v11).} A pivotal finding is the interaction between pre-training and learning rate. In variants v6--v8 (low LR, 2e-5), Chest ImaGenome (CIG) pre-training yielded only marginal gains over the baseline. However, increasing the learning rate to 2e-4 (v9--v11) unlocked substantial improvements. Comparing \textbf{v8} (Low LR) to \textbf{v11} (High LR, CURE), we observe a sharp increase in AGRG IoU (from 0.486 to \textbf{0.601}) and Phrase Grounding MS-CXR IoU (from 0.495 to 0.552). This implies that a higher learning rate is necessary to effectively adapt the visual encoder to fine-grained anatomical text after the initial pre-training phase. Among these high-LR variants, the 3000-step pre-training schedule (\textbf{v11}) provided the most consistent performance across tasks, serving as our final \textbf{CURE} configuration.

\paragraph{Sampling Strategy Analysis (v12--v15).} Finally, we investigate the impact of data mixing strategies. As detailed in Section~\ref{sec:CL}, our framework defines ``data sources'' as specific dataset-task pairs (e.g., PadChest-GR (task: GRG) vs. MS-CXR (task: PG)). To facilitate the analysis, we define the three sampling approaches used in variants v12--v15 as follows: \begin{itemize} \item \textit{Natural Sampling:} At the \textbf{Inter-level}, data sources are sampled strictly proportional to their size (heavily biasing training toward Chest ImaGenome). At the \textbf{Intra-level}, samples are drawn randomly without intervention, preserving the inherent clinical class imbalance.

\item \textit{Uniform Sampling:}
At the \textbf{Inter-level}, all data sources are sampled with equal probability ($1/K$).
At the \textbf{Intra-level}, samples are drawn such that each category (e.g., finding or anatomical region) has an equal probability of selection ($1/C$).

\item \textit{Curriculum Sampling:}
Sampling probabilities are dynamically re-weighted based on error rates.
At the \textbf{Inter-level}, this balances distinct data sources based on aggregate validation performance.
At the \textbf{Intra-level}, this re-weights specific intra-dataset categories based on per-class error.
\end{itemize}

\noindent We analyze the impact of these strategies on both in-domain tasks and the zero-shot out-of-distribution (OOD) benchmark, VinDr-CXR.

\begin{itemize} \item \textbf{The Risks of Natural Sampling (v15):} Variant \textbf{v15} employs a fully Natural strategy. While this achieves the absolute highest performance on the dominant Chest ImaGenome dataset (AGRG IoU \textbf{0.639}), it underperforms significantly on all other tasks. Most notably, it suffers a complete collapse on Grounded Report Generation (GRG), dropping to \textbf{0.000} IoU for both PadChest and VinDr-CXR. This confirms that without explicit re-balancing, the model overfits the largest data source and fails to acquire generalizable capabilities for auxiliary tasks.

\item \textbf{Uniform vs. Curriculum (v12 vs. v13):} Removing CL entirely and employing Uniform inter-dataset sampling (\textbf{v12}) effectively prevents the collapse seen in v15, yielding a very strong baseline that competes closely with the curriculum variants. However, applying Curriculum Learning to the inter-dataset mix (\textbf{v13}) offers marginal but consistent gains over the Uniform baseline across several benchmarks. For instance, v13 achieves higher grounding performance on PadChest (GRG IoU \textbf{0.272} vs. 0.263) and MS-CXR (IoU \textbf{0.567} vs. 0.557). This suggests that while Uniform sampling provides a robust foundation, dynamic re-weighting can squeeze out minor performance improvements by prioritizing harder tasks.

\item \textbf{Inter- vs. Intra-Dataset Dynamics (v13 vs. v14):} We further isolate the CL logic. Variant \textbf{v13} applies CL only to the \textit{Inter-dataset} mix, while \textbf{v14} applies CL only to the \textit{Intra-dataset} mix.
Variant \textbf{v13} outperforms \textbf{v14} on Grounded Report Generation tasks (e.g., GRG-PC IoU \textbf{0.272} vs. 0.246). Empirically, the curriculum logic at the inter-dataset level, present in v13, tends to assign higher sampling probabilities to the GRG task, as previously seen in Figure~\ref{fig:inter_weights}, compared to uniform or intra-only strategies, likely due to the higher difficulty of generating full grounded reports. While the differences are not dramatic, the results indicate that macro-level balancing between distinct data sources (Inter-CL) is a more effective driver of robustness than fine-grained category re-weighting (Intra-CL).
\end{itemize}

Ultimately, while Uniform sampling (\textbf{v12}) proves to be a highly effective strategy for multi-task stability, the Curriculum-based methods (v11/v13) demonstrate the capacity to further refine performance on challenging tasks like Grounded Report Generation without compromising the baseline capabilities.

\subsection{Sensitivity to $\alpha$ (Weighting Term)}
\label{sec:suppl_alpha_sensitivity}

As detailed in Section~\ref{sec:suppl_currlearn_details}, our curriculum protocol computes an aggregate performance score \(s_i\) for each data source and/or intra-dataset class using a weighted average of localization (IoU) and semantic alignment (CXRFEScore). This balance is governed by the parameter $\alpha$:
\begin{equation}
s_i = \alpha \cdot \text{IoU}_i + (1 - \alpha) \cdot \text{CXRFEScore}_i
\end{equation}
In our primary experiments, we set $\alpha=0.8$ to explicitly prioritize improvements in spatial localization accuracy. To empirically understand the sensitivity of the CURE framework to this parameter, we conducted an ablation study varying $\alpha \in \{0.0, 0.25, 0.5, 0.75, 0.8, 1.0\}$. 

Due to the computational cost of full training runs, these specific ablation experiments were restricted to 3000 steps of training in the AGRG task on the Chest ImaGenome dataset. The results, including bootstrapped standard deviations, are presented in Table~\ref{tab:suppl_alpha_ablation}.

\begin{table}[h!]
\centering
\caption{
    \textbf{Sensitivity Analysis of the Curriculum Weighting Term ($\alpha$).} 
    Performance metrics on the Chest ImaGenome dataset (AGRG task) after 3000 training steps across different values of $\alpha$. Higher values of $\alpha$ heavily weight the IoU metric during curriculum updates, while lower values prioritize the text-based semantic metric (CXRFEScore). We report mean Intersection-over-Union (\textbf{IoU}, \(\uparrow\)), CheXbert F1 (Micro/Macro averages, \(\uparrow\)), 
    CheXbert cosine similarity (\textbf{Cos.}, \(\uparrow\)), and CXRFEScore (\textbf{CXS}, \(\uparrow\)).
    \textbf{Bold} indicates the best result per column.
}
\label{tab:suppl_alpha_ablation}
\resizebox{\linewidth}{!}{%
\setlength{\tabcolsep}{4pt}
\begin{tabular}{@{}lccccc@{}}
\toprule
\textbf{Method} & \textbf{IoU} $\uparrow$ & \textbf{F1-Mi} $\uparrow$ & \textbf{F1-Ma} $\uparrow$ & \textbf{Cos.} $\uparrow$ & \textbf{CXS} $\uparrow$ \\
\midrule
CURE ($\alpha=0.0$)  & 0.582 $\pm$ 0.008 & 0.527 $\pm$ 0.017 & 0.226 $\pm$ 0.017 & 0.688 $\pm$ 0.009 & 0.556 $\pm$ 0.011 \\
CURE ($\alpha=0.25$) & 0.591 $\pm$ 0.008 & 0.522 $\pm$ 0.018 & 0.224 $\pm$ 0.019 & 0.692 $\pm$ 0.008 & 0.538 $\pm$ 0.012 \\
CURE ($\alpha=0.5$)  & 0.599 $\pm$ 0.008 & 0.529 $\pm$ 0.017 & \textbf{0.250} $\pm$ 0.017 & \textbf{0.697} $\pm$ 0.009 & 0.541 $\pm$ 0.012 \\
CURE ($\alpha=0.75$) & 0.603 $\pm$ 0.008 & 0.523 $\pm$ 0.017 & 0.228 $\pm$ 0.016 & 0.690 $\pm$ 0.009 & 0.545 $\pm$ 0.012 \\
CURE ($\alpha=0.8$)  & \textbf{0.616} $\pm$ 0.008 & 0.519 $\pm$ 0.018 & 0.215 $\pm$ 0.016 & 0.686 $\pm$ 0.009 & 0.545 $\pm$ 0.011 \\
CURE ($\alpha=1.0$)  & 0.608 $\pm$ 0.008 & \textbf{0.533} $\pm$ 0.017 & 0.236 $\pm$ 0.017 & 0.695 $\pm$ 0.009 & \textbf{0.558} $\pm$ 0.011 \\
\bottomrule
\end{tabular}%
}
\end{table}

As anticipated, increasing $\alpha$ generally yields consistent improvements in visual grounding. The model achieves its peak spatial accuracy at $\alpha=0.8$ (IoU \textbf{0.616}), confirming our design choice to prioritize visual grounding accuracy during curriculum updates. 

However, decreasing $\alpha$ toward 0.0 (which theoretically forces the curriculum to prioritize text generation quality) does not yield a monotonic improvement in clinical text metrics like F1-Macro or CXRFEScore. Instead, text performance fluctuates, peaking at moderate values ($\alpha=0.5$ or $\alpha=1.0$). 

We hypothesize that this behavior is a limitation of the current intra-dataset balancing strategy. As noted in Section~\ref{sec:task_formulation}, the intra-dataset curriculum for AGRG is designed to balance exposure across \textit{anatomical locations}, but it does not actively re-weight the distribution of \textit{clinical findings}. Consequently, even when a lower $\alpha$ signals the model to focus heavily on the text generation objective, the model's ability to improve its text metrics is constrained by the natural, long-tailed imbalance of pathological findings inherent to the dataset. Future iterations of the CURE framework could address this by implementing finer-grained, multi-dimensional balancing strategies that stratify samples based simultaneously on both anatomical regions and the prevalence of specific clinical findings.

\subsection{Full Results Tables}
The following tables contain the complete, unabridged results for each evaluation task and dataset, from which the summary tables in the main paper were derived.

\subsubsection{Anatomy-Grounded Report Generation (AGRG)}

Table \ref{tab:suppl_agrg_results} presents a comprehensive breakdown of performance on the Anatomy-Grounded Report Generation (AGRG) task, isolating the effects of pre-training strategies and multi-task fine-tuning configurations.

\begin{table*}[h!]
\centering
\caption{
    \textbf{Detailed Results for Anatomy-Grounded Report Generation (AGRG).}
    Performance of baseline models, pre-training-only checkpoints, and the full set of multi-task fine-tuning ablation variants (v1--v15) on the Chest ImaGenome test subset.
    We report mean Intersection-over-Union (\textbf{IoU}, \(\uparrow\)), CheXbert F1 (Micro/Macro averages, \(\uparrow\)), 
    CheXbert cosine similarity (\textbf{Cos.}, \(\uparrow\)), and CXRFEScore (\textbf{CXS}, \(\uparrow\)).
    \textbf{Bold} indicates the best result per column; \underline{underlined} indicates the second best.
}
\label{tab:suppl_agrg_results}

\input{tables_suppl_agrg_results}
\end{table*}

\paragraph{Trade-offs Between Grounding and Clinical Label Accuracy.}
Comparing the baseline (\textbf{v1}) with the optimized variants (v11--v15) highlights a stark contrast in the rate of improvement between spatial and text-based metrics. While the optimized models achieve substantial gains in spatial precision (with IoU jumping from 0.380 to $>0.59$), the text generation metrics do not scale proportionally. Semantic metrics such as CheXbert Cosine Similarity and CXRFEScore show only modest improvements, and the baseline model actually retains the highest CheXbert F1-Macro score (0.251), whereas most high-IoU variants fluctuate between 0.21 and 0.25. We hypothesize that this discrepancy stems from our current curriculum design. For the AGRG task, the intra-dataset re-weighting strategy is explicitly designed to balance \textit{anatomical locations} to ensure robust localization, but it does not currently account for the highly imbalanced distribution of \textit{clinical findings}. Consequently, while we observe dramatic improvements in spatial grounding and stable performance on dominant text classes (F1-Micro), the model does not fully benefit from re-balancing rare pathological conditions. This highlights a clear avenue for future work: designing a multidimensional re-weighting strategy that simultaneously targets anatomical diversity and the distribution of rare clinical findings to improve semantic report quality on long-tailed conditions.

\paragraph{Pre-training Efficiency.}
The ``Chest ImaGenome (CIG) Pre-training Only'' block highlights the impact of learning rate scaling. With a conservative learning rate (2e-5), extending pre-training from 1000 to 3000 steps yields only marginal IoU gains (0.378 $\rightarrow$ 0.430). Conversely, increasing the learning rate to 2e-4 results in a substantial improvement, with the 3000-step high-LR variant achieving an IoU of 0.596 even before multi-task fine-tuning. This confirms that aligning visual features with fine-grained anatomy-grounded reports benefits from more aggressive optimization during the initial training stages.

\paragraph{Performance of Sampling Strategies.}
Among the final sampling variants, we observe that \textbf{v15} (Natural Sampling) achieves the highest scores across all metrics in this specific task (IoU \textbf{0.639}, Cos. Sim. \textbf{0.694}, CXS \textbf{0.554}). This result is expected given the data distribution: \textbf{v15} undergoes 3000 steps of pre-training and 6000 steps of fine-tuning where samples are drawn proportional to dataset size. Since Chest ImaGenome dominates the training mixture, \textbf{v15} is effectively trained on AGRG for the vast majority of these $\sim$9,000 steps. However, as detailed in the ablation study (Section~\ref{sec:extended_ablation}), this specialization leads to severe degradation on complementary tasks (GRG and PG) on other datasets.
The strategies that actively intervene on the data distribution—Uniform (\textbf{v12}) and the Curriculum variants (\textbf{v11}, \textbf{v13}, \textbf{v14})—maintain competitive in-domain performance (IoU $\sim$0.60, CXS $\sim$0.53--0.55) while preventing the task collapse observed in \textbf{v15}. Notably, all proposed variants (v9--v15) significantly outperform the external MAIRA-2 baseline in both spatial grounding (IoU $\sim$0.59--0.64 vs. 0.249) and semantic alignment (Cos. Sim. $>0.67$ and CXS $>0.52$ vs. 0.662 and 0.467, respectively).

\subsubsection{Phrase Grounding (PG)}

Table~\ref{tab:suppl_pg_results} details the Phrase Grounding performance across three diverse benchmarks: MS-CXR (in-domain), PadChest-GR (in-domain), and VinDr-CXR (zero-shot, unseen distribution).

\begin{table*}[h!]
\footnotesize
\centering
\caption{
    \textbf{Detailed Results for Phrase Grounding (PG).} We report Micro-Average IoU (\textbf{IoU Mi.} \(\uparrow\)) and Macro-Average IoU (\textbf{IoU Ma.} \(\uparrow\)) on three test sets: MS-CXR, PadChest-GR, and zero-shot VinDr-CXR. \textbf{Bold} indicates best; \underline{underlined} indicates second best.
}
\label{tab:suppl_pg_results}

\input{tables_suppl_pg_results}
\end{table*}

\paragraph{Generalization via Augmentation.}
Adding data augmentation (\textbf{v2}) to the baseline (\textbf{v1}) yields modest but consistent spatial improvements. For instance, MS-CXR IoU Micro increases from 0.388 to 0.398, and zero-shot VinDr-CXR IoU improves from 0.191 to 0.203. This indicates that bounding-box-aware augmentations effectively reduce overfitting and help the model generalize to varied image acquisitions.

\paragraph{Progressive Improvements: Curriculum, Pre-training, and Learning Rate.}
The results demonstrate a cumulative benefit from each component of the CURE pipeline. First, introducing curriculum learning alone (variants \textbf{v3--v5}) yields a moderate gain over the augmented baseline (e.g., \textbf{v5} reaches 0.430 IoU on MS-CXR vs. 0.398 for \textbf{v2}). Second, adding Chest ImaGenome pre-training with a conservative learning rate (variants \textbf{v6--v8}) pushes performance further, with \textbf{v8} reaching 0.495 IoU. Finally, the most dramatic jump occurs when increasing the learning rate to 2e-4 for \textit{both} the pre-training and multi-task fine-tuning phases (variants \textbf{v9--v11}). Comparing \textbf{v8} (Low LR) to \textbf{v11} (High LR), we observe an improvement of over 5 points on MS-CXR (0.495 $\rightarrow$ 0.552). This confirms that a higher learning rate is essential throughout the entire pipeline to fully align visual features with text and escape local minima.

\paragraph{Sampling Strategy Dynamics.}
Analyzing the sampling strategies (v10--v15) reveals distinct performance profiles across datasets:
\begin{itemize}
    \item \textbf{Natural Sampling Failure:} Variant \textbf{v15} suffers a severe regression, dropping to 0.355 IoU on MS-CXR (worse than the un-augmented baseline \textbf{v1}). This confirms that without intervention, the dominance of AGRG data overwhelms the signal from smaller grounding datasets.
    \item \textbf{Peak Performance (v10 vs. v13):} While the proposed CURE model (\textbf{v11}) is highly competitive, the absolute peak performance for phrase grounding is split between \textbf{v10} and \textbf{v13}. Variant \textbf{v10} (2k pre-training) achieves the highest scores on MS-CXR (\textbf{0.574}) and the zero-shot VinDr-CXR (\textbf{0.248}), suggesting that a slightly shorter pre-training phase may occasionally favor pure localization tasks. Conversely, \textbf{v13} (Inter-CL) achieves the best performance on PadChest-GR (\textbf{0.464}), indicating that dynamic inter-dataset balancing effectively captures the nuances of that specific distribution.
    \item \textbf{Overall Robustness:} Despite these minor variations, all high-LR curriculum variants (v10, v11, v13, v14) significantly outperform the external MAIRA-2 baseline (e.g., $\sim$0.24 vs. 0.16 on VinDr-CXR), validating the general effectiveness of the proposed framework.
\end{itemize}

\subsubsection{Grounded Report Generation (GRG)}

Table \ref{tab:suppl_grg_results} presents the ablation results for the Grounded Report Generation (GRG) task on PadChest-GR and the zero-shot VinDr-CXR benchmark. This task is the most challenging in our suite, requiring the model to simultaneously generate a full radiology report and localize every mentioned finding.

\begin{table*}[t]
\centering
\footnotesize
\caption{
    \textbf{Detailed Results for Grounded Report Generation (GRG).}
    Performance on PadChest-GR and the zero-shot VinDr-CXR benchmark.
    We report mean IoU (\(\uparrow\)), CheXbert F1 (micro/macro, \(\uparrow\)), 
    CheXbert cosine similarity (\textbf{Cos.}, \(\uparrow\)), and CXRFEScore (\textbf{CXS}, \(\uparrow\)).
    High-learning-rate variants (v9--v13) consistently achieve superior localization (IoU) compared to baselines.
    \textbf{Bold} indicates best; \underline{underlined} indicates second best.
}
\label{tab:suppl_grg_results}

\input{tables_suppl_grg_results}
\end{table*}

\paragraph{Localization vs. Clinical Metrics.}
A clear divergence emerges when comparing the low-learning-rate variants (\textbf{v1--v8}) with the high-learning-rate variants (\textbf{v9--v14}). The low-LR models often achieve higher scores on clinical text metrics; for instance, \textbf{v5} achieves a zero-shot F1-Micro of \textbf{0.626} on VinDr-CXR but has limited localization accuracy (IoU 0.217). Conversely, the high-LR variants significantly boost visual grounding (e.g., \textbf{v13} reaches IoU \textbf{0.266}) but often see a regression in text metrics (F1-Micro drops to 0.434). This pattern suggests that while aggressive updates are necessary to learn the structural constraints of the GRG task (i.e., outputting bounding box coordinates), our current training protocol heavily favors the grounding objective. Future work is likely needed to design more sophisticated re-balancing strategies—such as balancing positive versus negative findings or specific finding classes—to simultaneously enhance clinical reporting metrics without sacrificing localization performance.

\paragraph{Benchmarking Against MAIRA-2.}
The PadChest-GR results are particularly significant. This task theoretically favors the MAIRA-2 baseline, which benefits from training on both PadChest-GR and the large, proprietary USMix dataset \cite{maira2} (containing $\sim$70k grounded reports). Despite this data disadvantage, our high-LR variants (\textbf{v10}, \textbf{v11}, \textbf{v12}, \textbf{v13}) consistently surpass MAIRA-2 in localization performance (e.g., \textbf{v13} IoU \textbf{0.272} vs. 0.256).
On the zero-shot VinDr-CXR benchmark, this trend holds, with most high-LR variants significantly outperforming MAIRA-2 (\textbf{0.266} vs. 0.217).
However, \textbf{v15} (Natural Sampling) illustrates a critical failure mode: it achieves an IoU of \textbf{0.000} on both datasets yet records the highest Cosine Similarity on VinDr-CXR (\textbf{0.858}). 
This suggests that due to the overwhelming dominance of AGRG data (Chest ImaGenome), the model fails to learn the specific formatting requirements of the minority GRG task. Despite this, when given the GRG instruction to ``Generate a grounded report'', it tends to behave like a standard captioning model: it generates clinically plausible text, but these descriptions are heavily biased by the style of the mini-reports used in the AGRG task. The task collapse here is thus a failure to acquire the correct output format due to extreme data imbalance, rather than a total loss of visual understanding.

\paragraph{Impact of Sampling Strategies.}
We focus on the comparison between \textbf{v12} (Uniform Inter-sampling) and \textbf{v13} (Curriculum Inter-sampling). As noted in Section~\ref{sec:CL}, PadChest-GR does not utilize intra-dataset curriculum re-weighting; therefore, differences in performance stem primarily from how the model balances the distinct data sources.
Variant \textbf{v13} achieves the highest IoU on both datasets, surpassing the Uniform baseline (\textbf{v12}). This indicates that dynamically up-weighting the GRG data sources (based on error rates) helps the model prioritize the visual grounding objective more effectively than static uniform sampling. While the margins are modest, \textbf{v13} consistently provides the most robust grounding performance across the evaluated benchmarks.

\begin{table*}[ht]
\centering
\footnotesize
\caption{
    \textbf{Results for Report Generation (RG) on the MIMIC-CXR test set.}
    We evaluate state-of-the-art baselines, a model fine-tuned solely for report generation (\textbf{MedGemma-FT (RG)}), and the proposed \textbf{CURE} model. 
    Notably, since CURE is multi-task, we explore different inference protocols: \textbf{GRG} (generating a single grounded report), \textbf{AGRG-$N$} (concatenating location-specific descriptions for $N$ anatomical regions), and their combinations.
    We report CheXbert F1, Precision (P), and Recall (R) (Micro/Macro averaged), CheXbert Cosine Similarity (\textbf{Cos.}), CXRFEScore (\textbf{CXS}), RaTEScore (\textbf{RaTES}), and RadGraph F1 (\textbf{RadF1}).
    \textbf{Bold} and \underline{underlined} values indicate the best and second-best scores.
}
\label{tab:suppl_mimiccxr_report_gen_results}

\input{tables_suppl_mimiccxr_report_gen_results}
\end{table*}

\subsubsection{Standard Report Generation (MIMIC-CXR)}

Table \ref{tab:suppl_mimiccxr_report_gen_results} provides a detailed breakdown of report generation performance on the MIMIC-CXR test set. We analyze how different inference protocols ranging from single-prompt generation to multi-location concatenation affect the trade-off between precision, recall, and semantic alignment.

\begin{table}[h!]
\centering
\scriptsize
\caption{
    \textbf{Anatomical Query Configurations.} 
    Definition of the location sets used for the AGRG inference protocols. 
    \textbf{AGRG-29} represents the set of locations with complete supervision (Bounding Box + Text) in Chest ImaGenome. 
    \textbf{AGRG-38} extends this to include locations that have only text supervision. 
    \textbf{AGRG-9} is a subset of core locations.
}
\label{tab:anatomical_sets}
\begin{tabularx}{\linewidth}{@{}l >{\raggedright\arraybackslash}X @{}}
\toprule
\textbf{Config.} & \textbf{Anatomical Locations Included} \\
\midrule
\textbf{AGRG-9} & \textit{Core Locations:} Abdomen, Cardiac Silhouette, Left/Right Costophrenic Angle, Left/Right Lung, Mediastinum, Spine, Trachea. \\
\midrule
\textbf{AGRG-29} & \textit{Includes all AGRG-9 plus:} Aortic Arch, Carina, Cavoatrial Junction, SVC, Upper Mediastinum, Left/Right Apical Zone, Left/Right Mid Lung Zone, Left/Right Lower Lung Zone, Left/Right Upper Lung Zone, Left/Right Hilar Structures, Left/Right Clavicle, Left/Right Hemidiaphragm, Right Atrium. \\
\midrule
\textbf{AGRG-38} & \textit{Includes all AGRG-29 plus:} Left/Right Arm, Left/Right Breast, Left/Right Chest Wall, Left/Right Shoulder, Neck. \\
\bottomrule
\end{tabularx}
\end{table}

\paragraph{Impact of Anatomical Granularity ($N$).}
A unique feature of the CURE framework is the ability to modulate the ``resolution'' of the generated report by varying the number of queried anatomical locations ($N$). The compositions of these sets are detailed in Table \ref{tab:anatomical_sets}.
\begin{itemize}
    \item \textbf{AGRG-9 (High Precision):} By querying only 9 core locations, the model achieves high precision (P-Mi: \textbf{0.639}), comparable to the grounded reports of MAIRA-2 (P-Mi: \textbf{0.639}). This configuration also ties for the highest RadGraph F1 score among the CURE variants, making it comparatively more suitable for scenarios where minimizing false positives is a priority.
    \item \textbf{AGRG-29 (Balanced Supervision):} This set comprises the 29 anatomical locations for which Chest ImaGenome provides both bounding box and text supervision. Using this configuration yields a strong balance of metrics, notably achieving the second-highest RaTEScore (\underline{0.592}), validating the quality of training on fully grounded data.
    \item \textbf{AGRG-38 (High Recall):} Expanding to 38 locations includes peripheral areas (e.g., neck, chest wall) that possess text supervision but lack bounding box annotations in the training data. Forcing the model to scrutinize these areas results in the highest Recall (R-Mi 0.770) and CheXbert Cosine Similarity (\underline{0.793}) among the purely anatomical approaches. However, this exhaustive search introduces a trade-off: while recall improves, precision drops significantly (P-Mi 0.408) compared to the concise AGRG-9 configuration (P-Mi 0.639), leading to slightly lower aggregate F1 scores compared to the AGRG-29 configuration.
\end{itemize}

\paragraph{Baseline Performance Analysis.}
We observe distinct performance profiles across the evaluated baselines. \textbf{CXRMate-RRG24} \cite{cxrmate-rrg24} establishes a strong benchmark, achieving the highest RadGraph F1 (\textbf{0.255}), CheXbert F1-Micro (\textbf{0.589}), and CXRFEScore (\textbf{0.656}), reflecting its optimization via reinforcement learning for clinical correctness. For \textbf{MAIRA-2}, the inference mode dictates a clear trade-off: the grounded mode maximizes precision (P-Mi \textbf{0.639}) but acts as a constraint that limits recall (0.397), whereas disabling grounding improves clinical finding detection (F1-Mi 0.554) but degrades semantic alignment (Cosine drops to 0.693). Finally, comparing the \textbf{MedGemma} variants reveals the impact of domain adaptation. The base \textbf{MedGemma-4B-IT} exhibits high recall (0.692) but low precision (0.452) and structural accuracy (RadF1 0.112). Surprisingly, fine-tuning solely on reports (\textbf{MedGemma-FT (RG)}) resulted in lower CheXbert F1 scores compared to the base model (e.g., F1-Ma 0.353 vs 0.382), although it significantly improved precision to \underline{0.621} and achieved the second-best RadGraph F1 (\underline{0.203}). However, this specialized baseline still lags behind the multi-task CURE variants in multiple metrics, such as RaTEScore (\textbf{0.597} vs. 0.536) and Cosine Similarity (0.792 vs. 0.753). This suggests that the explicit grounding tasks in CURE provide a more robust supervision signal for learning to describe radiological findings than standard text-only fine-tuning.

\paragraph{Benefits of Hybrid Inference.}
The standalone GRG prompt produces concise reports but yields lower recall (R-Mi: 0.376). Combining this global summary with fine-grained anatomical descriptions (\textbf{AGRG+GRG}) consistently yields the strongest empirical balance in our experiments. Specifically, the \textbf{AGRG-29 + GRG} configuration achieves the highest \textbf{RaTEScore} (\textbf{0.597}) in the table. Furthermore, this configuration achieves a CheXbert F1-Macro of \textbf{0.415}, marginally outperforming the state-of-the-art model CXRMate-RRG24~\cite{cxrmate-rrg24} (\underline{0.414}), the winner of a recent report generation competition. This indicates that fusing a holistic global grounded report with specific, visually grounded regional descriptions is a promising strategy to bridge the gap between precision and recall in radiology report generation.

\paragraph{Additional Comparison with Baselines.}
CURE demonstrates strong performance against specialized baselines. While \textbf{CXRMate-RRG24} \cite{cxrmate-rrg24} retains the top performance on RadGraph F1 (0.255 vs. 0.176), CURE outperforms it on \textbf{RaTEScore} (0.597 vs. 0.577) and CheXbert Cosine Similarity (0.792 vs. 0.764). The strong performance on RaTEScore—a recently proposed metric designed to assess medical entities and robustness to synonyms—highlights CURE's ability to generate clinically relevant content. Furthermore, the hybrid CURE configurations consistently surpass the MAIRA-2 baseline in semantic and factual consistency metrics; specifically, \textbf{AGRG-29 + GRG} achieves a CXRFEScore of 0.655 (compared to MAIRA-2's 0.603) and a RaTEScore of 0.597 (compared to MAIRA-2's 0.496).

\begin{table*}[b]
\centering
\caption{
\textbf{Per-anatomy comparison of MAIRA-2 and CURE performance on the Chest ImaGenome subset (Part 1/2).}
Each anatomical region reports hallucination and correctness rates (\%) for abnormalities and medical devices, 
along with Natural Language Inference (NLI) consistency metrics: Contradiction (Cont.) and Entailment (Ent.). 
Lower hallucination and contradiction values, together with higher correctness and entailment, 
indicate greater clinical agreement between the generated and ground-truth reports. 
See Table~\ref{tab:suppl_cig_hallucination_results_part2} for continuation.
}
\label{tab:suppl_cig_hallucination_results_part1}

\input{tables_suppl_cig_hallucination_results_part1}
\end{table*}

\begin{table*}[b]
\centering
\caption{
\textbf{Per-anatomy comparison of MAIRA-2 and CURE performance on the Chest ImaGenome subset (Part 2/2).}
Continuation of Table~\ref{tab:suppl_cig_hallucination_results_part1}. 
Each row corresponds to an anatomical region evaluated for abnormality and device hallucination (\%) and correctness (\%), 
as well as NLI-based Contradiction (Cont.) and Entailment (Ent.) rates. 
Lower hallucination and contradiction, and higher correctness and entailment, 
reflect more clinically faithful and factually consistent report generation.
}
\label{tab:suppl_cig_hallucination_results_part2}

\input{tables_suppl_cig_hallucination_results_part2}
\end{table*}

\subsection{Extended Hallucination Analysis}
\label{sec:suppl:extended-hallucination-analysis}
To provide a holistic view of model reliability, we extend the hallucination analysis from the main paper to the full spectrum of the Chest ImaGenome schema. As detailed in the anatomical configurations (Table~\ref{tab:anatomical_sets}), we focus specifically on the locations for which text supervision is available (typically accompanied by bounding boxes). This selection excludes locations with bounding-box-only supervision, resulting in a comprehensive evaluation set of 38 anatomical regions. Tables~\ref{tab:suppl_cig_hallucination_results_part1} and \ref{tab:suppl_cig_hallucination_results_part2} present the performance breakdown across these locations.

\paragraph{Methodology.}
Unlike the focused analysis in the main paper, this supplementary evaluation covers all 38 anatomical regions present in the training set with text supervision (out of 45 total locations in the schema). For each anatomy, we sampled 300 instances from the test set. To assess MAIRA-2, we utilized its \textbf{phrase grounding} capability, prompting the model with the specific anatomical name (e.g., ``left clavicle'') to elicit a grounded response using its official prompt template available at \url{https://huggingface.co/microsoft/maira-2}.

We employed \texttt{gemini-2.5-flash-lite} as an automated clinical judge to compare the anatomy-specific generation (GEN) against the full ground-truth report (GT). Using a Chain-of-Thought (CoT) prompting strategy, the judge evaluated:
\begin{enumerate}
    \item \textbf{Correctness:} Does GEN successfully retrieve findings present in GT?
    \item \textbf{Hallucination:} Does GEN invent findings not supported by GT?
    \item \textbf{NLI Consistency:} What is the logical relationship (Entailment, Contradiction, Neutral) between GEN and GT?
\end{enumerate}

\paragraph{Evaluation Prompt.}
The exact prompt utilized for the automated judge is provided below. It enforces an independent extraction step before comparison to minimize reasoning errors.

\begin{lstlisting}[style=promptstyle]
You are an expert radiologist. Your task is to compare a short anatomy-specific mini-report [GEN] against a full image ground-truth report [GT], where [GT] was generated by a radiologist over the entire image, whereas [GEN] was generated by a model over a specific anatomical location. You will assess the degree of hallucination and contradiction in [GEN] compared to [GT].

First, independently assess each report:
- If [GT] explicitly affirms the presence of any abnormality, set "gt_has_abnormalities" to "yes". Otherwise, set it to "no".
- If [GT] explicitly affirms the presence of any medical device (e.g., pacemaker, catheter, wires), set "gt_has_devices" to "yes". Otherwise, set it to "no".
- If [GEN] explicitly affirms the presence of any abnormality, set "gen_has_abnormalities" to "yes". Otherwise, set it to "no".
- If [GEN] explicitly affirms the presence of any medical device (e.g., pacemaker, catheter, wires), set "gen_has_devices" to "yes". Otherwise, set it to "no".

Next, perform the comparison based on [GT]:
- If [GEN] affirms the presence of an abnormality and this is clearly supported or reasonably suggested by [GT], set "gen_has_correct_abnormalities" to "yes". Otherwise, set it to "no".
- If [GEN] affirms the presence of an abnormality that is NOT affirmed nor supported by [GT], set "gen_has_hallucinated_abnormalities" to "yes". Otherwise, set it to "no".
- If [GEN] affirms the presence of a device that is clearly supported or reasonably suggested by [GT], set "gen_has_correct_devices" to "yes". Otherwise, set it to "no".
- If [GEN] affirms the presence of a device that is NOT affirmed nor supported by [GT], set "gen_has_hallucinated_devices" to "yes". Otherwise, set it to "no".
- Natural Language Inference:
    - If [GEN] makes at least one explicit statement that is clearly contradicted by [GT], set "nli_status" to "contradiction".
    - If all of [GEN]'s explicit statements are reasonably supported by [GT], set "nli_status" to "entailment".
    - Otherwise, set "nli_status" to "neutral".

You must respond ONLY with a single, valid JSON object in the following format. Do not add any text before or after the JSON object.

{
    "reason": "A detailed explanation of your reasoning for the comparison. Include a brief explanation of why you made your choices for each field. Focus on what is explicitly stated in [GEN] and [GT]. Do not make any assumptions about what is not explicitly stated.",
    "gt_has_abnormalities": "yes" | "no",
    "gt_has_devices": "yes" | "no",
    "gen_has_abnormalities": "yes" | "no",
    "gen_has_devices": "yes" | "no",
    "gen_has_correct_abnormalities": "yes" | "no",
    "gen_has_hallucinated_abnormalities": "yes" | "no",
    "gen_has_correct_devices": "yes" | "no",
    "gen_has_hallucinated_devices": "yes" | "no",
    "nli_status": "contradiction" | "entailment" | "neutral"
}
\end{lstlisting}

\paragraph{Results and Analysis.}
The comprehensive breakdown in Tables~\ref{tab:suppl_cig_hallucination_results_part1} and \ref{tab:suppl_cig_hallucination_results_part2} highlights distinct behavioral profiles:

\begin{itemize}
    \item \textbf{MAIRA-2 and NLI Neutrality:} MAIRA-2 exhibits a notably high \textit{Neutral} NLI rate (73.2\%) and low Abnormality Correctness (3.6\%). This is the behavior one would generally expect from MAIRA-2's phrase grounding formulation: when prompted with an anatomical phrase, the model frequently outputs the phrase verbatim with coordinates, without adding descriptive adjectives. Since the output merely identifies the anatomy without making a clinical claim, the NLI judge correctly labels the relationship to the ground truth as \textit{Neutral}. However, we observe that MAIRA-2 does occasionally append additional descriptions. When this occurs, it is highly prone to hallucination (14.9\% rate), suggesting that deviations from the standard localization behavior often result in factual errors rather than useful clinical insights.
    
    \item \textbf{Sensitivity-Specificity Trade-off:} CURE demonstrates a favorable shift in the trade-off between Sensitivity (the ability to correctly identify abnormalities) and Specificity (the ability to avoid false positives). In this context, we associate \textit{Abnormality Correctness} with Sensitivity and \textit{Hallucination Rate} with the inverse of Specificity. CURE maintains a hallucination rate comparable to MAIRA-2 (15.2\% vs. 14.9\%) while achieving a massive improvement in Correctness (17.8\% vs 3.6\%, an approximately $5\times$ increase). This indicates that CURE's generations are more clinically useful and aligned with radiologist findings (Entailment: \textbf{43.2\%} vs. 9.3\%), rather than defaulting to the ``safe silence'' of simple object localization.

    \item \textbf{Anatomical Specificity:} CURE demonstrates remarkable robustness on structures where MAIRA-2 fails. For instance, on the \textbf{Left and Right Clavicles}, MAIRA-2's attempts to describe the region result in hallucination rates of 59.0\% and 62.7\%, respectively. This likely reflects a bias in MAIRA-2's training distribution, where mentions of the clavicles were presumably highly correlated with fractures, leading the model to hallucinate pathology even when performing a grounding task. In contrast, CURE reduces this hallucination rate to $1.0\%$ in both cases.

    \item \textbf{Device Recognition:} CURE significantly outperforms MAIRA-2 in identifying medical devices (\textbf{14.0\%} Correctness vs. 1.3\%). It is important to note that MAIRA-2's low performance here is largely a consequence of the task formulation. MAIRA-2 is designed for specific tasks, and when performing phrase grounding using the standard prompt, it effectively localizes the structure but does not spontaneously describe the presence of devices (e.g., catheters). It is not designed as a flexible instruction-following model that can be prompted to exhaustively list findings. CURE, conversely, is trained on the AGRG objective to inherently describe the contents of the anatomical region, leading to stronger detection rates in device-heavy regions like the \textit{Cavoatrial Junction} (45.3\% Correctness).
\end{itemize}

\paragraph{Qualitative Example.}
Table~\ref{tab:qualitative_eval_example} provides a concrete instance of the evaluation protocol applied to the \textit{Left Clavicle}, validating the statistical trends observed above. In this scenario, the ground truth explicitly affirms that ``Bony structures are intact.'' MAIRA-2, reflecting the high hallucination bias observed for this anatomy ($\approx60\%$), generates a specific but incorrect finding: ``Left clavicle fracture is noted.'' The automated judge correctly identifies this incompatibility, marking it as a \textit{Contradiction} and a \textit{Hallucination}. In contrast, CURE correctly generates a negative finding (``No acute osseous abnormalities''), which the judge recognizes as supported by the ground truth (\textit{Entailment}). This example illustrates the robustness of the automated judge in discerning clinical nuances and highlights the tangible quality improvement achieved by CURE in avoiding specific anatomical hallucinations.

\begin{table*}[t]
\centering
\caption{\textbf{Qualitative Example of the Automated Evaluation Protocol.} 
We employ \texttt{gemini-2.5-flash-lite} to perform detailed hallucination and Natural Language Inference (NLI) analysis, utilizing the prompt defined in Section~\ref{sec:suppl:extended-hallucination-analysis}. 
This case illustrates the evaluation for the \textit{Left Clavicle}. 
The automated judge compares the model-generated anatomy-specific report (GEN) against the full ground-truth report (GT). 
MAIRA-2 hallucinates a fracture, leading to a \textit{Contradiction}, whereas CURE correctly identifies the lack of abnormalities, resulting in \textit{Entailment}.}
\label{tab:qualitative_eval_example}
\input{tables_suppl_gemini_eval_example}
\end{table*}

\section{Additional Qualitative Analysis}
\label{sec:suppl_qualitative}

To provide further insight into model behavior, this section includes additional qualitative examples from both the Grounded Report Generation (GRG) and Anatomy Grounded Report Generation (AGRG) tasks.

\begin{figure*}[t]
    \centering
    \setlength{\tabcolsep}{4pt}
    \renewcommand{\arraystretch}{1.2}
    
    \resizebox{0.95\textwidth}{!}{%
    \begin{tabular}{p{0.32\linewidth} p{0.32\linewidth} p{0.32\linewidth}}
        \includegraphics[width=\linewidth]{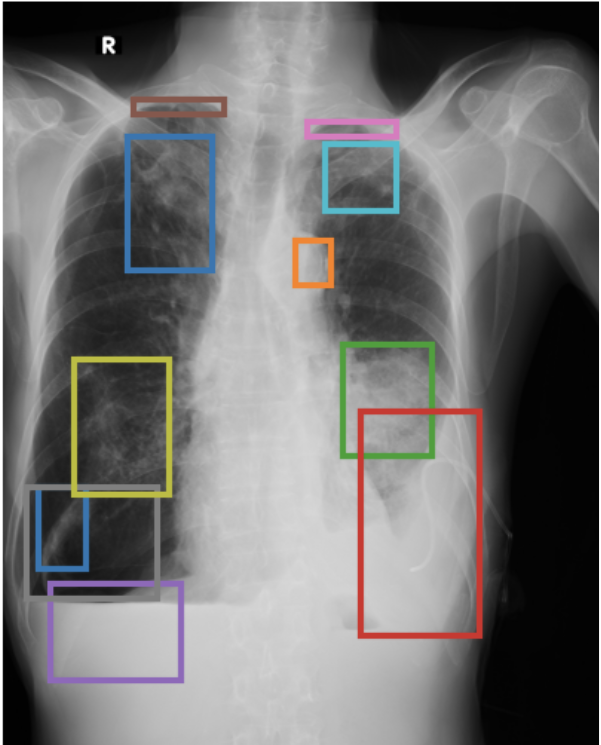} & 
        \includegraphics[width=\linewidth]{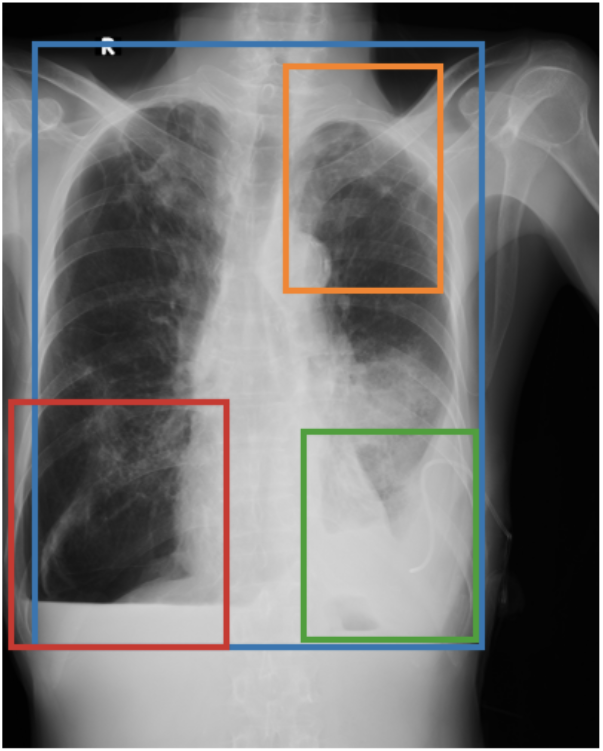} & 
        \includegraphics[width=\linewidth]{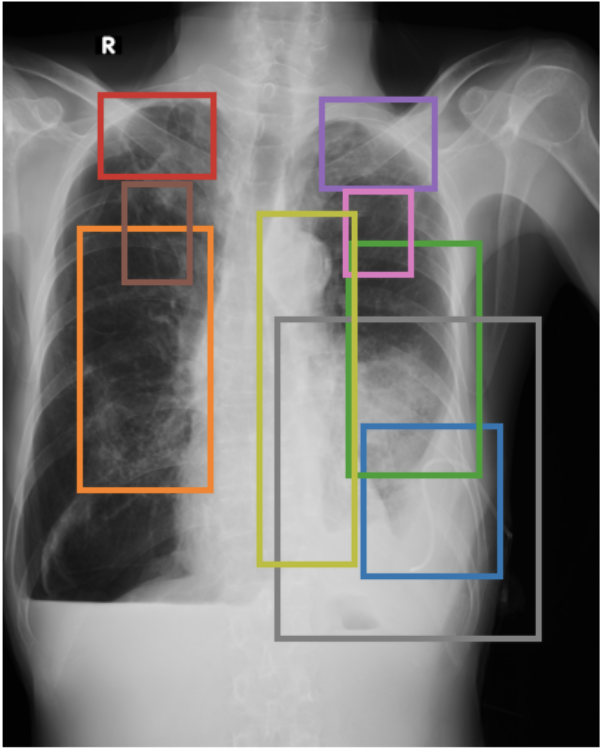} \\
        
        \textbf{(a) Ground Truth} & \textbf{(b) MAIRA-2} & \textbf{(c) CURE} \\
        
        \scriptsize \raggedright
        Atelectasis \textcolor[HTML]{1F77B4}{\textbf{[0.06,0.65,0.14,0.76]}}. Calcification \textcolor[HTML]{FF7F0E}{\textbf{[0.49,0.32,0.55,0.38]}}. Consolidation \textcolor[HTML]{2CA02C}{\textbf{[0.57,0.46,0.72,0.61]}}. Pleural effusion \textcolor[HTML]{D62728}{\textbf{[0.60,0.55,0.80,0.85]}} \textcolor[HTML]{9467BD}{\textbf{[0.08,0.78,0.30,0.91]}}. Pleural thickening \textcolor[HTML]{8C564B}{\textbf{[0.22,0.13,0.37,0.15]}} \textcolor[HTML]{E377C2}{\textbf{[0.51,0.16,0.66,0.18]}}. Pneumothorax \textcolor[HTML]{7F7F7F}{\textbf{[0.04,0.65,0.26,0.80]}}. Pulmonary fibrosis \textcolor[HTML]{BCBD22}{\textbf{[0.12,0.48,0.28,0.66]}} \textcolor[HTML]{17BECF}{\textbf{[0.54,0.19,0.66,0.28]}} \textcolor[HTML]{1F77B4}{\textbf{[0.21,0.18,0.35,0.36]}}. Tuberculosis. Other diseases.
        & 
        \scriptsize \raggedright
        The heart size is normal. Lung fields are hyperaerated bilaterally. \textcolor[HTML]{1F77B4}{\textbf{[0.055, 0.055, 0.805, 0.865]}} There is pleural calcification in the left upper lobe. \textcolor[HTML]{FF7F0E}{\textbf{[0.475, 0.085, 0.735, 0.385]}} There is blunting of the left cp angle. \textcolor[HTML]{2CA02C}{\textbf{[0.505, 0.575, 0.795, 0.855]}} There is pleural calcification in the right lower hemithorax. \textcolor[HTML]{D62728}{\textbf{[0.015, 0.535, 0.375, 0.865]}} There is blunting of the right cp angle.
        \vspace{0.3em} \\
        \textbf{Metrics:} \\
        \renewcommand{\arraystretch}{1.0}
        \begin{tabular}{@{}ll@{}}
        IoU: & 0.311 \\
        CheXbert Cos: & 0.767 \\
        CheXbert Acc: & 0.571 \\
        CXRFEScore: & 0.416 \\
        \end{tabular}
        & 
        \scriptsize \raggedright
        Left pleural effusion without significant changes \textcolor[HTML]{1F77B4}{\textbf{[0.72,0.67,0.23,0.20]}}. Chronic changes in the lung parenchyma \textcolor[HTML]{FF7F0E}{\textbf{[0.24,0.48,0.22,0.35]}} \textcolor[HTML]{2CA02C}{\textbf{[0.69,0.48,0.22,0.31]}}. Biapical pleuroparenchymal thickening \textcolor[HTML]{D62728}{\textbf{[0.26,0.18,0.19,0.11]}} \textcolor[HTML]{9467BD}{\textbf{[0.63,0.19,0.19,0.12]}}. Calcified granulomas \textcolor[HTML]{8C564B}{\textbf{[0.26,0.31,0.11,0.13]}} \textcolor[HTML]{E377C2}{\textbf{[0.63,0.31,0.11,0.11]}}. Changes from left mastectomy \textcolor[HTML]{7F7F7F}{\textbf{[0.68,0.64,0.44,0.43]}}. Aortic elongation \textcolor[HTML]{BCBD22}{\textbf{[0.51,0.52,0.16,0.47]}}
        \vspace{0.3em} \\
        \textbf{Metrics:} \\
        \renewcommand{\arraystretch}{1.0}
        \begin{tabular}{@{}ll@{}}
        IoU: & \textbf{0.320} \\
        CheXbert Cos: & \textbf{0.834} \\
        CheXbert Acc: & \textbf{0.643} \\
        CXRFEScore: & \textbf{0.523} \\
        \end{tabular}
    \end{tabular}%
    } 

    \vspace{0.3cm}
    \hrule
    \vspace{0.3cm}

    \resizebox{0.95\textwidth}{!}{%
    \begin{tabular}{p{0.32\linewidth} p{0.32\linewidth} p{0.32\linewidth}}
        \includegraphics[width=\linewidth]{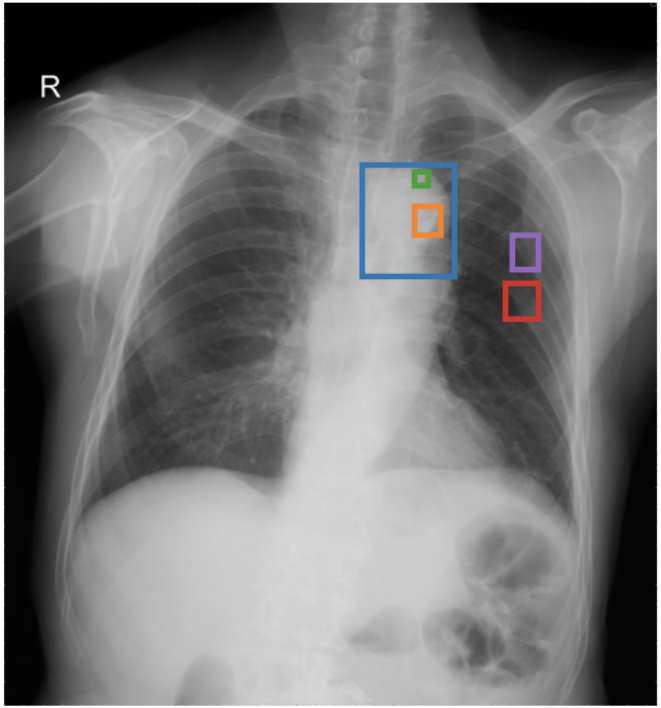} & 
        \includegraphics[width=\linewidth]{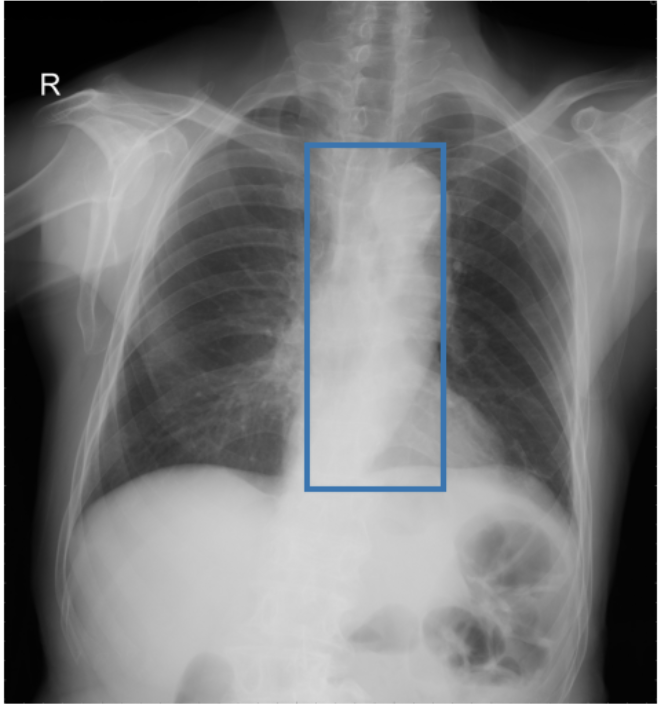} & 
        \includegraphics[width=\linewidth]{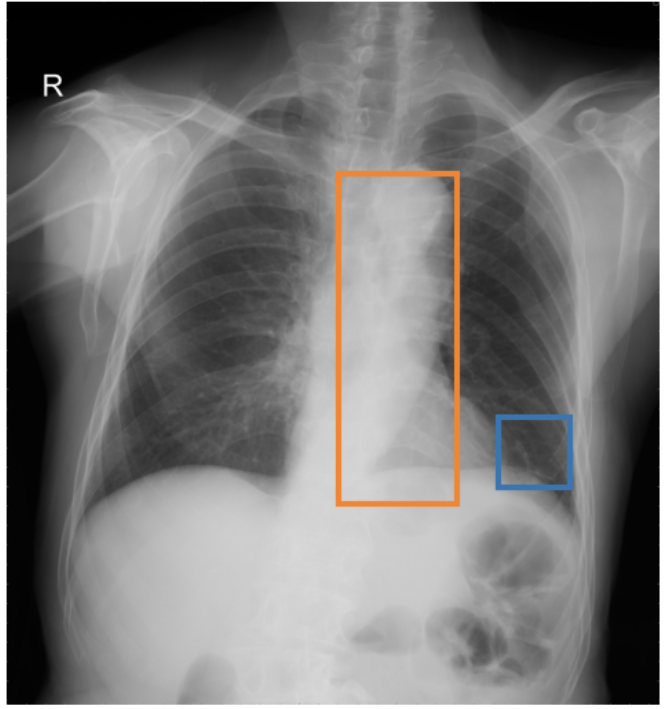} \\
        
        \textbf{(d) Ground Truth} & \textbf{(e) MAIRA-2} & \textbf{(f) CURE} \\
        
        \scriptsize \raggedright
        Aortic enlargement \textcolor[HTML]{1F77B4}{\textbf{[0.55,0.23,0.69,0.39]}}. Calcification \textcolor[HTML]{FF7F0E}{\textbf{[0.63,0.29,0.67,0.33]}} \textcolor[HTML]{2CA02C}{\textbf{[0.63,0.24,0.65,0.26]}}. Rib fracture \textcolor[HTML]{D62728}{\textbf{[0.77,0.40,0.82,0.45]}} \textcolor[HTML]{9467BD}{\textbf{[0.78,0.33,0.82,0.38]}}. Other diseases.
        & 
        \scriptsize \raggedright
        Normal cardiac silhouette. The lung fields are clear. No evidence of hilar adenopathy. No evidence of mediastinal adenopathy. Skeletal structures are unremarkable. The aorta is tortuous. \textcolor[HTML]{1F77B4}{\textbf{[0.465, 0.205, 0.675, 0.695]}}
        \vspace{0.3em} \\
        \textbf{Metrics:} \\
        \renewcommand{\arraystretch}{1.0}
        \begin{tabular}{@{}ll@{}}
        IoU: & 0.182 \\
        CheXbert Cos: & 0.738 \\
        CheXbert Acc: & \textbf{0.786} \\
        CXRFEScore: & 0.316 \\
        \end{tabular}
        & 
        \scriptsize \raggedright
        Left basal laminar atelectasis \textcolor[HTML]{1F77B4}{\textbf{[0.81,0.64,0.11,0.10]}}. Aortic elongation \textcolor[HTML]{FF7F0E}{\textbf{[0.60,0.48,0.18,0.47]}}. No other relevant findings
        \vspace{0.3em} \\
        \textbf{Metrics:} \\
        \renewcommand{\arraystretch}{1.0}
        \begin{tabular}{@{}ll@{}}
        IoU: & \textbf{0.199} \\
        CheXbert Cos: & \textbf{0.755} \\
        CheXbert Acc: & \textbf{0.786} \\
        CXRFEScore: & \textbf{0.350} \\
        \end{tabular}
    \end{tabular}%
    } 

    \caption{Qualitative comparison of two examples of grounded report generation (GRG) from the \textbf{VinDr-CXR} dataset. The \textbf{top row (a-c)} shows the first example, and the \textbf{bottom row (d-f)} shows the second example. For each, the left column represents ground-truth annotations, the middle is MAIRA-2, and the right is CURE. The colored coordinates in the text correspond to bounding boxes drawn in the images. Per-sample metrics are provided below the predicted reports.}
    \label{fig:vindr_grg_qualitative_examples}
\end{figure*}

\textbf{Grounded Report Generation on VinDr-CXR.} Figure~\ref{fig:vindr_grg_qualitative_examples} presents two additional test samples from the VinDr-CXR dataset. These examples highlight the difficulty of generating reports for complex cases with dense annotations. In these specific instances, both models capture salient clinical features, though challenges remain in fully recovering all localized findings present in the ground truth. Quantitative metrics for these samples remain comparable, with CURE showing a slight advantage in semantic similarity (CheXbert Cosine) and localization (IoU) in both examples.

\begin{figure*}[t]
    \centering
    \setlength{\tabcolsep}{3pt}
    \renewcommand{\arraystretch}{1.2}
    
    \newcommand{\SideBySideBlock}[4]{%
        \begin{minipage}[t]{0.32\textwidth}
            \textbf{#2} \\ 
            \begin{minipage}[t]{0.45\linewidth}
                \vspace{0pt} 
                \includegraphics[width=\linewidth]{#1}
            \end{minipage}%
            \hfill%
            \begin{minipage}[t]{0.52\linewidth}
                \vspace{0pt} 
                \scriptsize \raggedright
                #3
                
                \if\relax\detokenize{#4}\relax
                \else
                    \vspace{0.3em}
                    \textbf{Metrics:} \\
                    \renewcommand{\arraystretch}{0.9}
                    \begin{tabular}{@{}l@{\hspace{3pt}}l@{}}
                    #4
                    \end{tabular}
                \fi
            \end{minipage}
        \end{minipage}%
    }

    \resizebox{\textwidth}{!}{%
        \SideBySideBlock{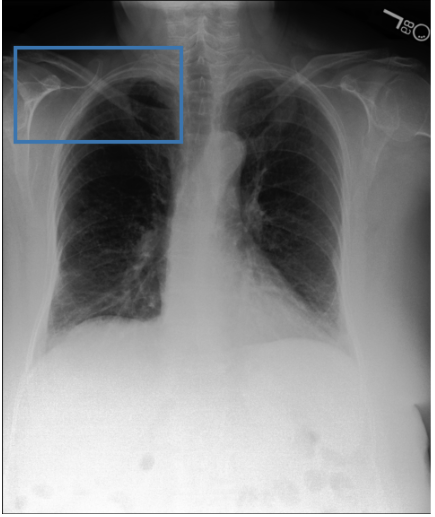}{(a) Ground Truth}{%
            \textbf{Loc: Right Clavicle} \\
            Osseous structures are diffusely demineralized. \textcolor[HTML]{1F77B4}{\textbf{[0.03, 0.09, 0.42, 0.28]}}
        }{}
        \SideBySideBlock{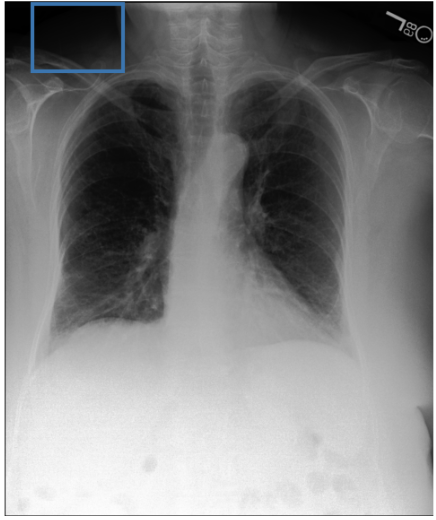}{(b) MAIRA-2}{%
            Right clavicle is fractured. \textcolor[HTML]{1F77B4}{\textbf{[0.07, 0.01, 0.28, 0.14]}}
        }{%
            IoU: & 0.097 \\
            CheXbert Cos: & 0.459 \\
            CheXbert Acc: & 0.857 \\
            CXRFEScore: & 0.119 \\
        }
        \SideBySideBlock{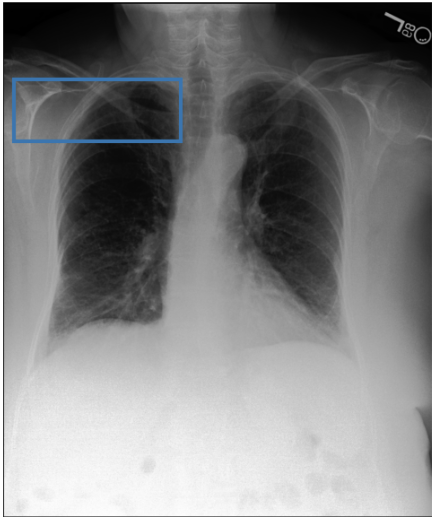}{(c) CURE}{%
            Location of the right clavicle: \textcolor[HTML]{1F77B4}{\textbf{[0.22,0.21,0.39,0.12]}}. Description: No acute osseous abnormalities.
        }{%
            IoU: & \textbf{0.643} \\
            CheXbert Cos: & \textbf{0.882} \\
            CheXbert Acc: & \textbf{1.000} \\
            CXRFEScore: & \textbf{0.513} \\
        }
    }

    \vspace{0.15cm}
    \hrule
    \vspace{0.15cm}

    \resizebox{\textwidth}{!}{%
        \SideBySideBlock{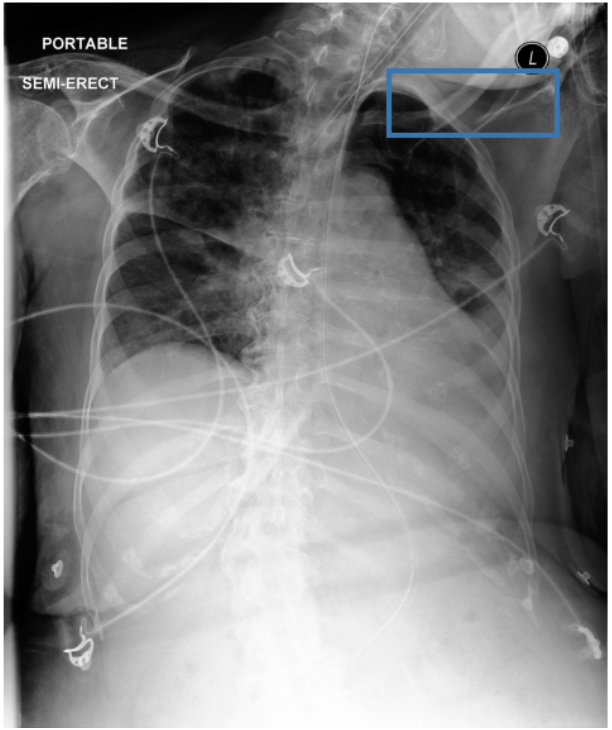}{(d) Ground Truth}{%
            \textbf{Loc: Left Clavicle} \\
            ETT tip at clavicle margin. No bone abnormalities. \textcolor[HTML]{1F77B4}{\textbf{[0.64, 0.09, 0.92, 0.18]}}
        }{}
        \SideBySideBlock{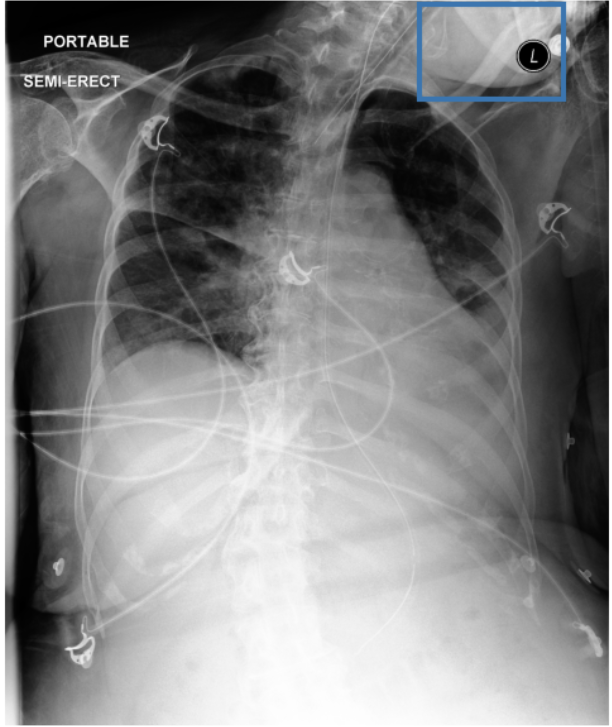}{(e) MAIRA-2}{%
            Left clavicle fracture. \textcolor[HTML]{1F77B4}{\textbf{[0.69, 0.01, 0.93, 0.14]}}
        }{%
            IoU: & 0.206 \\
            CheXbert Cos: & 0.303 \\
            CheXbert Acc: & 0.786 \\
            CXRFEScore: & 0.404 \\
        }
        \SideBySideBlock{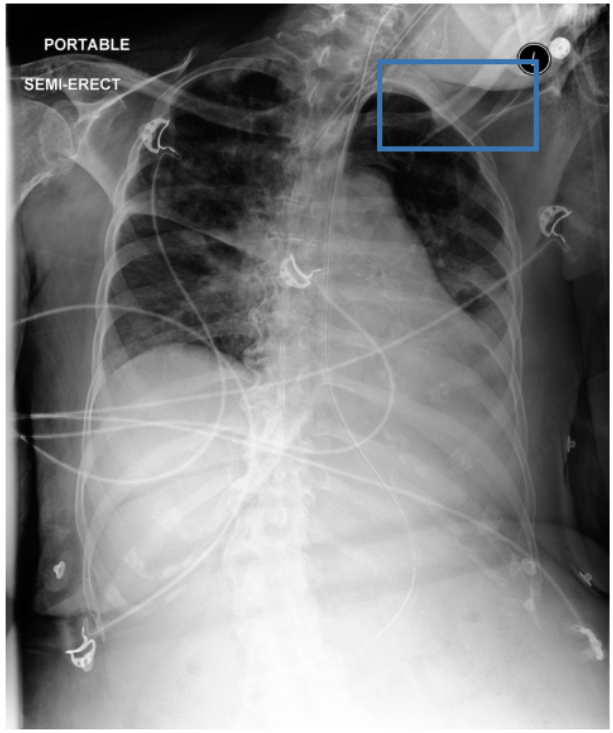}{(f) CURE}{%
            Location of the left clavicle: \textcolor[HTML]{1F77B4}{\textbf{[0.75,0.14,0.26,0.12]}}. Description: The patient is intubated, the ETT terminating in the trachea 4 cm above the level of the carina.
        }{%
            IoU: & \textbf{0.627} \\
            CheXbert Cos: & \textbf{0.880} \\
            CheXbert Acc: & \textbf{1.000} \\
            CXRFEScore: & \textbf{0.467} \\
        }
    }

    \vspace{0.15cm}
    \hrule
    \vspace{0.15cm}

    \resizebox{\textwidth}{!}{%
        \SideBySideBlock{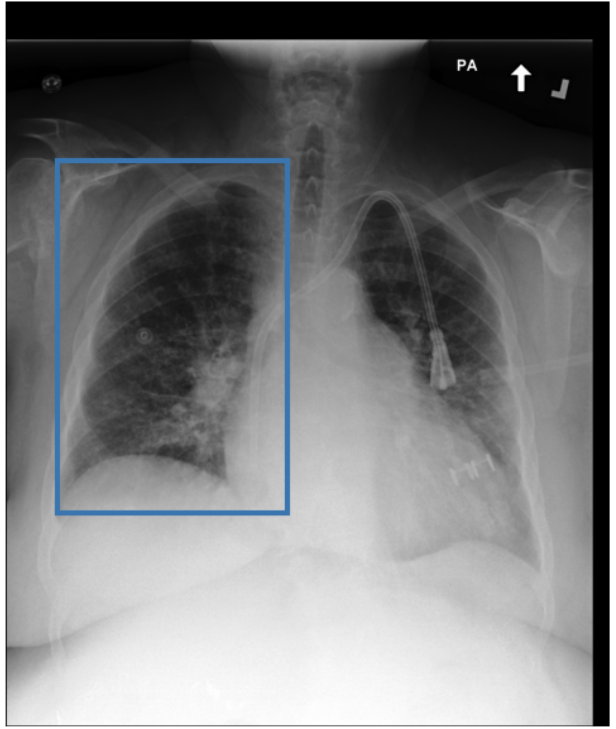}{(g) Ground Truth}{%
            \textbf{Loc: Right Lung} \\
            There is a diffuse mild interstitial abnormality in the right lung, unchanged from prior. There is no evidence of consolidation or edema. There is no pleural effusion or pneumothorax. There is evidence of stable pulmonary hypertension and vascular engorgement. \textcolor[HTML]{1F77B4}{\textbf{[0.086, 0.219, 0.466, 0.705]}}
        }{}
        \SideBySideBlock{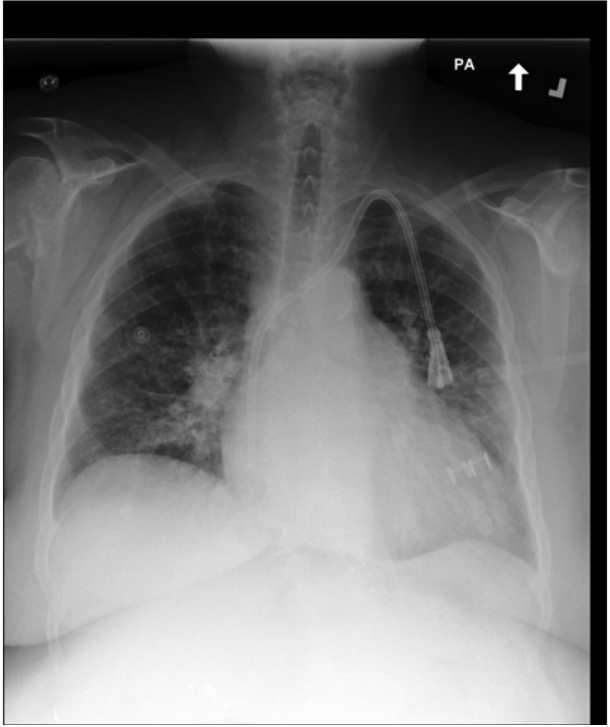}{(h) MAIRA-2}{%
            Right lung is clear.
        }{%
            IoU: & 0.000 \\
            CheXbert Cos: & \textbf{0.716} \\
            CheXbert Acc: & \textbf{0.929} \\
            CXRFEScore: & 0.530 \\
        }
        \SideBySideBlock{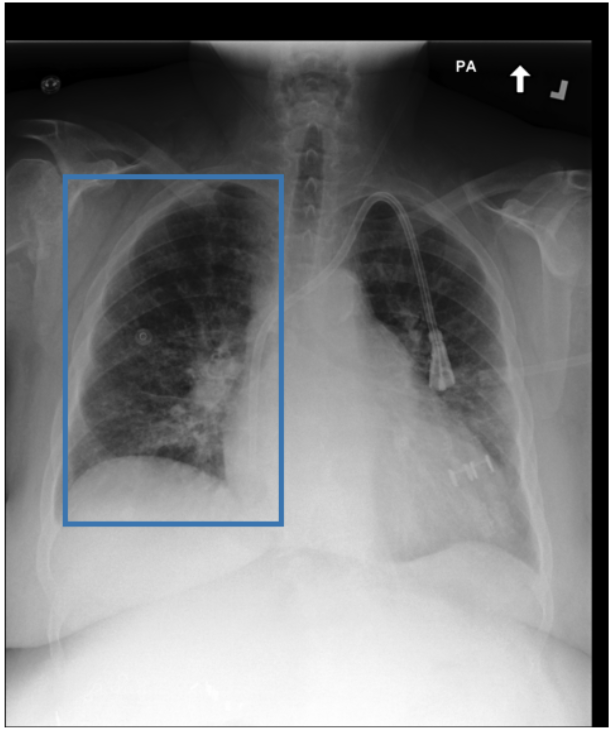}{(i) CURE}{%
            Location of the right lung: \textcolor[HTML]{1F77B4}{\textbf{[0.28,0.48,0.36,0.48]}}. Description: There is mild pulmonary vascular congestion. No pleural effusion or pneumothorax is seen. Mild pulmonary vascular congestion.
        }{%
            IoU: & \textbf{0.879} \\
            CheXbert Cos: & 0.699 \\
            CheXbert Acc: & \textbf{0.929} \\
            CXRFEScore: & \textbf{0.805} \\
        }
    }

    \vspace{0.15cm}
    \hrule
    \vspace{0.15cm}

    \resizebox{\textwidth}{!}{%
        \SideBySideBlock{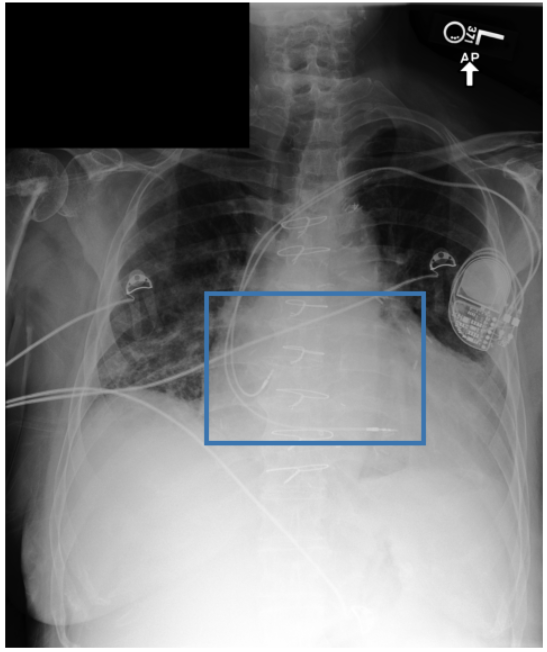}{(j) Ground Truth}{%
            \textbf{Loc: Cardiac Silhouette} \\
            Moderate cardiomegaly is unchanged. A left-sided dual-chamber pacemaker is present with leads in the right atrium and right ventricle. \textcolor[HTML]{1F77B4}{\textbf{[0.375, 0.451, 0.778, 0.683]}}
        }{}
        \SideBySideBlock{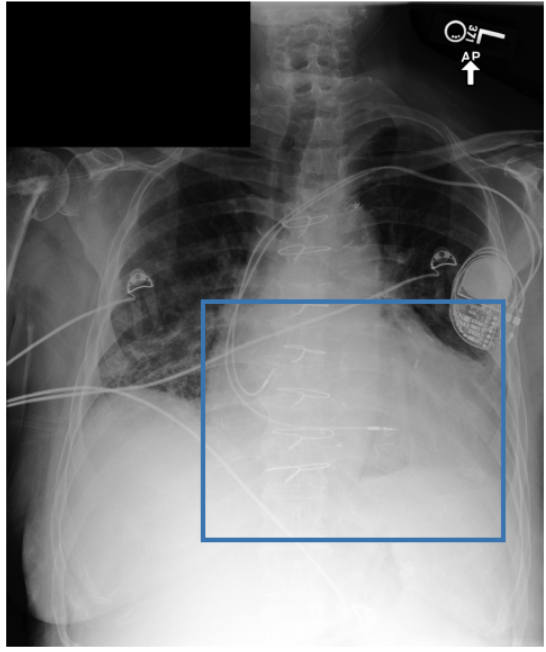}{(k) MAIRA-2}{%
            cardiac silhouette. \textcolor[HTML]{1F77B4}{\textbf{[0.37, 0.47, 0.93, 0.84]}}
        }{%
            IoU: & 0.411 \\
            CheXbert Cos: & 0.723 \\
            CheXbert Acc: & \textbf{0.929} \\
            CXRFEScore: & 0.515 \\
        }
        \SideBySideBlock{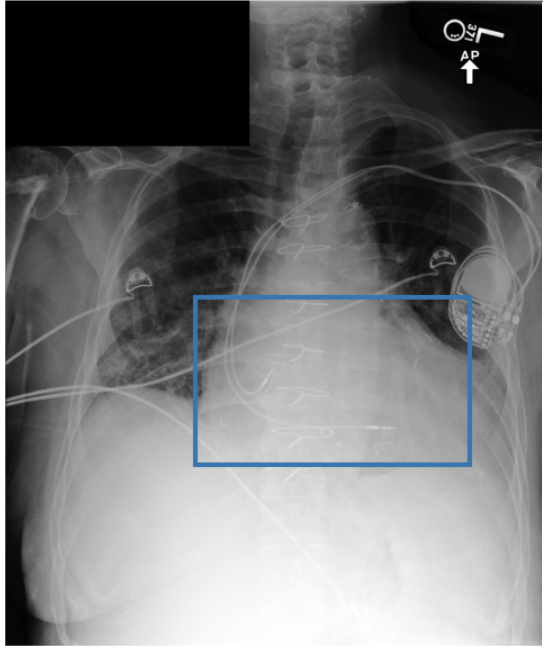}{(l) CURE}{%
            Location of the cardiac silhouette: \textcolor[HTML]{1F77B4}{\textbf{[0.61,0.59,0.51,0.26]}}. Description: FINDINGS: In comparison with the study of \_\_\_, there is continued enlargement of the cardiac silhouette with a dual-channel pacer device in place.
        }{%
            IoU: & \textbf{0.657} \\
            CheXbert Cos: & \textbf{0.725} \\
            CheXbert Acc: & \textbf{0.929} \\
            CXRFEScore: & \textbf{0.659} \\
        }
    }

    \caption{Qualitative comparison of Anatomy Grounded Report Generation (AGRG) on the \textbf{MIMIC-CXR} test set. We display images and reports side-by-side for compactness. Rows correspond to: \textbf{(a-c)} Right Clavicle, \textbf{(d-f)} Left Clavicle, \textbf{(g-i)} Right Lung, and \textbf{(j-l)} Cardiac Silhouette. The left block is ground-truth, middle is MAIRA-2, and right is CURE. Per-sample metrics are provided below each anatomy-grounded report.}
    \label{fig:mimiccxr_agrg_qualitative_examples}
\end{figure*}

\textbf{Anatomy Grounded Report Generation on MIMIC-CXR.} Figure~\ref{fig:mimiccxr_agrg_qualitative_examples} displays four examples of anatomy-specific generation. In this task, the model is prompted to locate and describe a specific region. The qualitative results highlight differences in robustness against hallucinations. For instance, in the second row (Left Clavicle), MAIRA-2 incorrectly predicts a fracture. In contrast, CURE accurately focuses its description on the placement of the endotracheal tube—which is the primary finding in the ground truth—avoiding the fracture hallucination and achieving significantly better localization (IoU 0.627 vs 0.206).

\clearpage

%% file: tables_hyperparameters.tex
\begin{tabular}{lc}
\toprule
\textbf{Hyperparameter} & \textbf{Value} \\
\midrule
\multicolumn{2}{l}{\textbf{Model \& Training}} \\
Base Model & MedGemma-4B-IT \\
Quantization & 4-bit NF4 \\
Precision & BF16 \\
Optimizer & Fused AdamW \\
Learning Rate & \(2 \times 10^{-4}\) \\
LR Scheduler & Linear \\
Warmup Ratio & 0.03 \\
Batch Size (per device) & 5 \\
Gradient Accum. Steps & 5 \\
Effective Batch Size & 25 \\
Max Grad Norm & 0.3 \\
\midrule
\multicolumn{2}{l}{\textbf{LoRA Configuration}} \\
Rank (\(r\)) & 16 \\
Alpha (\(\alpha\)) & 32 \\
Dropout & 0.05 \\
Target Modules & All linear layers \\
Modules to Save & \texttt{lm\_head, embed\_tokens} \\
\bottomrule
\end{tabular}

%% file: tables_task_summary.tex
\sloppy

\newcolumntype{M}[1]{>{\raggedright\arraybackslash}m{#1}}

\renewcommand{\arraystretch}{1.25}

\begin{small}
\begin{tabularx}{\linewidth}{
  @{}
  M{1.6cm} 
  M{1.0cm} 
  M{2.3cm} 
  X        
  @{}
}

\toprule
\textbf{Dataset} & \textbf{Task} & \textbf{Prompt} & \textbf{Example Output} \\
\midrule

MS-CXR &
PG &
\texttt{Ground the phrase: \textit{\{phrase\}}} &
\texttt{\textit{\{phrase\}}: [cx$_1$,cy$_1$,w$_1$,h$_1$] [cx$_2$,cy$_2$,w$_2$,h$_2$] ...} \\
\midrule

\multirow{2}{*}{PadChest-GR} &
PG &
\texttt{Ground the phrase: \textit{\{phrase\}}} &
\texttt{\textit{\{phrase\}}: [cx$_1$,cy$_1$,w$_1$,h$_1$] [cx$_2$,cy$_2$,w$_2$,h$_2$] ...} \\
\cmidrule(l){2-4}

& GRG &
\texttt{Generate a grounded report.} &
\texttt{Slight residual atelectasis in the right pulmonary base [0.29,0.66,0.18,0.20]. Minimal blunting of the costophrenic angle [0.81,0.74,0.33,0.39].} \\
\midrule

\multirow{3}{2cm}{Chest ImaGenome} &
\multirow{3}{*}{AGRG} &
\texttt{Locate and describe the \textit{\{location\}}.} &
\texttt{Location of the abdomen: [0.48,0.78,0.73,0.45]. Description: No free air below the right hemidiaphragm is seen.} \\
\cmidrule(l){3-4}

& &
\texttt{Locate the \textit{\{location\}}.} &
\texttt{Location of the right cardiophrenic angle: [0.33,0.71,0.09,0.12].} \\
\cmidrule(l){3-4}

& &
\texttt{Describe the \textit{\{location\}}.} &
\texttt{Description of the left chest wall: Left chest wall pacer defibrillator is unchanged in position.} \\
\midrule

\multirow{2}{2cm}{VinDr-CXR (Eval., zero-shot)} &
PG &
\texttt{Ground the phrase: \textit{\{phrase\}}} &
\texttt{Cardiomegaly: [0.57,0.65,0.55,0.37]} \\
\cmidrule(l){2-4}

& GRG &
\texttt{Generate a grounded report.} &
\texttt{Consolidation [0.29,0.54,0.18,0.13]. Infiltration [0.27,0.49,0.27,0.24]. Pleural effusion [0.82,0.72,0.25,0.14].} \\
\midrule

\multirow{3}{2cm}{MIMIC-CXR (Eval.)} &
RG via GRG &
\texttt{Generate a grounded report.} &
\textit{\{Grounded report\}} \\
\cmidrule(l){2-4}

& RG via AGRG &
\texttt{Locate and describe the \textit{\{location\}}. ($\times N$ anatomical locations)} &
\textit{\{Report for location 1\}} ... \textit{\{Report for location N\}} \\
\cmidrule(l){2-4}

& RG via AGRG + GRG &
\texttt{Combine AGRG and GRG generations.} &
\textit{\{Report for location 1\}} ... \textit{\{Report for location N\}} \textit{\{GRG report\}} \\

\bottomrule
\end{tabularx}
\end{small}

\renewcommand{\arraystretch}{1.0}

%% file: tables_suppl_configs.tex
\begin{tabularx}{\textwidth}{@{} >{\raggedright\arraybackslash}p{5.8cm} X @{}}
\toprule
\textbf{Model Configuration} & \textbf{Description} \\
\midrule
\multicolumn{2}{@{}l}{\textit{--- Baseline \& Augmentation ---}} \\
\textbf{v1}: Base (w/o Aug, w/o CL, w/o CIG, lr=2e-5) & \textbf{Baseline:} Basic multi-task fine-tuning using uniform sampling across all datasets (inter-dataset) and within datasets (intra-dataset). No data augmentation or pre-training is applied. Fine-tuned with a base learning rate of 2e-5 for 6k steps. \\
\textbf{v2}: + Aug & Identical to \textbf{v1}, but enables \textbf{bounding-box-aware} augmentations (stochastic CLAHE, RandomResizedCrop, and affine transforms). Ground-truth text coordinates are dynamically updated to match spatial changes. Horizontal flipping and color distortions are disabled. \\
\midrule
\multicolumn{2}{@{}l}{\textit{--- Curriculum Learning (CL) Frequency ---}} \\
\textbf{v3}: + Aug + CL(1.5k) & Extends \textbf{v2} by introducing curriculum learning (CL). The sampling distribution is re-weighted based on model performance every \textbf{1500 steps}. \\
\textbf{v4}: + Aug + CL(2k) & Same as \textbf{v3}, but the curriculum re-weighting interval is increased to every \textbf{2000 steps}. \\
\textbf{v5}: + Aug + CL(3k) & Same as \textbf{v3}, but the curriculum re-weighting interval is set to every \textbf{3000 steps}. This serves as the foundational CL schedule for subsequent experiments. \\
\midrule
\multicolumn{2}{@{}l}{\textit{--- CIG Pre-training Integration (Low LR: 2e-5) ---}} \\
\textbf{v6}: + Aug + CIG(1k) + CL(3k) & Introduces a pre-training phase on the Chest ImaGenome (CIG) dataset for \textbf{1000 steps} (lr=2e-5) before initializing the multi-task fine-tuning configuration of \textbf{v5}. \\
\textbf{v7}: + Aug + CIG(2k) + CL(3k) & Extends the CIG pre-training phase to \textbf{2000 steps} (lr=2e-5) before fine-tuning. \\
\textbf{v8}: + Aug + CIG(3k) + CL(3k) & Extends the CIG pre-training phase to \textbf{3000 steps} (lr=2e-5) before fine-tuning. \\
\midrule
\multicolumn{2}{@{}l}{\textit{--- Learning Rate Scaling (High LR: 2e-4) ---}} \\
\textbf{v9}: + Aug + CIG(1k) + CL(3k) + lr=2e-4 & Replicates the structure of \textbf{v6} (1k pre-train), but significantly increases the learning rate to \textbf{2e-4} for both pre-training and fine-tuning stages. \\
\textbf{v10}: + Aug + CIG(2k) + CL(3k) + lr=2e-4 & Replicates the structure of \textbf{v7} (2k pre-train) with the higher learning rate of \textbf{2e-4}. \\
\textbf{v11 (CURE)}: + Aug + CIG(3k) + CL(3k) + lr=2e-4 & \textbf{Proposed Method (CURE):} Replicates \textbf{v8} (3k pre-train) with the higher learning rate (2e-4). Combines prolonged pre-training, high learning rate, and 3k-step curriculum updates. \\
\midrule
\multicolumn{2}{@{}l}{\textit{--- Sampling Strategy Ablations (Based on v11) ---}} \\
\textbf{v12}: + Aug + CIG(3k) + Uni(Inter)/Nat(Intra) + lr=2e-4 & Modification of \textbf{v11} that removes Curriculum Learning entirely. Uses \textbf{Uniform} sampling between datasets (Inter) and \textbf{Natural} distribution sampling within datasets (Intra). \\
\textbf{v13}: + Aug + CIG(3k) + CL(Inter,3k)/Nat(Intra) + lr=2e-4 & Modification of \textbf{v11} that applies Curriculum Learning (3k re-weighting) \textbf{only} to the inter-dataset sampling ratios, while maintaining a \textbf{Natural} distribution for intra-dataset sampling. \\
\textbf{v14}: + Aug + CIG(3k) + Uni(Inter)/CL(Intra,3k) + lr=2e-4 & Modification of \textbf{v11} that applies \textbf{Uniform} sampling between datasets (Inter), while applying \textbf{Curriculum Learning} (3k re-weighting) exclusively to intra-dataset sampling. \\
\textbf{v15}: + Aug + CIG(3k) + Nat(Inter)/Nat(Intra) + lr=2e-4 & Modification of \textbf{v11} using a fully \textbf{Natural} sampling strategy (proportional to dataset size) for both inter-dataset and intra-dataset distributions. \\
\bottomrule
\end{tabularx}

%% file: tables_suppl_ablations.tex
\resizebox{\textwidth}{!}{%
\begin{tabular}{@{}l@{\hspace{1mm}}c@{\hspace{1mm}}c@{\hspace{1mm}}c@{\hspace{1mm}}c@{\hspace{1mm}}c@{\hspace{1mm}}c@{\hspace{1mm}}c@{\hspace{1mm}}c@{\hspace{1mm}}c@{\hspace{1mm}}c@{\hspace{1mm}}c@{\hspace{1mm}}c@{}}
\toprule
\multicolumn{1}{c}{\multirow{2}{*}{\textbf{Model Configuration}}} & \multicolumn{3}{c}{\textbf{AGRG (CIG)}} & \multicolumn{3}{c}{\textbf{GRG (PC)}} & \multicolumn{3}{c}{\textbf{GRG (VD)}} & \multicolumn{3}{c}{\textbf{PG (IoU $\uparrow$)}} \\
\cmidrule(l){2-4} \cmidrule(l){5-7} \cmidrule(l){8-10} \cmidrule(l){11-13}
& IoU $\uparrow$ & F1 $\uparrow$ & CXS $\uparrow$ & IoU $\uparrow$ & F1 $\uparrow$ & CXS $\uparrow$ & IoU $\uparrow$ & F1 $\uparrow$ & CXS $\uparrow$ & MS & PC & VD \\ 
\midrule
MAIRA-2 (External Baseline) & 0.249 $\pm$ 0.008 & 0.377 $\pm$ 0.016 & 0.357 $\pm$ 0.010 & 0.256 $\pm$ 0.011 & \textbf{0.591} $\pm$ 0.015 & \textbf{0.616} $\pm$ 0.011 & 0.217 $\pm$ 0.007 & 0.546 $\pm$ 0.008 & 0.591 $\pm$ 0.005 & 0.495 $\pm$ 0.016 & 0.280 $\pm$ 0.008 & 0.161 $\pm$ 0.005 \\
\midrule
v1: Base (w/o Aug, w/o CL, w/o CIG, lr=2e-5) & 0.380 $\pm$ 0.008 & 0.517 $\pm$ 0.017 & 0.517 $\pm$ 0.011 & 0.171 $\pm$ 0.009 & 0.557 $\pm$ 0.015 & 0.589 $\pm$ 0.011 & 0.207 $\pm$ 0.006 & 0.586 $\pm$ 0.008 & 0.630 $\pm$ 0.007 & 0.388 $\pm$ 0.017 & 0.356 $\pm$ 0.007 & 0.191 $\pm$ 0.004 \\
v2: + Aug & 0.360 $\pm$ 0.009 & 0.500 $\pm$ 0.018 & 0.522 $\pm$ 0.011 & 0.185 $\pm$ 0.010 & 0.564 $\pm$ 0.015 & \underline{0.599} $\pm$ 0.011 & 0.221 $\pm$ 0.007 & \underline{0.614} $\pm$ 0.008 & 0.648 $\pm$ 0.007 & 0.398 $\pm$ 0.019 & 0.366 $\pm$ 0.007 & 0.203 $\pm$ 0.005 \\
\midrule
v3: + Aug + CL(1.5k) & 0.399 $\pm$ 0.009 & 0.487 $\pm$ 0.018 & 0.521 $\pm$ 0.011 & 0.179 $\pm$ 0.010 & 0.564 $\pm$ 0.016 & 0.592 $\pm$ 0.011 & 0.224 $\pm$ 0.007 & 0.605 $\pm$ 0.008 & 0.630 $\pm$ 0.007 & 0.409 $\pm$ 0.019 & 0.383 $\pm$ 0.007 & 0.210 $\pm$ 0.005 \\
v4: + Aug + CL(2k) & 0.394 $\pm$ 0.009 & 0.504 $\pm$ 0.017 & 0.513 $\pm$ 0.011 & 0.193 $\pm$ 0.010 & 0.568 $\pm$ 0.015 & 0.596 $\pm$ 0.011 & 0.222 $\pm$ 0.006 & 0.611 $\pm$ 0.008 & \underline{0.651} $\pm$ 0.007 & 0.393 $\pm$ 0.017 & 0.383 $\pm$ 0.007 & 0.196 $\pm$ 0.005 \\
v5: + Aug + CL(3k) & 0.411 $\pm$ 0.009 & 0.493 $\pm$ 0.017 & 0.526 $\pm$ 0.011 & 0.180 $\pm$ 0.010 & \underline{0.578} $\pm$ 0.014 & 0.595 $\pm$ 0.012 & 0.217 $\pm$ 0.007 & \textbf{0.626} $\pm$ 0.008 & \textbf{0.671} $\pm$ 0.007 & 0.430 $\pm$ 0.018 & 0.393 $\pm$ 0.007 & 0.205 $\pm$ 0.005 \\
\midrule
v6: + Aug + CIG(1k) + CL(3k) & 0.454 $\pm$ 0.008 & 0.512 $\pm$ 0.017 & 0.518 $\pm$ 0.011 & 0.195 $\pm$ 0.009 & 0.553 $\pm$ 0.015 & 0.591 $\pm$ 0.011 & 0.232 $\pm$ 0.006 & 0.582 $\pm$ 0.008 & 0.628 $\pm$ 0.007 & 0.457 $\pm$ 0.016 & 0.394 $\pm$ 0.007 & 0.219 $\pm$ 0.005 \\
v7: + Aug + CIG(2k) + CL(3k) & 0.448 $\pm$ 0.008 & \underline{0.532} $\pm$ 0.017 & 0.533 $\pm$ 0.011 & 0.203 $\pm$ 0.010 & 0.535 $\pm$ 0.015 & 0.586 $\pm$ 0.011 & 0.227 $\pm$ 0.006 & 0.553 $\pm$ 0.008 & 0.590 $\pm$ 0.007 & 0.467 $\pm$ 0.016 & 0.403 $\pm$ 0.007 & 0.222 $\pm$ 0.005 \\
v8: + Aug + CIG(3k) + CL(3k) & 0.486 $\pm$ 0.008 & 0.530 $\pm$ 0.017 & 0.521 $\pm$ 0.011 & 0.207 $\pm$ 0.010 & 0.552 $\pm$ 0.015 & 0.582 $\pm$ 0.011 & 0.233 $\pm$ 0.006 & 0.568 $\pm$ 0.008 & 0.601 $\pm$ 0.007 & 0.495 $\pm$ 0.016 & 0.421 $\pm$ 0.007 & 0.224 $\pm$ 0.005 \\
\midrule
v9: + Aug + CIG(1k) + CL(3k) + lr=2e-4 & 0.607 $\pm$ 0.008 & 0.517 $\pm$ 0.017 & 0.531 $\pm$ 0.011 & 0.253 $\pm$ 0.010 & 0.522 $\pm$ 0.016 & 0.563 $\pm$ 0.010 & 0.259 $\pm$ 0.007 & 0.491 $\pm$ 0.008 & 0.529 $\pm$ 0.006 & 0.564 $\pm$ 0.016 & 0.453 $\pm$ 0.007 & \underline{0.247} $\pm$ 0.005 \\
v10: + Aug + CIG(2k) + CL(3k) + lr=2e-4 & 0.606 $\pm$ 0.008 & 0.515 $\pm$ 0.017 & 0.535 $\pm$ 0.011 & 0.258 $\pm$ 0.010 & 0.523 $\pm$ 0.015 & 0.569 $\pm$ 0.010 & 0.262 $\pm$ 0.007 & 0.449 $\pm$ 0.008 & 0.477 $\pm$ 0.006 & \textbf{0.574} $\pm$ 0.015 & \underline{0.457} $\pm$ 0.007 & \textbf{0.248} $\pm$ 0.005 \\
\textbf{v11 (CURE)}: + Aug + CIG(3k) + CL(3k) + lr=2e-4 & 0.601 $\pm$ 0.008 & 0.529 $\pm$ 0.017 & \underline{0.549} $\pm$ 0.011 & \underline{0.265} $\pm$ 0.011 & 0.507 $\pm$ 0.015 & 0.574 $\pm$ 0.010 & 0.262 $\pm$ 0.007 & 0.505 $\pm$ 0.008 & 0.540 $\pm$ 0.007 & 0.552 $\pm$ 0.015 & 0.453 $\pm$ 0.006 & 0.243 $\pm$ 0.005 \\
\midrule
v12: + Aug + CIG(3k) + Uni(Inter)/Nat(Intra) + lr=2e-4 & 0.595 $\pm$ 0.008 & 0.509 $\pm$ 0.017 & 0.526 $\pm$ 0.011 & 0.263 $\pm$ 0.010 & 0.529 $\pm$ 0.015 & 0.575 $\pm$ 0.010 & \underline{0.264} $\pm$ 0.007 & 0.489 $\pm$ 0.008 & 0.514 $\pm$ 0.007 & 0.557 $\pm$ 0.015 & \underline{0.457} $\pm$ 0.007 & 0.245 $\pm$ 0.005 \\
v13: + Aug + CIG(3k) + CL(Inter,3k)/Nat(Intra) + lr=2e-4 & \underline{0.612} $\pm$ 0.008 & \textbf{0.533} $\pm$ 0.017 & 0.529 $\pm$ 0.011 & \textbf{0.272} $\pm$ 0.010 & 0.503 $\pm$ 0.016 & 0.550 $\pm$ 0.010 & \textbf{0.266} $\pm$ 0.007 & 0.434 $\pm$ 0.008 & 0.469 $\pm$ 0.006 & \underline{0.567} $\pm$ 0.016 & \textbf{0.464} $\pm$ 0.007 & 0.243 $\pm$ 0.005 \\
v14: + Aug + CIG(3k) + Uni(Inter)/CL(Intra,3k) + lr=2e-4 & 0.603 $\pm$ 0.008 & 0.525 $\pm$ 0.017 & 0.547 $\pm$ 0.011 & 0.246 $\pm$ 0.010 & 0.544 $\pm$ 0.015 & 0.570 $\pm$ 0.010 & 0.262 $\pm$ 0.007 & 0.496 $\pm$ 0.008 & 0.536 $\pm$ 0.007 & 0.555 $\pm$ 0.015 & 0.456 $\pm$ 0.007 & 0.244 $\pm$ 0.005 \\
v15: + Aug + CIG(3k) + Nat(Inter)/Nat(Intra) + lr=2e-4 & \textbf{0.639} $\pm$ 0.008 & 0.530 $\pm$ 0.017 & \textbf{0.554} $\pm$ 0.011 & 0.000 $\pm$ 0.000 & 0.498 $\pm$ 0.016 & 0.483 $\pm$ 0.011 & 0.000 $\pm$ 0.000 & 0.604 $\pm$ 0.009 & 0.534 $\pm$ 0.006 & 0.355 $\pm$ 0.019 & 0.221 $\pm$ 0.006 & 0.162 $\pm$ 0.005 \\
\bottomrule
\end{tabular}%
}

%% file: tables_suppl_agrg_results.tex
\resizebox{\textwidth}{!}{%
\begin{tabular}{@{}lccccc@{}}
\toprule
\textbf{Model Variant} &
\textbf{IoU} $\uparrow$ &
\textbf{F1-Mi} $\uparrow$ &
\textbf{F1-Ma} $\uparrow$ &
\textbf{Cos.} $\uparrow$ &
\textbf{CXS} $\uparrow$ \\ 
\midrule
\multicolumn{6}{@{}l}{\textit{--- Baselines ---}} \\
MAIRA-2 & 0.249 $\pm$ 0.008 & 0.377 $\pm$ 0.016 & 0.098 $\pm$ 0.009 & 0.587 $\pm$ 0.010 & 0.357 $\pm$ 0.010 \\
MedGemma-4B-IT & -- & 0.266 $\pm$ 0.012 & 0.227 $\pm$ 0.014 & 0.662 $\pm$ 0.004 & 0.467 $\pm$ 0.006 \\
\midrule
\multicolumn{6}{@{}l}{\textit{--- Chest ImaGenome (CIG) Pre-training Only ---}} \\
CIG Pre-train (1k steps, lr=2e-5) & 0.378 $\pm$ 0.008 & 0.530 $\pm$ 0.017 & 0.249 $\pm$ 0.023 & 0.670 $\pm$ 0.009 & 0.510 $\pm$ 0.012 \\
CIG Pre-train (2k steps, lr=2e-5) & 0.402 $\pm$ 0.008 & 0.528 $\pm$ 0.016 & 0.215 $\pm$ 0.015 & 0.666 $\pm$ 0.009 & 0.512 $\pm$ 0.011 \\
CIG Pre-train (3k steps, lr=2e-5) & 0.430 $\pm$ 0.008 & 0.526 $\pm$ 0.017 & 0.221 $\pm$ 0.015 & 0.672 $\pm$ 0.009 & 0.513 $\pm$ 0.012 \\
CIG Pre-train (1k steps, lr=2e-4) & 0.501 $\pm$ 0.008 & 0.525 $\pm$ 0.017 & 0.237 $\pm$ 0.017 & 0.675 $\pm$ 0.009 & 0.545 $\pm$ 0.011 \\
CIG Pre-train (2k steps, lr=2e-4) & 0.590 $\pm$ 0.008 & 0.525 $\pm$ 0.017 & 0.242 $\pm$ 0.018 & 0.686 $\pm$ 0.009 & 0.544 $\pm$ 0.011 \\
CIG Pre-train (3k steps, lr=2e-4) & 0.596 $\pm$ 0.008 & 0.524 $\pm$ 0.017 & 0.235 $\pm$ 0.017 & 0.688 $\pm$ 0.009 & \underline{0.551} $\pm$ 0.011 \\
\midrule
\multicolumn{6}{@{}l}{\textit{--- Multi-task Fine-tuning Variants (v1--v15) ---}} \\
\textbf{v1}: Base (w/o Aug, w/o CL, w/o CIG, lr=2e-5) & 0.380 $\pm$ 0.008 & 0.517 $\pm$ 0.017 & \textbf{0.251} $\pm$ 0.017 & 0.665 $\pm$ 0.009 & 0.517 $\pm$ 0.011 \\
\textbf{v2}: + Aug & 0.360 $\pm$ 0.009 & 0.500 $\pm$ 0.018 & 0.211 $\pm$ 0.015 & 0.660 $\pm$ 0.009 & 0.522 $\pm$ 0.011 \\
\textbf{v3}: + Aug + CL(1.5k) & 0.399 $\pm$ 0.009 & 0.487 $\pm$ 0.018 & 0.207 $\pm$ 0.015 & 0.653 $\pm$ 0.009 & 0.521 $\pm$ 0.011 \\
\textbf{v4}: + Aug + CL(2k) & 0.394 $\pm$ 0.009 & 0.504 $\pm$ 0.017 & 0.212 $\pm$ 0.015 & 0.660 $\pm$ 0.009 & 0.513 $\pm$ 0.011 \\
\textbf{v5}: + Aug + CL(3k) & 0.411 $\pm$ 0.009 & 0.493 $\pm$ 0.017 & 0.197 $\pm$ 0.011 & 0.655 $\pm$ 0.009 & 0.526 $\pm$ 0.011 \\
\textbf{v6}: + Aug + CIG(1k) + CL(3k) & 0.454 $\pm$ 0.008 & 0.512 $\pm$ 0.017 & 0.213 $\pm$ 0.016 & 0.656 $\pm$ 0.009 & 0.518 $\pm$ 0.011 \\
\textbf{v7}: + Aug + CIG(2k) + CL(3k) & 0.448 $\pm$ 0.008 & \underline{0.532} $\pm$ 0.017 & \textbf{0.251} $\pm$ 0.021 & 0.673 $\pm$ 0.009 & 0.533 $\pm$ 0.011 \\
\textbf{v8}: + Aug + CIG(3k) + CL(3k) & 0.486 $\pm$ 0.008 & 0.530 $\pm$ 0.017 & 0.226 $\pm$ 0.016 & 0.671 $\pm$ 0.009 & 0.521 $\pm$ 0.011 \\
\textbf{v9}: + Aug + CIG(1k) + CL(3k) + lr=2e-4 & 0.607 $\pm$ 0.008 & 0.517 $\pm$ 0.017 & 0.223 $\pm$ 0.016 & 0.679 $\pm$ 0.009 & 0.531 $\pm$ 0.011 \\
\textbf{v10}: + Aug + CIG(2k) + CL(3k) + lr=2e-4 & 0.606 $\pm$ 0.008 & 0.515 $\pm$ 0.017 & 0.212 $\pm$ 0.014 & 0.675 $\pm$ 0.009 & 0.535 $\pm$ 0.011 \\
\textbf{v11 (CURE)}: + Aug + CIG(3k) + CL(3k) + lr=2e-4 & 0.601 $\pm$ 0.008 & 0.529 $\pm$ 0.017 & 0.234 $\pm$ 0.018 & \underline{0.691} $\pm$ 0.009 & 0.549 $\pm$ 0.011 \\
\textbf{v12}: + Aug + CIG(3k) + Uni(Inter)/Nat(Intra) + lr=2e-4 & 0.595 $\pm$ 0.008 & 0.509 $\pm$ 0.017 & 0.218 $\pm$ 0.020 & 0.674 $\pm$ 0.009 & 0.526 $\pm$ 0.011 \\
\textbf{v13}: + Aug + CIG(3k) + CL(Inter,3k)/Nat(Intra) + lr=2e-4 & \underline{0.612} $\pm$ 0.008 & \textbf{0.533} $\pm$ 0.017 & 0.249 $\pm$ 0.021 & 0.687 $\pm$ 0.009 & 0.529 $\pm$ 0.011 \\
\textbf{v14}: + Aug + CIG(3k) + Uni(Inter)/CL(Intra,3k) + lr=2e-4 & 0.603 $\pm$ 0.008 & 0.525 $\pm$ 0.017 & 0.227 $\pm$ 0.018 & 0.685 $\pm$ 0.009 & 0.547 $\pm$ 0.011 \\
\textbf{v15}: + Aug + CIG(3k) + Nat(Inter)/Nat(Intra) + lr=2e-4 & \textbf{0.639} $\pm$ 0.008 & 0.530 $\pm$ 0.017 & \underline{0.250} $\pm$ 0.018 & \textbf{0.694} $\pm$ 0.009 & \textbf{0.554} $\pm$ 0.011 \\
\bottomrule
\end{tabular}%
}

%% file: tables_suppl_pg_results.tex
\resizebox{\textwidth}{!}{%
\begin{tabular}{@{}lcccccc@{}}
\toprule
\small
\multirow{2}{*}{\textbf{Model Variant}} &
\multicolumn{2}{c}{\textbf{MS-CXR}} &
\multicolumn{2}{c}{\textbf{PadChest-GR}} &
\multicolumn{2}{c}{\textbf{VinDr-CXR (Zero-Shot)}} \\ 
\cmidrule(l){2-3} \cmidrule(l){4-5} \cmidrule(l){6-7}
& \textbf{IoU Mi.} $\uparrow$ & \textbf{IoU Ma.} $\uparrow$ & \textbf{IoU Mi.} $\uparrow$ & \textbf{IoU Ma.} $\uparrow$ & \textbf{IoU Mi.} $\uparrow$ & \textbf{IoU Ma.} $\uparrow$\\ 
\midrule
MAIRA-2 & 0.495 $\pm$ 0.016 & 0.453 $\pm$ 0.016 & 0.280 $\pm$ 0.008 & 0.288 $\pm$ 0.009 & 0.161 $\pm$ 0.005 & 0.114 $\pm$ 0.010 \\
\midrule

\textbf{v1}: Base (w/o Aug, w/o CL, w/o CIG, lr=2e-5) & 0.388 $\pm$ 0.017 & 0.344 $\pm$ 0.016 & 0.356 $\pm$ 0.007 & 0.345 $\pm$ 0.007 & 0.191 $\pm$ 0.004 & 0.144 $\pm$ 0.010 \\
\textbf{v2}: + Aug & 0.398 $\pm$ 0.019 & 0.353 $\pm$ 0.016 & 0.366 $\pm$ 0.007 & 0.360 $\pm$ 0.007 & 0.203 $\pm$ 0.005 & 0.153 $\pm$ 0.007 \\
\textbf{v3}: + Aug + CL(1.5k) & 0.409 $\pm$ 0.019 & 0.369 $\pm$ 0.016 & 0.383 $\pm$ 0.007 & 0.382 $\pm$ 0.007 & 0.210 $\pm$ 0.005 & 0.154 $\pm$ 0.007 \\
\textbf{v4}: + Aug + CL(2k) & 0.393 $\pm$ 0.017 & 0.348 $\pm$ 0.015 & 0.383 $\pm$ 0.007 & 0.381 $\pm$ 0.008 & 0.196 $\pm$ 0.005 & 0.145 $\pm$ 0.007 \\
\textbf{v5}: + Aug + CL(3k) & 0.430 $\pm$ 0.018 & 0.377 $\pm$ 0.016 & 0.393 $\pm$ 0.007 & 0.397 $\pm$ 0.008 & 0.205 $\pm$ 0.005 & 0.155 $\pm$ 0.006 \\
\textbf{v6}: + Aug + CIG(1k) + CL(3k) & 0.457 $\pm$ 0.016 & 0.405 $\pm$ 0.014 & 0.394 $\pm$ 0.007 & 0.391 $\pm$ 0.008 & 0.219 $\pm$ 0.005 & 0.167 $\pm$ 0.011 \\
\textbf{v7}: + Aug + CIG(2k) + CL(3k) & 0.467 $\pm$ 0.016 & 0.428 $\pm$ 0.015 & 0.403 $\pm$ 0.007 & 0.399 $\pm$ 0.008 & 0.222 $\pm$ 0.005 & 0.160 $\pm$ 0.006 \\
\textbf{v8}: + Aug + CIG(3k) + CL(3k) & 0.495 $\pm$ 0.016 & 0.446 $\pm$ 0.015 & 0.421 $\pm$ 0.007 & 0.419 $\pm$ 0.008 & 0.224 $\pm$ 0.005 & 0.173 $\pm$ 0.011 \\
\textbf{v9}: + Aug + CIG(1k) + CL(3k) + lr=2e-4 & 0.564 $\pm$ 0.016 & 0.514 $\pm$ 0.016 & 0.453 $\pm$ 0.007 & 0.443 $\pm$ 0.007 & \underline{0.247} $\pm$ 0.005 & 0.203 $\pm$ 0.011 \\
\textbf{v10}: + Aug + CIG(2k) + CL(3k) + lr=2e-4 & \textbf{0.574} $\pm$ 0.015 & \textbf{0.526} $\pm$ 0.014 & \underline{0.457} $\pm$ 0.007 & 0.445 $\pm$ 0.008 & \textbf{0.248} $\pm$ 0.005 & \textbf{0.206} $\pm$ 0.011 \\
\textbf{v11 (CURE)}: + Aug + CIG(3k) + CL(3k) + lr=2e-4 & 0.552 $\pm$ 0.015 & 0.495 $\pm$ 0.015 & 0.453 $\pm$ 0.006 & 0.438 $\pm$ 0.007 & 0.243 $\pm$ 0.005 & \underline{0.205} $\pm$ 0.012 \\
\textbf{v12}: + Aug + CIG(3k) + Uni(Inter)/Nat(Intra) + lr=2e-4 & 0.557 $\pm$ 0.015 & 0.507 $\pm$ 0.014 & \underline{0.457} $\pm$ 0.007 & 0.448 $\pm$ 0.007 & 0.245 $\pm$ 0.005 & 0.198 $\pm$ 0.007 \\
\textbf{v13}: + Aug + CIG(3k) + CL(Inter,3k)/Nat(Intra) + lr=2e-4 & \underline{0.567} $\pm$ 0.016 & \underline{0.515} $\pm$ 0.014 & \textbf{0.464} $\pm$ 0.007 & \textbf{0.455} $\pm$ 0.007 & 0.243 $\pm$ 0.005 & 0.199 $\pm$ 0.011 \\
\textbf{v14}: + Aug + CIG(3k) + Uni(Inter)/CL(Intra,3k) + lr=2e-4 & 0.555 $\pm$ 0.015 & 0.496 $\pm$ 0.013 & 0.456 $\pm$ 0.007 & \underline{0.451} $\pm$ 0.008 & 0.244 $\pm$ 0.005 & 0.203 $\pm$ 0.013 \\
\textbf{v15}: + Aug + CIG(3k) + Nat(Inter)/Nat(Intra) + lr=2e-4 & 0.355 $\pm$ 0.019 & 0.277 $\pm$ 0.016 & 0.221 $\pm$ 0.006 & 0.206 $\pm$ 0.005 & 0.162 $\pm$ 0.005 & 0.099 $\pm$ 0.004 \\
\bottomrule
\end{tabular}%
}

%% file: tables_suppl_grg_results.tex
\resizebox{\textwidth}{!}{%
\setlength{\tabcolsep}{2pt} 
\begin{tabular}{@{}lccccccccccc@{}}
\toprule
\multicolumn{1}{c}{\multirow{2}{*}{\textbf{Model Variant}}} &
\multicolumn{5}{c}{\textbf{PadChest-GR}} &
\multicolumn{5}{c}{\textbf{VinDr-CXR (Zero-Shot)}} \\
\cmidrule(l){2-6} \cmidrule(l){7-11}
& \textbf{IoU} $\uparrow$ & \textbf{F1-Mi} $\uparrow$ & \textbf{F1-Ma} $\uparrow$ & \textbf{Cos.} $\uparrow$ & \textbf{CXS} $\uparrow$ &
  \textbf{IoU} $\uparrow$ & \textbf{F1-Mi} $\uparrow$ & \textbf{F1-Ma} $\uparrow$ & \textbf{Cos.} $\uparrow$ & \textbf{CXS} $\uparrow$ \\
\midrule
\multicolumn{11}{@{}l}{\textit{--- Baselines ---}} \\
MAIRA-2 & 0.256 $\pm$ 0.011 & \textbf{0.591} $\pm$ 0.015 & \textbf{0.321} $\pm$ 0.019 & \textbf{0.844} $\pm$ 0.004 & \textbf{0.616} $\pm$ 0.011 & 0.217 $\pm$ 0.007 & 0.546 $\pm$ 0.008 & 0.256 $\pm$ 0.011 & 0.824 $\pm$ 0.002 & 0.591 $\pm$ 0.005 \\
MedGemma-4B-IT & -- & 0.144 $\pm$ 0.009 & 0.203 $\pm$ 0.014 & 0.733 $\pm$ 0.003 & 0.517 $\pm$ 0.005 & -- & 0.209 $\pm$ 0.006 & 0.212 $\pm$ 0.008 & 0.779 $\pm$ 0.001 & 0.596 $\pm$ 0.003 \\
\midrule
\multicolumn{11}{@{}l}{\textit{--- Multi-task Fine-tuning Variants (v1--v15) ---}} \\
\textbf{v1}: Base (w/o Aug, w/o CL, w/o CIG, lr=2e-5) & 0.171 $\pm$ 0.009 & 0.557 $\pm$ 0.015 & 0.246 $\pm$ 0.012 & 0.829 $\pm$ 0.005 & 0.589 $\pm$ 0.011 & 0.207 $\pm$ 0.006 & 0.586 $\pm$ 0.008 & 0.247 $\pm$ 0.012 & 0.835 $\pm$ 0.003 & 0.630 $\pm$ 0.007 \\
\textbf{v2}: + Aug & 0.185 $\pm$ 0.010 & 0.564 $\pm$ 0.015 & 0.246 $\pm$ 0.012 & 0.839 $\pm$ 0.005 & \underline{0.599} $\pm$ 0.011 & 0.221 $\pm$ 0.007 & \underline{0.614} $\pm$ 0.008 & \underline{0.277} $\pm$ 0.011 & 0.834 $\pm$ 0.003 & 0.648 $\pm$ 0.007 \\
\textbf{v3}: + Aug + CL(1.5k) & 0.179 $\pm$ 0.010 & 0.564 $\pm$ 0.016 & 0.239 $\pm$ 0.011 & 0.835 $\pm$ 0.005 & 0.592 $\pm$ 0.011 & 0.224 $\pm$ 0.007 & 0.605 $\pm$ 0.008 & 0.251 $\pm$ 0.011 & 0.816 $\pm$ 0.003 & 0.630 $\pm$ 0.007 \\
\textbf{v4}: + Aug + CL(2k) & 0.193 $\pm$ 0.010 & 0.568 $\pm$ 0.015 & 0.246 $\pm$ 0.012 & 0.838 $\pm$ 0.005 & 0.596 $\pm$ 0.011 & 0.222 $\pm$ 0.006 & 0.611 $\pm$ 0.008 & 0.247 $\pm$ 0.010 & 0.842 $\pm$ 0.003 & \underline{0.651} $\pm$ 0.007 \\
\textbf{v5}: + Aug + CL(3k) & 0.180 $\pm$ 0.010 & \underline{0.578} $\pm$ 0.014 & 0.256 $\pm$ 0.010 & \underline{0.843} $\pm$ 0.005 & 0.595 $\pm$ 0.012 & 0.217 $\pm$ 0.007 & \textbf{0.626} $\pm$ 0.008 & 0.253 $\pm$ 0.009 & \underline{0.843} $\pm$ 0.003 & \textbf{0.671} $\pm$ 0.007 \\
\textbf{v6}: + Aug + CIG(1k) + CL(3k) & 0.195 $\pm$ 0.009 & 0.553 $\pm$ 0.015 & 0.279 $\pm$ 0.016 & 0.837 $\pm$ 0.005 & 0.591 $\pm$ 0.011 & 0.232 $\pm$ 0.006 & 0.582 $\pm$ 0.008 & \textbf{0.280} $\pm$ 0.013 & 0.838 $\pm$ 0.003 & 0.628 $\pm$ 0.007 \\
\textbf{v7}: + Aug + CIG(2k) + CL(3k) & 0.203 $\pm$ 0.010 & 0.535 $\pm$ 0.015 & 0.277 $\pm$ 0.021 & 0.829 $\pm$ 0.005 & 0.586 $\pm$ 0.011 & 0.227 $\pm$ 0.006 & 0.553 $\pm$ 0.008 & 0.252 $\pm$ 0.009 & 0.828 $\pm$ 0.003 & 0.590 $\pm$ 0.007 \\
\textbf{v8}: + Aug + CIG(3k) + CL(3k) & 0.207 $\pm$ 0.010 & 0.552 $\pm$ 0.015 & 0.293 $\pm$ 0.020 & 0.834 $\pm$ 0.005 & 0.582 $\pm$ 0.011 & 0.233 $\pm$ 0.006 & 0.568 $\pm$ 0.008 & 0.248 $\pm$ 0.009 & 0.841 $\pm$ 0.003 & 0.601 $\pm$ 0.007 \\
\textbf{v9}: + Aug + CIG(1k) + CL(3k) + lr=2e-4 & 0.253 $\pm$ 0.010 & 0.522 $\pm$ 0.016 & 0.313 $\pm$ 0.026 & 0.820 $\pm$ 0.004 & 0.563 $\pm$ 0.010 & 0.259 $\pm$ 0.007 & 0.491 $\pm$ 0.008 & 0.235 $\pm$ 0.009 & 0.814 $\pm$ 0.003 & 0.529 $\pm$ 0.006 \\
\textbf{v10}: + Aug + CIG(2k) + CL(3k) + lr=2e-4 & 0.258 $\pm$ 0.010 & 0.523 $\pm$ 0.015 & 0.265 $\pm$ 0.013 & 0.820 $\pm$ 0.005 & 0.569 $\pm$ 0.010 & 0.262 $\pm$ 0.007 & 0.449 $\pm$ 0.008 & 0.236 $\pm$ 0.008 & 0.799 $\pm$ 0.003 & 0.477 $\pm$ 0.006 \\
\textbf{v11 (CURE)}: + Aug + CIG(3k) + CL(3k) + lr=2e-4 & \underline{0.265} $\pm$ 0.011 & 0.507 $\pm$ 0.015 & 0.270 $\pm$ 0.013 & 0.819 $\pm$ 0.005 & 0.574 $\pm$ 0.010 & 0.262 $\pm$ 0.007 & 0.505 $\pm$ 0.008 & 0.246 $\pm$ 0.009 & 0.832 $\pm$ 0.003 & 0.540 $\pm$ 0.007 \\
\textbf{v12}: + Aug + CIG(3k) + Uni(Inter)/Nat(Intra) + lr=2e-4 & 0.263 $\pm$ 0.010 & 0.529 $\pm$ 0.015 & 0.277 $\pm$ 0.014 & 0.823 $\pm$ 0.005 & 0.575 $\pm$ 0.010 & \underline{0.264} $\pm$ 0.007 & 0.489 $\pm$ 0.008 & 0.244 $\pm$ 0.009 & 0.819 $\pm$ 0.003 & 0.514 $\pm$ 0.007 \\
\textbf{v13}: + Aug + CIG(3k) + CL(Inter,3k)/Nat(Intra) + lr=2e-4 & \textbf{0.272} $\pm$ 0.010 & 0.503 $\pm$ 0.016 & 0.280 $\pm$ 0.018 & 0.812 $\pm$ 0.005 & 0.550 $\pm$ 0.010 & \textbf{0.266} $\pm$ 0.007 & 0.434 $\pm$ 0.008 & 0.231 $\pm$ 0.009 & 0.795 $\pm$ 0.003 & 0.469 $\pm$ 0.006 \\
\textbf{v14}: + Aug + CIG(3k) + Uni(Inter)/CL(Intra,3k) + lr=2e-4 & 0.246 $\pm$ 0.010 & 0.544 $\pm$ 0.015 & \underline{0.317} $\pm$ 0.019 & 0.824 $\pm$ 0.005 & 0.570 $\pm$ 0.010 & 0.262 $\pm$ 0.007 & 0.496 $\pm$ 0.008 & 0.245 $\pm$ 0.010 & 0.815 $\pm$ 0.003 & 0.536 $\pm$ 0.007 \\
\textbf{v15}: + Aug + CIG(3k) + Nat(Inter)/Nat(Intra) + lr=2e-4 & 0.000 $\pm$ 0.000 & 0.498 $\pm$ 0.016 & 0.193 $\pm$ 0.016 & 0.788 $\pm$ 0.005 & 0.483 $\pm$ 0.011 & 0.000 $\pm$ 0.000 & 0.604 $\pm$ 0.009 & 0.193 $\pm$ 0.016 & \textbf{0.858} $\pm$ 0.003 & 0.534 $\pm$ 0.006 \\
\bottomrule
\end{tabular}%
}

%% file: tables_suppl_mimiccxr_report_gen_results.tex
\resizebox{\textwidth}{!}{%
\setlength{\tabcolsep}{2pt} 
\begin{tabular}{@{}lcccccccccc@{}}
\toprule
\textbf{Model / Inference Protocol} &
\textbf{F1-Ma} $\uparrow$ & \textbf{F1-Mi} $\uparrow$ &
\textbf{P-Ma} $\uparrow$ & \textbf{P-Mi} $\uparrow$ &
\textbf{R-Ma} $\uparrow$ & \textbf{R-Mi} $\uparrow$ &
\textbf{Cos.} $\uparrow$ & \textbf{CXS} $\uparrow$ & \textbf{RaTES} $\uparrow$ & \textbf{RadF1} $\uparrow$\\
\midrule
\multicolumn{10}{@{}l}{\textit{--- Baselines ---}} \\
CXRMate-RRG24 & \underline{0.414} $\pm$ 0.006 & \textbf{0.589} $\pm$ 0.004 & \textbf{0.493} $\pm$ 0.012 & 0.617 $\pm$ 0.005 & 0.415 $\pm$ 0.006 & 0.563 $\pm$ 0.005 & 0.764 $\pm$ 0.001 & \textbf{0.656} $\pm$ 0.002 & 0.577 $\pm$ 0.002 & \textbf{0.255} $\pm$ 0.002 \\
MAIRA-2 (w/ grounding) & 0.304 $\pm$ 0.006 & 0.489 $\pm$ 0.005 & 0.442 $\pm$ 0.021 & \textbf{0.639} $\pm$ 0.006 & 0.283 $\pm$ 0.006 & 0.397 $\pm$ 0.005 & 0.751 $\pm$ 0.002 & 0.603 $\pm$ 0.002 & 0.496 $\pm$ 0.002 & 0.120 $\pm$ 0.002 \\
MAIRA-2 (w/o grounding) & 0.386 $\pm$ 0.006 & 0.554 $\pm$ 0.004 & 0.425 $\pm$ 0.009 & 0.578 $\pm$ 0.005 & 0.384 $\pm$ 0.006 & 0.533 $\pm$ 0.005 & 0.693 $\pm$ 0.002 & 0.576 $\pm$ 0.002 & 0.501 $\pm$ 0.002 & 0.143 $\pm$ 0.002 \\
MedGemma-4B-IT & 0.382 $\pm$ 0.004 & 0.547 $\pm$ 0.004 & 0.332 $\pm$ 0.005 & 0.452 $\pm$ 0.004 & 0.494 $\pm$ 0.005 & 0.692 $\pm$ 0.005 & 0.714 $\pm$ 0.001 & 0.580 $\pm$ 0.002 & 0.532 $\pm$ 0.001 & 0.112 $\pm$ 0.001 \\
\midrule
\multicolumn{10}{@{}l}{\textit{--- Specialized Fine-tuning ---}} \\
MedGemma-FT (RG only) & 0.353 $\pm$ 0.006 & 0.520 $\pm$ 0.005 & \underline{0.469} $\pm$ 0.016 & \underline{0.621} $\pm$ 0.006 & 0.323 $\pm$ 0.005 & 0.447 $\pm$ 0.005 & 0.753 $\pm$ 0.002 & 0.624 $\pm$ 0.002 & 0.536 $\pm$ 0.002 & \underline{0.203} $\pm$ 0.002 \\
\midrule
\multicolumn{10}{@{}l}{\textit{--- CURE Inference Strategies (Single Model) ---}} \\
CURE (GRG Prompt) & 0.314 $\pm$ 0.006 & 0.463 $\pm$ 0.005 & 0.442 $\pm$ 0.009 & 0.605 $\pm$ 0.006 & 0.290 $\pm$ 0.006 & 0.376 $\pm$ 0.004 & 0.725 $\pm$ 0.002 & 0.526 $\pm$ 0.002 & 0.447 $\pm$ 0.002 & 0.077 $\pm$ 0.002 \\
CURE (AGRG-9) & 0.230 $\pm$ 0.004 & 0.443 $\pm$ 0.005 & 0.432 $\pm$ 0.014 & \textbf{0.639} $\pm$ 0.006 & 0.225 $\pm$ 0.004 & 0.339 $\pm$ 0.004 & 0.762 $\pm$ 0.002 & 0.608 $\pm$ 0.002 & 0.557 $\pm$ 0.002 & 0.200 $\pm$ 0.002 \\
CURE (AGRG-9 + GRG) & 0.355 $\pm$ 0.006 & 0.528 $\pm$ 0.004 & 0.436 $\pm$ 0.008 & 0.600 $\pm$ 0.006 & 0.342 $\pm$ 0.006 & 0.472 $\pm$ 0.005 & 0.784 $\pm$ 0.001 & 0.640 $\pm$ 0.002 & 0.572 $\pm$ 0.002 & 0.200 $\pm$ 0.002 \\
CURE (AGRG-29) & 0.400 $\pm$ 0.005 & 0.559 $\pm$ 0.004 & 0.355 $\pm$ 0.008 & 0.446 $\pm$ 0.004 & 0.539 $\pm$ 0.005 & 0.749 $\pm$ 0.004 & 0.783 $\pm$ 0.001 & 0.645 $\pm$ 0.002 & \underline{0.592} $\pm$ 0.001 & 0.181 $\pm$ 0.001 \\
CURE (AGRG-29 + GRG) & \textbf{0.415} $\pm$ 0.005 & \underline{0.562} $\pm$ 0.004 & 0.365 $\pm$ 0.010 & 0.439 $\pm$ 0.004 & 0.582 $\pm$ 0.005 & \underline{0.781} $\pm$ 0.004 & 0.792 $\pm$ 0.001 & \underline{0.655} $\pm$ 0.002 & \textbf{0.597} $\pm$ 0.001 & 0.176 $\pm$ 0.001 \\
CURE (AGRG-38) & 0.395 $\pm$ 0.004 & 0.534 $\pm$ 0.004 & 0.354 $\pm$ 0.017 & 0.408 $\pm$ 0.004 & \underline{0.593} $\pm$ 0.005 & 0.770 $\pm$ 0.004 & \underline{0.793} $\pm$ 0.001 & 0.632 $\pm$ 0.001 & 0.577 $\pm$ 0.001 & 0.172 $\pm$ 0.001 \\
CURE (AGRG-38 + GRG) & 0.406 $\pm$ 0.004 & 0.536 $\pm$ 0.004 & 0.360 $\pm$ 0.022 & 0.404 $\pm$ 0.004 & \textbf{0.628} $\pm$ 0.005 & \textbf{0.798} $\pm$ 0.004 & \textbf{0.800} $\pm$ 0.001 & 0.642 $\pm$ 0.001 & 0.583 $\pm$ 0.001 & 0.169 $\pm$ 0.001 \\
\bottomrule
\end{tabular}%
}

%% file: tables_suppl_cig_hallucination_results_part1.tex
\resizebox{\textwidth}{!}{%
\begin{tabular}{l l c c c c c c c c }
\toprule
Anatomy & Model & Abn. Halluc. $\downarrow$ & Abn. Corr. $\uparrow$ & Dev. Halluc. $\downarrow$ & Dev. Corr. $\uparrow$ & Contra. $\downarrow$ & Entail. $\uparrow$ & Neutral & N \\
\midrule
Abdomen & MAIRA-2 & 24.0\% & \textbf{6.0\%} & \textbf{3.0\%} & 0.7\% & 35.7\% & 16.3\% & 48.0\% & 300 \\
 & CURE & \textbf{2.0\%} & 1.7\% & 14.3\% & \textbf{33.0\%} & \textbf{9.0\%} & \textbf{54.3\%} & 36.7\% & 300 \\
\midrule
Aortic Arch & MAIRA-2 & \textbf{1.3\%} & 0.3\% & \textbf{0.0\%} & 0.0\% & \textbf{0.3\%} & 16.3\% & 83.3\% & 300 \\
 & CURE & 64.7\% & \textbf{26.3\%} & 2.0\% & \textbf{1.7\%} & 6.3\% & \textbf{26.0\%} & 67.7\% & 300 \\
\midrule
Cardiac Silhouette & MAIRA-2 & \textbf{2.0\%} & 2.7\% & \textbf{0.0\%} & 0.3\% & \textbf{8.0\%} & 25.0\% & 67.0\% & 300 \\
 & CURE & 25.7\% & \textbf{26.3\%} & 1.0\% & \textbf{6.3\%} & 27.7\% & \textbf{47.3\%} & 25.0\% & 300 \\
\midrule
Carina & MAIRA-2 & \textbf{0.0\%} & 1.0\% & \textbf{0.0\%} & 0.0\% & \textbf{1.0\%} & 5.7\% & 93.3\% & 300 \\
 & CURE & 6.0\% & \textbf{2.3\%} & 41.0\% & \textbf{14.7\%} & 34.0\% & \textbf{11.0\%} & 55.0\% & 300 \\
\midrule
Cavoatrial Junction & MAIRA-2 & \textbf{0.0\%} & 0.0\% & \textbf{0.0\%} & 0.3\% & \textbf{0.3\%} & 12.3\% & 87.3\% & 300 \\
 & CURE & 6.3\% & \textbf{5.7\%} & 32.3\% & \textbf{45.3\%} & 15.0\% & \textbf{47.0\%} & 38.0\% & 300 \\
\midrule
Left Apical Zone & MAIRA-2 & 7.0\% & 2.0\% & \textbf{0.0\%} & 0.7\% & \textbf{3.7\%} & 3.7\% & 92.7\% & 300 \\
 & CURE & \textbf{2.0\%} & \textbf{10.3\%} & \textbf{0.0\%} & \textbf{4.0\%} & 14.0\% & \textbf{61.0\%} & 25.0\% & 300 \\
\midrule
Left Arm & MAIRA-2 & 21.3\% & 2.0\% & \textbf{0.7\%} & 0.7\% & \textbf{13.3\%} & 6.7\% & 80.0\% & 300 \\
 & CURE & \textbf{12.7\%} & \textbf{18.7\%} & 1.7\% & \textbf{7.7\%} & 23.7\% & \textbf{42.7\%} & 33.7\% & 300 \\
\midrule
Left Breast & MAIRA-2 & 27.0\% & 3.7\% & \textbf{1.3\%} & 0.7\% & \textbf{13.7\%} & 6.0\% & 80.3\% & 300 \\
 & CURE & \textbf{10.3\%} & \textbf{20.0\%} & \textbf{1.3\%} & \textbf{7.7\%} & 21.0\% & \textbf{44.3\%} & 34.7\% & 300 \\
\midrule
Left Chest Wall & MAIRA-2 & 19.3\% & 4.7\% & 11.0\% & 8.3\% & 40.0\% & 9.3\% & 50.7\% & 300 \\
 & CURE & \textbf{4.3\%} & \textbf{18.0\%} & \textbf{5.3\%} & \textbf{28.0\%} & \textbf{11.7\%} & \textbf{68.3\%} & 20.0\% & 300 \\
\midrule
Left Clavicle & MAIRA-2 & 59.0\% & 2.3\% & \textbf{0.0\%} & 0.3\% & 22.7\% & 5.0\% & 72.3\% & 300 \\
 & CURE & \textbf{1.0\%} & \textbf{4.3\%} & 8.3\% & \textbf{11.3\%} & \textbf{7.0\%} & \textbf{32.7\%} & 60.3\% & 300 \\
\midrule
Left Costophrenic Angle & MAIRA-2 & \textbf{2.0\%} & 1.0\% & \textbf{0.0\%} & 0.0\% & \textbf{2.0\%} & 1.3\% & 96.7\% & 300 \\
 & CURE & 9.3\% & \textbf{18.7\%} & \textbf{0.0\%} & \textbf{2.0\%} & 26.3\% & \textbf{53.7\%} & 20.0\% & 300 \\
\midrule
Left Hemidiaphragm & MAIRA-2 & \textbf{0.7\%} & 0.7\% & \textbf{0.0\%} & 0.0\% & \textbf{0.3\%} & 4.3\% & 95.3\% & 300 \\
 & CURE & 11.7\% & \textbf{9.0\%} & 8.0\% & \textbf{23.3\%} & 6.0\% & \textbf{33.7\%} & 60.3\% & 300 \\
\midrule
Left Hilar Structures & MAIRA-2 & 17.7\% & 2.7\% & \textbf{0.0\%} & 0.7\% & \textbf{6.3\%} & 5.0\% & 88.7\% & 300 \\
 & CURE & \textbf{9.7\%} & \textbf{31.3\%} & \textbf{0.0\%} & \textbf{1.3\%} & 28.3\% & \textbf{52.3\%} & 19.3\% & 300 \\
\midrule
Left Lower Lung Zone & MAIRA-2 & \textbf{11.7\%} & 10.0\% & \textbf{0.0\%} & 0.0\% & \textbf{8.3\%} & 11.7\% & 80.0\% & 300 \\
 & CURE & 25.3\% & \textbf{40.0\%} & \textbf{0.0\%} & \textbf{5.7\%} & 17.3\% & \textbf{53.0\%} & 29.7\% & 300 \\
\midrule
Left Lung & MAIRA-2 & 12.0\% & 9.3\% & 0.3\% & 3.0\% & 56.3\% & 21.7\% & 22.0\% & 300 \\
 & CURE & \textbf{7.0\%} & \textbf{12.0\%} & \textbf{0.0\%} & \textbf{3.3\%} & \textbf{32.3\%} & \textbf{41.7\%} & 26.0\% & 300 \\
\midrule
Left Mid Lung Zone & MAIRA-2 & \textbf{11.0\%} & 4.7\% & \textbf{0.0\%} & 0.0\% & \textbf{4.7\%} & 4.3\% & 91.0\% & 300 \\
 & CURE & 61.0\% & \textbf{37.7\%} & \textbf{0.0\%} & \textbf{2.7\%} & 40.3\% & \textbf{29.3\%} & 30.3\% & 300 \\
\midrule
Left Shoulder & MAIRA-2 & 31.3\% & 2.7\% & \textbf{1.3\%} & 0.3\% & \textbf{15.7\%} & 4.7\% & 79.7\% & 300 \\
 & CURE & \textbf{10.3\%} & \textbf{20.0\%} & 5.7\% & \textbf{8.0\%} & 22.7\% & \textbf{44.0\%} & 33.3\% & 300 \\
\midrule
Left Upper Lung Zone & MAIRA-2 & \textbf{15.3\%} & 8.3\% & \textbf{0.0\%} & 0.0\% & \textbf{8.0\%} & 7.3\% & 84.7\% & 300 \\
 & CURE & 38.7\% & \textbf{38.3\%} & \textbf{0.0\%} & \textbf{2.3\%} & 15.7\% & \textbf{38.3\%} & 46.0\% & 300 \\
\midrule
Mediastinum & MAIRA-2 & 23.7\% & \textbf{14.3\%} & \textbf{0.0\%} & 4.3\% & 67.0\% & 14.0\% & 19.0\% & 300 \\
 & CURE & \textbf{5.3\%} & 6.0\% & 12.3\% & \textbf{33.7\%} & \textbf{20.7\%} & \textbf{49.7\%} & 29.7\% & 300 \\
\bottomrule
\end{tabular}%
}

%% file: tables_suppl_cig_hallucination_results_part2.tex
\resizebox{\textwidth}{!}{%
\begin{tabular}{l l c c c c c c c c }
\toprule
Anatomy & Model & Abn. Halluc. $\downarrow$ & Abn. Corr. $\uparrow$ & Dev. Halluc. $\downarrow$ & Dev. Corr. $\uparrow$ & Contra. $\downarrow$ & Entail. $\uparrow$ & Neutral & N \\
\midrule
Neck & MAIRA-2 & 17.0\% & \textbf{3.7\%} & \textbf{2.7\%} & 1.7\% & 58.3\% & 4.3\% & 37.3\% & 300 \\
 & CURE & \textbf{1.0\%} & 2.0\% & 27.0\% & \textbf{62.0\%} & \textbf{20.7\%} & \textbf{41.7\%} & 37.7\% & 300 \\
\midrule
Right Apical Zone & MAIRA-2 & 7.3\% & 3.7\% & \textbf{0.0\%} & 2.0\% & \textbf{5.7\%} & 6.3\% & 88.0\% & 300 \\
 & CURE & \textbf{6.3\%} & \textbf{13.0\%} & \textbf{0.0\%} & \textbf{8.0\%} & 20.0\% & \textbf{54.0\%} & 26.0\% & 300 \\
\midrule
Right Arm & MAIRA-2 & 21.7\% & 2.0\% & \textbf{1.7\%} & 1.0\% & \textbf{16.3\%} & 5.3\% & 78.3\% & 300 \\
 & CURE & \textbf{7.0\%} & \textbf{16.3\%} & 2.3\% & \textbf{5.3\%} & 23.0\% & \textbf{40.0\%} & 37.0\% & 300 \\
\midrule
Right Atrium & MAIRA-2 & \textbf{0.7\%} & 0.3\% & \textbf{0.0\%} & 0.3\% & \textbf{1.0\%} & 11.3\% & 87.7\% & 300 \\
 & CURE & 8.3\% & \textbf{3.3\%} & 37.0\% & \textbf{45.3\%} & 12.3\% & \textbf{39.7\%} & 48.0\% & 300 \\
\midrule
Right Breast & MAIRA-2 & 21.0\% & 1.0\% & \textbf{1.3\%} & 1.0\% & \textbf{11.3\%} & 5.7\% & 83.0\% & 300 \\
 & CURE & \textbf{9.7\%} & \textbf{19.3\%} & 3.0\% & \textbf{4.7\%} & 22.0\% & \textbf{41.3\%} & 36.7\% & 300 \\
\midrule
Right Chest Wall & MAIRA-2 & 23.0\% & 6.0\% & \textbf{5.0\%} & 3.7\% & 39.7\% & 10.7\% & 49.7\% & 300 \\
 & CURE & \textbf{2.0\%} & \textbf{23.7\%} & 6.0\% & \textbf{20.0\%} & \textbf{12.7\%} & \textbf{67.0\%} & 20.3\% & 300 \\
\midrule
Right Clavicle & MAIRA-2 & 62.7\% & 0.3\% & \textbf{0.0\%} & 0.7\% & 20.3\% & 1.7\% & 78.0\% & 300 \\
 & CURE & \textbf{1.0\%} & \textbf{4.3\%} & 4.3\% & \textbf{8.7\%} & \textbf{7.3\%} & \textbf{27.7\%} & 65.0\% & 300 \\
\midrule
Right Costophrenic Angle & MAIRA-2 & \textbf{1.3\%} & 4.7\% & \textbf{0.0\%} & 0.3\% & \textbf{0.3\%} & 5.7\% & 94.0\% & 300 \\
 & CURE & 10.7\% & \textbf{25.0\%} & 0.3\% & \textbf{3.3\%} & 25.3\% & \textbf{49.0\%} & 25.7\% & 300 \\
\midrule
Right Hemidiaphragm & MAIRA-2 & \textbf{0.7\%} & 0.0\% & \textbf{0.0\%} & 0.0\% & \textbf{0.0\%} & 6.3\% & 93.7\% & 300 \\
 & CURE & 8.7\% & \textbf{8.7\%} & 7.3\% & \textbf{21.7\%} & 8.7\% & \textbf{45.3\%} & 46.0\% & 300 \\
\midrule
Right Hilar Structures & MAIRA-2 & 17.0\% & 2.7\% & \textbf{0.0\%} & 0.7\% & \textbf{7.0\%} & 6.3\% & 86.7\% & 300 \\
 & CURE & \textbf{13.0\%} & \textbf{32.3\%} & \textbf{0.0\%} & \textbf{2.3\%} & 31.3\% & \textbf{49.0\%} & 19.7\% & 300 \\
\midrule
Right Lower Lung Zone & MAIRA-2 & \textbf{8.0\%} & 5.3\% & \textbf{0.0\%} & 0.0\% & \textbf{6.0\%} & 5.3\% & 88.7\% & 300 \\
 & CURE & 25.3\% & \textbf{43.7\%} & \textbf{0.0\%} & \textbf{4.0\%} & 18.7\% & \textbf{53.3\%} & 28.0\% & 300 \\
\midrule
Right Lung & MAIRA-2 & \textbf{10.7\%} & 9.7\% & \textbf{0.0\%} & 3.0\% & 53.7\% & 29.3\% & 17.0\% & 300 \\
 & CURE & 11.7\% & \textbf{21.3\%} & 0.3\% & \textbf{7.3\%} & \textbf{27.0\%} & \textbf{46.0\%} & 27.0\% & 300 \\
\midrule
Right Mid Lung Zone & MAIRA-2 & \textbf{8.7\%} & 2.0\% & \textbf{0.0\%} & 0.0\% & \textbf{6.0\%} & 2.3\% & 91.7\% & 300 \\
 & CURE & 66.0\% & \textbf{33.0\%} & \textbf{0.0\%} & \textbf{2.7\%} & 43.3\% & \textbf{23.0\%} & 33.7\% & 300 \\
\midrule
Right Shoulder & MAIRA-2 & 32.0\% & 1.3\% & \textbf{0.7\%} & 0.0\% & \textbf{16.3\%} & 4.7\% & 79.0\% & 300 \\
 & CURE & \textbf{8.7\%} & \textbf{22.7\%} & 10.0\% & \textbf{16.3\%} & 23.7\% & \textbf{45.3\%} & 31.0\% & 300 \\
\midrule
Right Upper Lung Zone & MAIRA-2 & \textbf{9.7\%} & 3.0\% & \textbf{0.0\%} & 0.0\% & \textbf{6.3\%} & 3.3\% & 90.3\% & 300 \\
 & CURE & 47.0\% & \textbf{36.0\%} & \textbf{0.0\%} & \textbf{4.0\%} & 24.7\% & \textbf{31.7\%} & 43.7\% & 300 \\
\midrule
Spine & MAIRA-2 & 12.7\% & 6.0\% & 5.7\% & \textbf{4.7\%} & 38.3\% & 13.0\% & 48.7\% & 300 \\
 & CURE & \textbf{6.3\%} & \textbf{11.7\%} & \textbf{0.7\%} & 2.7\% & \textbf{3.3\%} & \textbf{41.7\%} & 55.0\% & 300 \\
\midrule
SVC & MAIRA-2 & 9.3\% & 1.0\% & \textbf{4.0\%} & 5.0\% & \textbf{11.0\%} & 19.0\% & 70.0\% & 300 \\
 & CURE & \textbf{5.7\%} & \textbf{1.7\%} & 34.3\% & \textbf{42.0\%} & 17.7\% & \textbf{34.7\%} & 47.7\% & 300 \\
\midrule
Trachea & MAIRA-2 & \textbf{7.7\%} & 0.3\% & \textbf{0.0\%} & 0.0\% & 26.3\% & 16.0\% & 57.7\% & 300 \\
 & CURE & 19.3\% & \textbf{2.0\%} & 21.0\% & \textbf{21.7\%} & \textbf{24.3\%} & \textbf{19.7\%} & 56.0\% & 300 \\
\midrule
Upper Mediastinum & MAIRA-2 & 10.0\% & 5.7\% & \textbf{0.3\%} & 4.0\% & 33.0\% & 16.3\% & 50.7\% & 300 \\
 & CURE & \textbf{7.7\%} & \textbf{10.7\%} & 4.3\% & \textbf{8.7\%} & \textbf{21.7\%} & \textbf{61.0\%} & 17.3\% & 300 \\
\midrule
\textbf{Mean} & \textbf{MAIRA-2} & \textbf{14.9\%} & 3.6\% & \textbf{1.0\%} & 1.3\% & \textbf{17.5\%} & 9.3\% & 73.2\% & 300 \\
\rowcolor{gray!15}
\textbf{Mean} & \textbf{CURE} & 15.2\% & \textbf{17.8\%} & 7.7\% & \textbf{14.0\%} & 20.2\% & \textbf{43.2\%} & 36.6\% & 300 \\
\bottomrule
\end{tabular}%
}

%% file: tables_suppl_gemini_eval_example.tex
\footnotesize
\renewcommand{\arraystretch}{1.3}

\begin{tabularx}{\textwidth}{l X}
\toprule

\multicolumn{2}{c}{%
    \begin{minipage}[m]{0.2\textwidth}
        \centering
        \includegraphics[width=\linewidth, keepaspectratio]{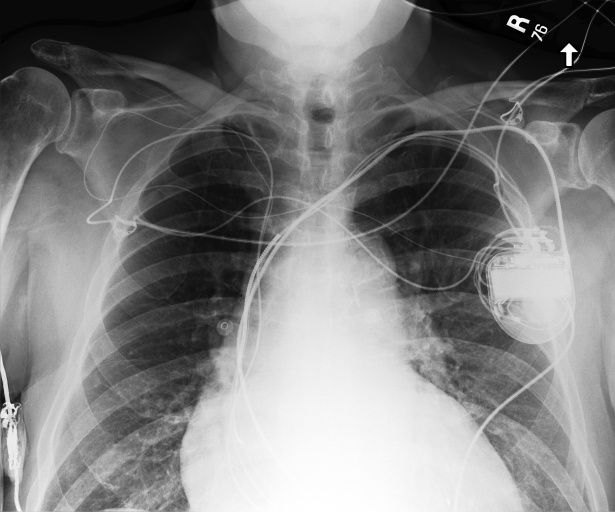} 
    \end{minipage}%
    \hfill
    \begin{minipage}[m]{0.75\textwidth}
        \textbf{Context: Ground Truth (GT) Report} \newline
        \textit{``AP upright portable chest radiograph is obtained. A left chest wall pacer device is again seen with lead tips extending into the right atrium and ventricle. Abandoned pacing leads are also seen in the right chest wall, extending into the right heart, not significantly changed. The heart is mildly enlarged. The lungs appear clear without definite signs of pneumonia or CHF. No large effusion or pneumothorax is seen. The overall cardiomediastinal silhouette is stable. \textbf{Bony structures are intact}. No acute findings in the chest. Stable mild cardiomegaly. Multiple pacer wires are unchanged in position.''}
    \end{minipage}%
} \\

\multicolumn{2}{l}{\textbf{Target Anatomy:} Left Clavicle} \\
\midrule
\textbf{Model} & \textbf{Generated Output \& Gemini 2.5 Flash-Lite's Evaluation} \\
\midrule

\textbf{MAIRA-2} & 
\textbf{Generation:} \newline
\raw{Left clavicle fracture is noted. [0.645, 0.145, 0.995, 0.345]} \par\vspace{0.5em}
\textbf{Gemini 2.5 Flash-Lite's Evaluation:} \newline
\raw{{
  "reason": "The Ground Truth ([GT]) report describes a left chest wall pacer device, abandoned pacing leads in the right chest wall, mild cardiomegaly, and stable findings, but it explicitly states that 'Bony structures are intact' and 'No acute findings in the chest.' The Mini-Report ([GEN]) states 'Left clavicle fracture is noted.' Since the GT explicitly states bony structures are intact and no acute findings are seen, the mention of a left clavicle fracture in [GEN] is a clear contradiction to the findings in [GT].\n\n- gt_has_abnormalities: Yes, due to mild cardiomegaly.\n- gt_has_devices: Yes, due to the presence of pacer devices and leads.\n- gen_has_abnormalities: Yes, a fracture is an abnormality.\n- gen_has_devices: No devices are mentioned in [GEN].\n- gen_has_correct_abnormalities: No, the fracture is not supported; in fact, bony structures are reported as intact.\n- gen_has_hallucinated_abnormalities: Yes, the fracture is not mentioned in [GT] and contradicts the statement that bony structures are intact.\n- gen_has_correct_devices: No, [GEN] does not mention devices.\n- gen_has_hallucinated_devices: No, [GEN] does not mention devices.\n- nli_status: Contradiction, because [GEN] reports a fracture while [GT] reports bony structures as intact and no acute findings.",
  "gt_has_abnormalities": "yes",
  "gt_has_devices": "yes",
  "gen_has_abnormalities": "yes",
  "gen_has_devices": "no",
  "gen_has_correct_abnormalities": "no",
  "gen_has_hallucinated_abnormalities": "yes",
  "gen_has_correct_devices": "no",
  "gen_has_hallucinated_devices": "no",
  "nli_status": "contradiction"
}}
\\

\midrule

\textbf{CURE} & 
\textbf{Generation:} \newline
\raw{Location of the left clavicle: [0.75,0.28,0.36,0.19]. Description: No acute osseous abnormalities.} \par\vspace{0.5em}
\textbf{Gemini 2.5 Flash-Lite's Evaluation:} \newline
\raw{{
  "reason": "The ground truth report [GT] states, \"Bony structures are intact\" and \"No acute findings in the chest,\" which implies no acute osseous abnormalities. The mini-report [GEN] states, \"No acute osseous abnormalities.\" This statement is directly supported by [GT].\n\nFor the initial assessment:\n- [GT] mentions a pacer device and leads, so gt_has_devices is 'yes'. It mentions mild cardiomegaly, so gt_has_abnormalities is 'yes'.\n- [GEN] mentions no abnormalities, so gen_has_abnormalities is 'no'. [GEN] mentions no devices, so gen_has_devices is 'no'.\n\nComparison:\n- [GEN] does not affirm any abnormality, so gen_has_correct_abnormalities and gen_has_hallucinated_abnormalities are 'no'.\n- [GEN] does not affirm any device, so gen_has_correct_devices and gen_has_hallucinated_devices are 'no'.\n- Since the statement in [GEN] (\"No acute osseous abnormalities\") is directly supported by the statement in [GT] (\"Bony structures are intact\"), the NLI status is 'entailment'.",
  "gt_has_abnormalities": "yes",
  "gt_has_devices": "yes",
  "gen_has_abnormalities": "no",
  "gen_has_devices": "no",
  "gen_has_correct_abnormalities": "no",
  "gen_has_hallucinated_abnormalities": "no",
  "gen_has_correct_devices": "no",
  "gen_has_hallucinated_devices": "no",
  "nli_status": "entailment"
}}
\\

\bottomrule
\end{tabularx}